\documentclass[twoside]{article}

\usepackage[accepted]{aistats2024}
\usepackage{hyperref}
\usepackage[capitalize,noabbrev]{cleveref}
\usepackage{url}
\usepackage{amsthm,amsmath,amssymb}
\usepackage[pdftex]{graphicx}
\usepackage[labelformat=simple]{subcaption}
\usepackage{wrapfig}
\usepackage[ruled]{algorithm2e}
\usepackage{hhline}
\usepackage{enumitem}
\usepackage{comment}
\usepackage{dsfont}

\theoremstyle{plain}
\newtheorem{theorem}{Theorem}[section]
\newtheorem{proposition}[theorem]{Proposition}
\newtheorem{lemma}[theorem]{Lemma}

\newtheorem{definition}[theorem]{Definition}

\theoremstyle{remark}
\newtheorem{remark}{Remark}[section]

\usepackage[round]{natbib}

\usepackage{amsmath,amsfonts,bm}

\def\1{\bm{1}}

\def\rvb{{\mathbf{b}}}

\def\rve{{\mathbf{e}}}

\def\rvg{{\mathbf{g}}}
\def\rvh{{\mathbf{h}}}
\def\rvu{{\mathbf{i}}}

\def\rvs{{\mathbf{s}}}

\def\rvu{{\mathbf{u}}}
\def\rvv{{\mathbf{v}}}

\def\rvx{{\mathbf{x}}}
\def\rvy{{\mathbf{y}}}

\def\rmA{{\mathbf{A}}}
\def\rmB{{\mathbf{B}}}
\def\rmC{{\mathbf{C}}}
\def\rmD{{\mathbf{D}}}

\def\rmH{{\mathbf{H}}}
\def\rmI{{\mathbf{I}}}
\def\rmJ{{\mathbf{J}}}
\def\rmK{{\mathbf{K}}}

\def\rmM{{\mathbf{M}}}

\def\rmP{{\mathbf{P}}}
\def\rmQ{{\mathbf{Q}}}

\def\rmU{{\mathbf{U}}}
\def\rmV{{\mathbf{V}}}

\def\vzero{{\bm{0}}}
\def\vone{{\bm{1}}}
\def\vmu{{\bm{\mu}}}
\def\vtheta{{\bm{\theta}}}
\def\vpi{{\bm{\pi}}}

\def\mP{{\bm{P}}}

\def\mV{{\bm{V}}}

\DeclareMathAlphabet{\mathsfit}{\encodingdefault}{\sfdefault}{m}{sl}
\SetMathAlphabet{\mathsfit}{bold}{\encodingdefault}{\sfdefault}{bx}{n}

\def\gA{{\mathcal{A}}}

\def\gE{{\mathcal{E}}}
\def\gF{{\mathcal{F}}}

\def\gN{{\mathcal{N}}}

\def\gS{{\mathcal{S}}}

\def\gV{{\mathcal{V}}}

\def\sP{{\mathbb{P}}}

\def\sR{{\mathbb{R}}}

\newcommand{\E}{\mathbb{E}}

\newcommand{\R}{\mathbb{R}}

\begin{document}

%
\runningtitle{CLT for TTSA with Markovian Noise: Theory and Applications}

%

\twocolumn[

\aistatstitle{Central Limit Theorem for Two-Timescale Stochastic Approximation with Markovian Noise: Theory and Applications}

\aistatsauthor{ Jie Hu \And Vishwaraj Doshi \And  Do Young Eun }

\aistatsaddress{ North Carolina State University \And   IQVIA Inc. \And North Carolina State University } ]

\begin{abstract}
Two-timescale stochastic approximation (TTSA) is among the most general frameworks for iterative stochastic algorithms. This includes well-known stochastic optimization methods such as SGD variants and those designed for bilevel or minimax problems, as well as reinforcement learning like the family of gradient-based temporal difference (GTD) algorithms. In this paper, we conduct an in-depth asymptotic analysis of TTSA under controlled Markovian noise via central limit theorem (CLT), uncovering the coupled dynamics of TTSA influenced by the underlying Markov chain, which has not been addressed by previous CLT results of TTSA only with Martingale difference noise.  Building upon our CLT, we expand its application horizon of efficient sampling strategies from vanilla SGD to a wider TTSA context in distributed learning, thus broadening the scope of \citet{hu2022efficiency}. In addition, we leverage our CLT result to deduce the statistical properties of GTD algorithms with nonlinear function approximation using Markovian samples and show their identical asymptotic performance, a perspective not evident from current finite-time bounds.
\end{abstract}

\section{INTRODUCTION}

Two-timescale stochastic approximation (TTSA) serves as a cornerstone algorithm for identifying the root $(\rvx^*,\rvy^*)$ of two coupled functions, i.e.,
\vspace{-0mm}
\begin{equation}\label{eqn:root_finding_problem}
\vspace{-0mm}
\begin{split}
    &\bar{h}_1 (\rvx^*,\rvy^*) \triangleq \E_{\xi\sim \vmu}[h_1(\rvx^*,\rvy^*,\xi)] = 0,\\ \vspace{-0mm}
    &\bar{h}_2(\rvx^*,\rvy^*) \triangleq \E_{\xi\sim \vmu}[h_2(\rvx^*,\rvy^*,\xi)] = 0,
\end{split}
\end{equation}
where $\vmu$ is a probability vector and typically only noisy observations $h_1(\rvx,\rvy,\xi), h_2(\rvx,\rvy,\xi)$ are accessible \citep{kushner2003stochastic,borkar2009stochastic}. If either of the two functions $h_1, h_2$ is decoupled, e.g., $h_1(\rvx,\rvy,\xi) \equiv h_1(\rvx,\xi)$, TTSA degenerates into stochastic approximation (SA) as a special case, which itself has a wide range of applications, including, but not limited to, stochastic optimization \citep{bottou2018optimization,gower2019sgd}, reinforcement learning (RL) \citep{srikant2019finite,dalal2020tale,patil2023finite}, and adaptive Markov chain Monte Carlo (MCMC) \citep{benaim2012strongly,avrachenkov2021dynamic,doshi2023self}. In this paper, our primary focus is the analysis of the asymptotic behavior exhibited by a general nonlinear TTSA with Markovian noise, establishing a central limit theorem (CLT) to explore the effect of coupled variables $(\rvx,\rvy)$. By leveraging this CLT, we address two applications: improvement of asymptotic performance in optimization algorithms, and the derivation of statistical property from a family of gradient-based TD (GTD) algorithms in RL.

\begin{table*}[ht]
    \caption{Overview of TTSA literature. \textbf{Loc. Lipschitz}: locally Lipschitz; \textbf{high-prob. bound}: high-probability bound; \textbf{Mart. diff.}: Martingale difference noise; \textbf{exo. MC}: exogenous Markov chain, independent of TTSA iterates $(\rvx,\rvy)$; \textbf{ctrl. MC}: controlled Markov chain, where the transition kernel is determined by iterates $(\rvx,\rvy)$. Except for a.s. convergence, all other result types inherently include a.s. convergence.}
    \label{tab:related_works_TTSA}
    \begin{center}
    \vspace{-0mm}
    \begin{tabular}{|c||c|c|c|c|}
        \hline
        \textbf{Existing Works} & \textbf{Result Type} & \textbf{Noise Type} & \textbf{Loc. Lipschitz} & \textbf{Nonlinear} \\
        \hhline{|=||=|=|=|=|}
        \cite{konda2004convergence} & CLT  & Mart. diff. & $\times$ & $\times$\\
        \hline
        \cite{mokkadem2006convergence}  & CLT & Mart. diff. & \checkmark & \checkmark \\
        \hline
        \cite{dalal2018finite} & high-prob. bound & Mart. diff. & $\times$ & $\times$ \\
        \hline
        \cite{borkar2018concentration} & high-prob. bound  & Mart. diff. & $\times$  & \checkmark\\
        \hline
        \cite{doan2022nonlinear,doan2024fast,hong2023two} & finite-time bound  & Mart. diff.  & $\times$  & \checkmark \\
        \hline
        \cite{karmakar2018two} & a.s. convergence & ctrl. MC & $\times$ & \checkmark \\
        \hline
        \cite{yaji2020stochastic} & a.s. convergence & ctrl. MC & \checkmark & \checkmark \\
        \hline
\cite{gupta2019finite,haque2023tight} & finite-time bound  & exo. MC & $\times$ & $\times$ \\
        \hline 
        \cite{doan2021finite} & finite-time bound  & exo. MC & $\times$ & \checkmark \\
        \hline 
        \cite{khodadadian2022finite} & finite-time bound  & ctrl. MC & $\times$ & $\times$ \\ 
        \hline
        \cite{barakat2022analysis} & finite-time bound  & ctrl. MC & $\times$ & $\times$ \\   
        \hline
        \cite{zeng2021two} & finite-time bound  & ctrl. MC & $\times$ & \checkmark \\
        \hline
        Our Work & CLT  & ctrl. MC & \checkmark & \checkmark \\
        \hline
    \end{tabular}
    \end{center}
    \vspace{-0mm}
\end{table*}

The recursion of the TTSA algorithm considered in this work is described as follows:
\begin{equation}\label{eqn:general_two_timescale_SA}
    \begin{cases}
        \rvx_{n+1} = \rvx_n + \beta_{n+1}h_1(\rvx_n, \rvy_n, \xi_{n+1}), \\
        \rvy_{n+1} = \rvy_n + \gamma_{n+1} h_2(\rvx_n, \rvy_n, \xi_{n+1}), 
    \end{cases} 
\end{equation}
where $\beta_{n},\gamma_{n}$ are decreasing step sizes at different rates,\footnote{For example, when the step size $\beta_n$ is much smaller than $\gamma_n$, i.e., $\beta_n = o(\gamma_n)$, iterates $\rvx_n$ converges slower than iterates $\rvy_n$, thereby $\rvx_n$ is on the slow timescale and $\rvy_n$ is on the fast timescale.} $\{\xi_n\}_{n\geq 0}$ is a random sequence over a finite set $\Xi$. For instance, in stochastic bilevel optimizations, \citet{hong2023two} deploys TTSA to simultaneously optimize both primal and dual variables. Likewise, the work by \citet{lin2020gradient} highlights the applicability of TTSA in solving minimax problems for optimizing two competing objectives. In RL, a family of GTD algorithms utilize the two-timescale structure \citep{sutton2009fast,dalal2018finite,dalal2020tale,li2023sharp}. Specifically, in these algorithms, the primary value function estimates update on slower timescale, while auxiliary variables or correction terms update on faster timescale. Furthermore, modern energy systems, such as power systems and smart grids, use TTSA for dynamic decision making \citep{lopez2017two,yang2019two}. In the realm of game theory, a noteworthy application is Generative Adversarial Networks, where the game between a generator and a discriminator can be tackled using TTSA \citep{prasad2015two,heusel2017gans}.

In this paper, we focus on the Markovian sequence $\{\xi_n\}$, which plays an important role in the TTSA algorithm and is inherent in many applications.\footnote{While the noise sequence $\{\xi_n\}$ is common to both recursions in the TTSA algorithm \eqref{eqn:general_two_timescale_SA}, it allows for two distinct Markov chains for each recursion. Further details can be found in Section \ref{section:assumptions}.} In distributed learning, token algorithms utilize a random walk, enabling tokens to traverse distributed agents over a graph, each possessing local datasets, and iteratively update model parameters, thus facilitating collaborative stochastic optimization across agents \citep{hu2022efficiency,triastcyn2022decentralized,hendrikx2023principled,even2023stochastic}. Apart from employing SGD iterates to minimize an objective function, such token algorithms of the form \eqref{eqn:general_two_timescale_SA} can also address distributed bilevel or minimax problems that have been recently studied in \citet{gao2022decentralized,gao2023convergence}. Meanwhile, in RL, the environment itself is modeled as a Markov Decision Process (MDP), which by design incorporates Markovian properties. When an agent interacts with this environment, the trajectory $\{\xi_n\}$ it follows, i.e., a sequence of states, actions, and rewards, is inherently Markovian. Notably, this Markovian sequence can be influenced by the agent's adaptive policy, as seen in actor-critic algorithms, yielding a \textit{controlled} Markov chain dependent on the iterates $(\rvx_n,\rvy_n)$ \citep{karmakar2018two,yaji2020stochastic,zeng2021two}. These examples underscore the importance of the Markovian sequence in the development of both theoretical frameworks and practical implementations of various learning algorithms.

\subsection{Related Works}\label{section:related work} \vspace{-0mm}
\textbf{Finite-time vs Asymptotic Analysis:} The convergence properties of SA have been studied extensively using both asymptotic \citep{kushner2003stochastic,fort2015central,borkar2009stochastic,li2023online} and finite-time \citep{srikant2019finite,karimi2019non,chen2022finite} analyses. While recent trends have shown a preference for non-asymptotic analysis, discussions in \citet[Chapter 1.2]{meyn2022control} point out the often-underestimated significance of asymptotic statistics. This notion is highlighted in \citet{mou2020linear, chen2020explicit, srikant2024rates}, which demonstrate the broader applicability of CLT beyond purely asymptotic contexts.  Specifically, the limiting covariance matrix, central to the CLT, finds its presence in high-probability bounds \citep{mou2020linear}, and in finite-time bounds on mean square error \citep{chen2020explicit} as well as $1$-Wasserstein distance to measure the rate of convergence to normality \citep{srikant2024rates}. Further underscoring its significance, \citet{hu2022efficiency} showcases its accuracy in capturing the rate of convergence compared to the mixing rate of the underlying Markov chain, frequently employed in finite-time analysis \citep{karimi2019non,chen2022finite}.

\textbf{TTSA with Martingale Difference Noise:} For the TTSA algorithm \eqref{eqn:general_two_timescale_SA}, the stochastic sequence being an \textit{i.i.d.} sequence $\{\xi_n\}$ allows for the decomposition of the noisy observation $h_1(\rvx,\rvy,\xi_n)$ into $\bar{h}_1(\rvx,\rvy)$ and a Martingale difference noise term $h_1(\rvx,\rvy,\xi_n) \!-\! \bar{h}_1(\rvx,\rvy)$; a similar decomposition applying to $h_2$. In the case of Martingale difference noise, an extensive body of research focuses on the analysis of CLT results \citep{konda2004convergence,mokkadem2006convergence}, high-probability bounds \citep{dalal2018finite,borkar2018concentration}, and finite-time bounds \citep{doan2022nonlinear,doan2024fast,hong2023two} for both linear and nonlinear TTSA, as shown in Table \ref{tab:related_works_TTSA}.

\textbf{TTSA with Markovian Noise --- Asymptotic Results and Suboptimal Finite-Time Bounds:} Recently, increasing attention has been shifted towards analyzing TTSA with Markovian noise sequences $\{\xi_n\}$, which introduces technical challenges due to inherent bias in $h_i(\rvx,\rvy,\xi_n)$ as an estimator of $\bar{h}_i(\rvx,\rvy)$ for $i = 1, 2$. \citet{karmakar2018two,yaji2020stochastic} delve into the almost sure convergence of nonlinear TTSA with Markovian noise, showing that the two iterates $\rvx_n, \rvy_n$ asymptotically estimate the related differential inclusions, which are a generalized version of ordinary differential equations (ODEs). \citet{yaji2020stochastic} further relax to the locally Lipschitz functions $h_1, h_2$, which is commonly seen in the machine learning literature such as low-rank matrix recovery \citep{recht2010guaranteed}, tensor factorization problem \citep{kolda2009tensor}, and deep neural networks with unbounded Hessian matrices \citep[Appendix H]{zhang2020first}. 

Meanwhile, the mixing rate properties of Markov chains have been predominantly utilized for the finite-time analysis of both linear \citep{gupta2019finite,kaledin2020finite,doan2021finitelinearTTSA,khodadadian2022finite,barakat2022analysis,haque2023tight} and nonlinear TTSA \citep{doan2021finite,zeng2021two} with Markovian noise.\footnote{While nonlinear TTSA with Markovian noise is currently the most general framework, our emphasis is not solely on generalization. As we will demonstrate in Section \ref{section:3}, this setting has substantive implications in both stochastic optimization and RL.} Notably, the latter two works align closely with our TTSA settings. However, \citet{doan2021finite} only provided a finite-time bound for the combined error of both iterations, i.e., $\E[\|\rvx_n-\rvx^*\|^2+\frac{\beta_n}{\gamma_n}\|\rvy_n-\rvy^*\|^2]$ at a suboptimal rate of $O(n^{-2/3})$ with a specific choice of step sizes $\beta_n = (n+1)^{-1}$ and $\gamma_n = (n+1)^{-2/3}$, while we show in Section \ref{section:main_results} that for large $n$, the combined error should approximately decrease to zero at the speed of $O(\beta_n) = O(n^{-1})$. A similar bound for $\E[\|\rvx_n - \rvx^*\|^2]$ under the more general controlled Markov noise setting is provided in \citet{zeng2021two} at the suboptimal rate of $O(n^{-2/3})$ with the same choice of step sizes. Thus, even the state-of-the-art finite-time bounds in \citet{doan2021finite,zeng2021two} do not preciously capture the leading term that determines the performance of each iterates $\rvx_n, \rvy_n$. A comprehensive non-asymptotic analysis with rate matching the CLT scale (i.e., $O(\beta_n), O(\gamma_n)$) has yet to be performed in the nonlinear TTSA with controlled Markovian noise under general decreasing step sizes $\beta_n, \gamma_n$. 

\subsection{Our Contributions}
In this paper, we study the CLT of both iterates $\rvx_n$ and $\rvy_n$ in nonlinear TTSA with controlled Markovian noise, where $h_1, h_2$ are only locally Lipschitz continuous. Although \citet{yaji2020stochastic} considered more general set-valued functions $h_1, h_2$, they only obtained almost sure convergence. In contrast, we here target single-valued functions that are more common in the machine learning literature and extend the scope to include CLT results. Our work further generalizes the CLT analysis of the two-timescale framework in \citet{mokkadem2006convergence} - still a state-of-the-art CLT result for Martingale difference noise - by necessitating a deeper exploration into the Markovian noise $\{\xi_n\}_{n\geq 0}$, given that $h_i(\rvx,\rvy,\xi_n) - \bar{h}_i(\rvx,\rvy), i= 1, 2$ are no longer Martingale difference. 

Utilizing our CLT results, we demonstrate the impact of sampling strategies on the limiting covariance across a wide class of distributed optimization algorithms. Extending beyond the vanilla SGD setting studied in \citet{hu2022efficiency}, we show that improved sampling strategies lead to better performance for general TTSA including, but not restricted to, SGD variants and algorithms tailored for stochastic bilevel and minimax problems. Moreover, in the RL context, we introduce first of its kind statistical characterization of GTD2 and TDC algorithms with nonlinear function approximation \citep{maei2009convergent} using Markovian samples. Using both theoretical and empirical results, we show that their asymptotic performance coincides, as evidenced by identical covariance matrix in our CLT. Such conclusions are not possible via current finite-time bounds \citep{doan2021finite,zeng2021two}.

\textbf{Notations.}  We use $\|\!\cdot\!\|$ to denote both the Euclidean norm of vectors and the spectral norm of matrices. Two symmetric matrices $\rmM_1, \rmM_2$ follow Loewner ordering $\rmM_1 >_L \rmM_2$ (resp. `$\geq_L$') if $\rmM_1 -\rmM_2$ is positive definite (resp. positive semi-definite). A matrix is \textit{Hurwitz} if all its eigenvalues possess strictly negative real parts. The function $\mathds{1}_{(\cdot)}$ is an indicator function. $\nabla_{\rvx} h(\rvx,\rvy)$ stands for the Jacobian matrix of the vector-valued function $h(\rvx,\rvy)$ with respect to the variable $\rvx$. $C^1$ function $f$ means that function $f$ is both continuous and differentiable. We use `$\xrightarrow{~d~}$' for the convergence in distribution and $N(0,\rmV)$ is the Gaussian random vector with covariance matrix $\rmV$.
\section{MAIN RESULTS}\label{section:2}
In this section, we analyze the asymptotic behavior of the TTSA algorithm \eqref{eqn:general_two_timescale_SA} with Markovian noise. First, we provide assumptions and the almost sure convergence result in Section \ref{section:assumptions}. Before presenting our main CLT result in Section \ref{section:main_results}, we explain how our result is achieved by transforming the TTSA iteration into a single-timescale SA-like recursion, and introduce some key components related to asymptotic covariance of the iterates. This transformation resembles that in \citet{konda2004convergence, mokkadem2006convergence} but with a fresh perspective by accounting for biased errors due to Markovian noise, as elaborated upon in Section \ref{section:insights}.

\subsection{Key Assumptions and a.s. Convergence}\label{section:assumptions}
\begin{enumerate}[label=A\arabic*., ref=(A\arabic*)]
  \item The step sizes $\beta_n \triangleq (n+1)^{-b}$ and $\gamma_n \triangleq (n+1)^{-a}$, where $0.5 < a < b \leq 1$. \label{assump:one} \vspace{-0mm}
  \item For the $C^1$ function $h_1 : \sR^{d_1} \times \sR^{d_2} \times \Xi \to \sR^{d_1}$, there exists a positive constant $L_1$ such that $\|h_1(\rvx,\rvy,\xi)\| \leq L_1(1 + \|\rvx\| + \|\rvy\|)$ for every $\rvx \in \sR^{d_1}, \rvy \in \sR^{d_2}, \xi \in \Xi$. The same condition holds for the $C^1$ function $h_2$ as well. \label{assump:two} \vspace{-0mm}
  \item  Consider a $C^1$ function $\lambda : \sR^{d_1} \to \sR^{d_2}$. For every $\rvx \in \sR^{d_1}$, the following three properties hold: (i) $\lambda(\rvx)$ is the globally attracting point of the related ODE $\dot \rvy = \bar h_2(\rvx,\rvy)$; (ii) $\nabla_{\rvy}\bar{h}_2(\rvx,\lambda(\rvx))$ is Hurwitz; (iii) $\|\lambda(\rvx)\| \leq L_2(1+\|\rvx\|)$ for some positive constant $L_2$. Additionally, let $\hat{h}_1(\rvx) \triangleq \bar{h}_1(\rvx,\lambda(\rvx))$, there exists a set of disjoint roots $\Lambda \triangleq \{\rvx^*: \hat{h}_1(\rvx^*)  = 0, \nabla_{\rvx} \hat{h}_1(\rvx^*) + \frac{\mathds{1}_{\{b=1\}}}{2}\rmI \text{ is Hurwitz}\}$, which is also the globally attracting set for trajectories of the related ODE $\dot \rvx = \hat h_1(\rvx)$. \label{assump:three} \vspace{-0mm}
  \item $\{\xi_n\}_{n\geq 0}$ is an iterate-dependent Markov chain on finite state space $\Xi$. For every $n\geq 0$, $P(\xi_{n+1} = j | \rvx_m,\rvy_m, \xi_m, 0 \leq m \leq n) = P(\xi_{n+1} = j | \rvx_n,\rvy_n, \xi_n=i) = \rmP_{i,j}[\rvx_n,\rvy_n]$, where the transition kernel $\rmP[\rvx,\rvy]$ is continuous in $\rvx, \rvy$, and the Markov chain generated by $\rmP[\rvx,\rvy]$ is ergodic so that it admits a stationary distribution $\vpi(\rvx,\rvy)$, and $\vpi(\rvx^*,\lambda(\rvx^*)) = \vmu$. \label{assump:four} \vspace{-0mm}
  \item $\sup_{n\geq 0} (\|\rvx_n\| + \|\rvy_n\|) < \infty$ a.s. \label{assump:five}\vspace{-0mm}
\end{enumerate}
In Assumption \ref{assump:one}, the step sizes $\beta_n,\gamma_n$ decay polynomially at distinct rates, i.e., $\beta_n = o(\gamma_n)$, which is standard in the TTSA literature \citep{zeng2021two,doan2021finite,hong2023two}. Assumption \ref{assump:two} ensures that $C^1$ functions $h_1, h_2$ are locally Lipschitz and grow at most linearly with respect to the norms of their parameters, as also assumed in \citet{yaji2020stochastic}. This is a far less stringent condition compared to the globally Lipschitz assumption used in most of the recent works, as listed in \cref{tab:related_works_TTSA}.

Assumption \ref{assump:three} is crucial for the analysis of iterates $(\rvx_n,\rvy_n)$, which can be seen as a stochastic discretization of the ODEs $\dot{\rvx} = \hat{h}_1(\rvx)$ and $\dot{\rvy} = \bar{h}_2(\rvx,\rvy)$. This assumption guarantees the global asymptotic stability of these two ODEs, as demonstrated in \citet{yaji2020stochastic,doan2021finite}. The linear growth of $\lambda(\rvx)$ is a milder condition than the globally Lipschitz assumption in \citet{borkar2018concentration,karmakar2018two,zeng2021two,doan2021finite}. 

Assumption \ref{assump:four} is standard to guarantee the asymptotic unbiasedness of $h_1, h_2$ in the existing literature on TTSA with Markovian noise \citep{karmakar2018two,yaji2020stochastic,khodadadian2022finite,barakat2022analysis}. It is worth noting that $\{\xi_n\}$ naturally allows for an augmentation of the form $\xi_n \triangleq (X_n, Y_n)$, with two independent Markovian noise sequences $\{X_n\}$, $\{Y_n\}$ corresponding to iterates $\{\rvx_n\}$ and $\{\rvy_n\}$, respectively. In this case, the functions $h_1$ and $h_2$ act only on the entries of $\xi_n$ related to $X_n$ and $Y_n$.

Assumption \ref{assump:five} assumes the a.s. boundedness of the coupled iterates $(\rvx_n,\rvy_n)$, which is commonly seen in the TTSA literature \citep{karmakar2018two,yaji2020stochastic}. A similar stability condition is also found in the SA literature \citep{delyon1999convergence,borkar2009stochastic,li2023online}. In practice, to stabilize the TTSA algorithm \eqref{eqn:general_two_timescale_SA} under Markovian noise, one could adopt algorithmic modifications from the SA literature, including the projection method onto (possibly expanding) compact sets \citep{chen2006stochastic,andrieu2014markovian} or the truncation method with a restart process \citep{fort2015central,fort2016convergence}.

\begin{lemma}[Almost Sure Convergence]\label{lemma:almost_sure_convergence}
    Under Assumptions \ref{assump:one} - \ref{assump:five}, iterates $(\rvx_n,\rvy_n)$ in \eqref{eqn:general_two_timescale_SA} almost surely converge to a set of roots, i.e., $(\rvx_n, \rvy_n) \to \bigcup_{\rvx^* \in \Lambda}      (\rvx^*, \lambda(\rvx^*))$ a.s. \vspace{-0mm}
\end{lemma}
Lemma \ref{lemma:almost_sure_convergence} follows from \citet[Theorem 4]{yaji2020stochastic} by verifying the conditions therein and we defer the details to Appendix $A.1$. While they studied broader set-valued functions $h_1, h_2$ within the realm of stochastic recursive inclusion, they did not explore the CLT result. This is likely due to existing gaps in the CLT analysis even for single-timescale stochastic recursive inclusion, as mentioned in \citet[Chapter 5]{borkar2009stochastic}. In contrast, we focus on single-valued functions $h_1, h_2$, as prevalent in the machine learning literature. This paves the way for the first CLT result, Theorem \ref{theorem:CLT_general_TTSA}, for the general TTSA with \textit{controlled} Markovian noise, as demonstrated in \cref{tab:related_works_TTSA}. In the following section, we will conduct a more detailed analysis of the asymptotic behavior of iterates $(\rvx_n,\rvy_n)$ near equilibrium $(\rvx^*, \lambda(\rvx^*))$ for some $\rvx^* \in \Lambda$.

\subsection{Overview of the CLT Analysis for $(\rvx_n, \rvy_n)$} \label{section:insights}
Assumption \ref{assump:one} puts $\{\rvy_n\}_{n \geq 0}$ on a `faster timescale' compared to $\{\rvx_n\}_{n \geq 0}$, and has implications on convergence rates of the two sequences. Under additional conditions on the function $\bar h_2(\cdot,\cdot)$ in Assumption \ref{assump:three}, the sequence $\{\rvy_n\}$ can be approximated by $\{\lambda(\rvx_n)\}$ for large time step $n$, where $\lambda(\rvx)$ is an implicit function solving $\bar h_2 (\rvx,\lambda(\rvx)) = 0$. Loosely speaking, when $n$ is large enough, the fast iterates $\rvy_n$ are nearly convergent to the root $\lambda(\rvx_n)$ of $\bar h_2(\rvx_n,\cdot)$. Iterates $\rvx_n$ on the slower timescale then guide the roots $\lambda(\rvx_n)$ of the iterates $\rvy_n$ until they reach $\rvy^* = \lambda(\rvx^*)$, which also satisfies $\bar h_1 (\rvx^*,\lambda(\rvx^*)) = 0$.
Consequently, resembling \citet[Section 2]{konda2004convergence} and \citet[Section 2.3]{mokkadem2006convergence}, we can show that $\{\rvx_n\}$ is now approximated by iterating a \textit{single-timescale} SA update rule, independent of $\{\rvy_n\}$ but instead driven by $\{\lambda(\rvx_n)\}$, whose derivation we detail in what follows. For $i \in \{1,2\}$, define \vspace{-0mm}
\begin{align*}
&\rmQ_{i1} \triangleq \nabla_{\rvx}\bar{h}_i(\rvx,\rvy) \big|_{(\rvx,\rvy) = (\rvx^*\!,~\! \rvy^*),} \\
&\rmQ_{i2} \triangleq \nabla_{\rvy}\bar{h}_i(\rvx,\rvy) \big|_{(\rvx,\rvy) = (\rvx^*\!,~\! \rvy^*),}  \\
&\Delta_{n}^{(i)} \triangleq h_i(\rvx_n, \rvy_n, \xi_{n+1}) - \bar{h}_i(\rvx_n,\rvy_n), \\
&\tilde \Delta_{n}^{(i)} \triangleq h_i(\rvx_n, \lambda(\rvx_n), \xi_{n+1}) - \bar{h}_i(\rvx_n,\lambda(\rvx_n)). \vspace{-0mm}
\end{align*}
Adding and subtracting $\bar{h}_1(\rvx_n,\rvy_n)$ and $\bar{h}_2(\rvx_n,\rvy_n)$ to iterates $\rvx_n$ and $\rvy_n$ in \eqref{eqn:general_two_timescale_SA} respectively, and taking their Taylor expansions at $(\rvx_n,\rvy_n) \!=\! (\rvx^*, \rvy^*)$, gives us \vspace{-0mm}
\begin{equation}\label{eqn:iteration_x_2}
\begin{split}
    \rvx_{n+1}&  \!=\rvx_{n} \!+\! \beta_{n+1}(\rmQ_{11}(\rvx_n\!-\!\rvx^*) + \rmQ_{12} (\rvy_n\!-\!\rvy^*)) \\
    \!+\! &~\beta_{n+1}\Delta_{n}^{(1)} \!\!+\! \beta_{n+1} O(\|\rvx_n\!\!-\! \rvx^*\|^2 \!+\! \|\rvy_n\!\!-\! \rvy^*\|^2),
\end{split}
\end{equation}
\begin{equation}\label{eqn:iteration_y_2}
\begin{split}
    \rvy_{n+1}& \!= \rvy_{n} \!+\! \gamma_{n+1}(\rmQ_{21} (\rvx_n \!-\! \rvx^*) + \rmQ_{22}(\rvy_n \!-\! \rvy^*)) \\
    \!+\! &~\gamma_{n+1}\Delta_{n}^{(2)}\!\!+\! \gamma_{n+1} O(\|\rvx_n \!\!-\! \rvx^*\|^2 + \|\rvy_n \!\!-\! \rvy^*\|^2).
\end{split}
\end{equation}
Re-arranging \eqref{eqn:iteration_y_2} by placing the $\rvy_n-\rvy^*$ on the left-hand side yields \vspace{-0mm}
\begin{equation}\label{eqn:y_n-y^*}
\begin{split}
    \rvy_n\!-\!\rvy^* &\!\!=\! \gamma_{n+1}^{\!-1}\rmQ_{22}^{-1}(\rvy_{n+\!1}\!-\!\rvy_n) \!-\! \rmQ_{22}^{\!-1}\rmQ_{21}(\rvx_n \!-\! \rvx^*) \\
    &+ \rmQ_{22}^{-1}\Delta_n^{(2)} \!+\! O(\|\rvx_n \!\!-\! \rvx^*\|^2 + \|\rvy_n \!\!-\! \rvy^*\|^2). \vspace{-0mm}
\end{split}
\end{equation}
By substituting the above into \eqref{eqn:iteration_x_2}, and then replacing (approximating) $\rvy_n$ with $\lambda(\rvx_n)$, we get \vspace{-0mm}
\begin{equation}\label{eqn:iteration_x_3} 
\begin{split}
    \rvx_{n+\!1} \!=\! \rvx_n \!+\! \beta_{n+\!1}\rmK_{\rvx}(\rvx_n\!\!-\!\rvx^*) \!+\! \beta_{n+\!1}\tilde\Delta_n^{\rvx} \!+\! \beta_{n+\!1} R_n,
\end{split} \vspace{-0mm}
\end{equation}
where  \vspace{-0mm}
\begin{equation}\label{eqn:K_x}
\rmK_{\rvx} \!\triangleq\! \rmQ_{11} \!-\!\rmQ_{12}\rmQ_{22}^{-1}\rmQ_{21}, ~~\tilde\Delta_n^{\rvx} \!\triangleq\! \tilde{\Delta}_n^{(1)} \!-\! \rmQ_{12}\rmQ_{22}^{-1} \tilde{\Delta}_n^{(2)}\!\!, \vspace{-0mm}
\end{equation}
and $R_n$ is comprised of residual errors from the earlier Taylor expansion and approximation of $\rvy_n$ by $\lambda(\rvx_n)$. The term $\tilde\Delta_n^{\rvx}$ can be further decomposed using the Poisson equation technique \citep{benveniste2012adaptive,meyn2022control} as \vspace{-0mm}
\begin{align*}
&\tilde\Delta_n^{\rvx} \!=\! [M_{n+1}^{(1)} \!-\! \rmQ_{12}\rmQ_{22}^{-1}M_{n+1}^{(2)}] \\
&~~~~~~~+ [\tilde{H}(\rvx_n,\lambda(\rvx_n),\xi_{n+1}) - \tilde{H}(\rvx_n,\lambda(\rvx_n),\xi_{n})],
\end{align*}
where $M_{n+1}^{(1)}$ and $M_{n+1}^{(2)}$ are Martingale difference terms adapted to filtration $\gF_n \triangleq \sigma(\rvx_0,\rvy_0,\xi_0,\cdots,\xi_n)$. The exact expressions for the Martingale difference terms can be found in Appendix $A.2.1$, equations 9(a) and 9(b). The second summand including the $\tilde{H}$ terms, whose exact expression is provided in Appendix $A.2.1$, involves consecutive Markovian noise terms $ \xi_{n+1}$ and $\xi_n$ which are responsible for biased errors in the iteration for $\rvx_n$. These additional terms are not present in existing works that focus only on \textit{i.i.d.} stochastic inputs \citep{konda2004convergence,mokkadem2006convergence}, even though their analysis leads to equations similar to \eqref{eqn:iteration_x_3}. In Appendix $A.2.2$, we show that the $\tilde H$ terms along with residual errors $R_n$ at each step are $o(\sqrt{\beta_n})$, and thus do not influence the CLT result for iterates $\rvx_n$ of the slower timescale.

Consequently, the approximation $\rvy_n = \lambda(\rvx_n)$, together with the aforementioned analysis leading to $o(\sqrt{\beta_n})$, now allows us to analyze \eqref{eqn:iteration_x_3} as essentially a single-timescale SA with Markovian noise. We then apply \citet[Proposition 4.1]{fort2015central} to extract a CLT result, i.e., we prove that \vspace{-0mm}
\begin{equation}\label{eqn:CLT_single_timescale_SA}
    \beta_{n}^{-1/2}(\rvx_n - \rvx^*) \xrightarrow{d} N(0, \rmV_{\rvx}), \vspace{-0mm}
\end{equation}
where $\rmV_{\rvx}$ solves the Lyapunov equation $\rmU_{\rvx} + (\rmK_{\rvx}+ \frac{\mathds{1}_{\{\beta_n = O(1/n)\}}}{2}\rmI)\rmV_{\rvx} + \rmV_{\rvx}(\rmK_{\rvx}+ \frac{\mathds{1}_{\{\beta_n = O(1/n)\}}}{2}\rmI)^T = 0$, \vspace{-0mm}
 \begin{equation}\label{eqn:def_K_U}
    \rmU_{\rvx} \triangleq \lim_{s \to \infty}\frac{1}{s} \E\left[\left(\sum_{n=1}^s\tilde\Delta_n^{\rvx^*}\right)\left(\sum_{n=1}^s\tilde\Delta_n^{\rvx^*}\right)^T\right], \vspace{-0mm}
\end{equation}   
and $\tilde\Delta_n^{\rvx^*}$ represents $\tilde\Delta_n^{\rvx}$ measured at $\rvx_n = \rvx^*$ for all $n$. Through $\rmQ_{22}$, $\rmQ_{21}$ and $\tilde \Delta_n^{(2)}$, both $\rmK_{\rvx}$ and $\rmU_{\rvx}$ capture the effect of deterministic field $\bar h_2(\cdot,\cdot)$ on the asymptotic behavior of $\rvx_n$. The matrix $\rmU_{\rvx}$ incorporates the effect of Markovian noise sequence $\{\xi_n\}_{n \geq 0}$ through $\tilde \Delta_n^{(1)}$ and $\tilde \Delta_n^{(2)}$, which will be utilized in Proposition \ref{proposition:performance_ordering} to identify the effect of the underlying Markov chain on the asymptotic behavior of iterates $\rvx_n$. 
We show in Appendix $A.2.1$ that $\rmU_{\rvx}$ can also be written as \vspace{-0mm}
\begin{equation}\label{eqn:alt_form_of_matrix_U_x}
    \rmU_{\rvx} \!=\! \begin{bmatrix}
        \rmI & -\rmQ_{12}\rmQ_{22}^{-1}
    \end{bmatrix} \begin{bmatrix}
        \rmU_{11} & \rmU_{12} \\ \rmU_{21} & \rmU_{22}
    \end{bmatrix} \begin{bmatrix}
        \rmI & -\rmQ_{12}\rmQ_{22}^{-1}
    \end{bmatrix}^T\!\!\!, \vspace{-0mm}
\end{equation}
where   \vspace{-0mm}
\begin{equation}\label{eqn:U_ij}
    \rmU_{ij} \triangleq \lim_{s\to\infty} \frac{1}{s}\E\left[\left(\sum_{n=1}^{s} \tilde{\Delta}_n^{(i)^{\dagger}}\right)\left(\sum_{n=1}^{s} \tilde{\Delta}_n^{(j)^{\dagger}}\right)^T\right], \vspace{-0mm}
\end{equation}
$\tilde{\Delta}_n^{(i)^{\dagger}}$ denotes $\tilde{\Delta}_n^{(i)}$ measured at $\rvx_n = \rvx^*$ for all $n$ and $i$, and $\rmU_{12} = \rmU_{21}^T$. For an \textit{i.i.d.} sequence $\{\xi_n\}$ with marginal $\vmu$, $\rmU_{ij} = \E_{\xi \sim \vmu}[h_i(\rvx^*,\rvy^*,\xi)h_j(\rvx^*,\rvy^*,\xi)^T]$ degenerates to the marginal covariance of functions $h_i(\rvx^*,\rvy^*,\cdot), h_j(\rvx^*,\rvy^*,\cdot)$, and \eqref{eqn:CLT_single_timescale_SA} aligns with previously established CLT results for linear \citep{konda2004convergence} and nonlinear TTSA \citep{mokkadem2006convergence}, both with Martingale difference noise.

\subsection{Central Limit Theorem of TTSA with Controlled Markovian Noise}\label{section:main_results}
Without loss of generality, our remaining results are stated while conditioning on the event that $\{\rvx_n \to \rvx^*, \rvy_n \to \rvy^*\}$, for some $\rvx^* \in \Lambda$ and $\rvy^* = \lambda(\rvx^*)$. Our main CLT result is as follows, with its proof deferred to Appendix $A.2$.
\begin{theorem}[Central Limit Theorem]\label{theorem:CLT_general_TTSA}
    Under Assumptions \ref{assump:one} -- \ref{assump:five},
        \begin{equation}\label{eqn:weak_convergence_general_TTSA}
    \begin{pmatrix}
       \beta_n^{\!-1/2} (\rvx_n - \rvx^*) \\\gamma_n^{\!-1/2} (\rvy_n - \rvy^*)
    \end{pmatrix} \!\xrightarrow{~d~} N\left(\vzero, \begin{pmatrix}
        \rmV_{\rvx} & \vzero \\ \vzero & \rmV_{\rvy}
    \end{pmatrix}\right), 
\end{equation}
where the limiting covariance matrices $\rmV_{\rvx} \in \sR^{d_1 \times d_1}, \rmV_{\rvy} \in \sR^{d_2 \times d_2}$ are given by
\begin{equation}\label{eqn:limiting_cov_matrix}
\begin{split}
    &\rmV_{\rvx} \!=\! \int_0^{\infty}\! e^{t\left(\rmK_{\rvx}+\frac{\mathds{1}_{\{b=1\}}}{2}\rmI\right)} \rmU_{\rvx} e^{t\left(\rmK_{\rvx}+\frac{\mathds{1}_{\{b=1\}}}{2}\rmI\right)^T} dt,\\
    &\rmV_{\rvy} \!=\! \int_0^{\infty} e^{t\rmQ_{22}} \rmU_{22} e^{t\rmQ_{22}^T} dt,
\end{split}
\end{equation} 
with $\rmK_{\rvx}$, $\rmU_{\rvx}$ and $\rmU_{22}$ defined in \eqref{eqn:K_x}, \eqref{eqn:def_K_U} and \eqref{eqn:U_ij}, respectively. 
\end{theorem}
Theorem \ref{theorem:CLT_general_TTSA} suggests that iterates $(\rvx_n, \rvy_n)$ evolve asymptotically independently, as evidenced by the zero covariance of off-diagonal terms in \eqref{eqn:weak_convergence_general_TTSA}. This is due to the diminishing correlation between $(\rvx_n-\rvx^*)$ and $(\rvy_n-\rvy^*)$ at a rate of $O(\beta_n/\gamma_n)$, a characteristic of the two-timescale setup, aligning with existing CLT findings for TTSA with Martingale difference noise \citep{konda2004convergence,mokkadem2006convergence}. The limiting covariance matrix $\rmV_{\rvy}$ is solely determined by the local function $h_2$ and $\rvx^*$ without an additional term $\frac{\mathds{1}_{\{a=1\}}}{2}\rmI$ due to $a\!<\!1$ by assumption \ref{assump:one}, implying minimal effect of $\rvx_n$ on the asymptotic behavior of $\rvy_n$. In contrast, $\rmV_{\rvx}$ is significantly impacted by iterates $\rvy_n$ since matrices $\rmK_{\rvx}, \rmU_{\rvx}$ are comprised of functions $h_2$ and $\bar{h}_2$.

As a special case, when $h_1(\rvx,\rvy,\xi)$ in the TTSA algorithm is independent of the variable $\rvy$, i.e., $h_1(\rvx,\rvy,\xi) \equiv h_1(\rvx,\xi)$, then $\nabla_{\rvy} h_1(\rvx,\xi) = 0$ for any $\rvy \in \sR^{d_2}$, implying $\rmQ_{12} = \vzero$. According to Theorem \ref{theorem:CLT_general_TTSA}, $\rvx_n$ is decoupled from iterates $\rvy_n$ and reduces to the single-timescale SA with Markovian noise, where $\rmV_{\rvx}$ in \eqref{eqn:limiting_cov_matrix}, in view of \eqref{eqn:alt_form_of_matrix_U_x} with $\rmQ_{12} = \vzero$, becomes
\begin{equation*}
    \rmV_{\rvx} \!=\! \int_0^{\infty}\!\! e^{t\left(\!\!\nabla \bar{h}_1(\rvx^*)+\frac{\mathds{1}_{\{b=1\}}}{2}\rmI\!\right)} \rmU_{11} e^{t\left(\!\!\nabla \bar{h}_1(\rvx^*)+\frac{\mathds{1}_{\{b=1\}}}{2}\rmI\!\right)^T}\!\! dt. \vspace{-0mm}
\end{equation*}
This $\rmV_{\rvx}$ is in line with the existing CLT result for the single-timescale SA with controlled Markovian noise \citep{delyon2000stochastic,benveniste2012adaptive,fort2015central} under the same locally Lipschitz condition on $h_1(\rvx,\xi)$, as stated in Assumption \ref{assump:two}. 

The limiting covariance matrices $\rmV_{\rvx}$ and $\rmV_{\rvy}$ are related to the mean square error (MSE) of their corresponding iterative errors $\rvx_n - \rvx^*$ and $\rvy_n - \rvy^*$. For large enough $n$, the diagonal entries of $\rmV_{\rvx}$ are approximated by $\rve_i^T \rmV_{\rvx}\rve_i \!\approx\! \rve_i^T\E[(\rvx_n \!-\! \rvx^*)(\rvx_n \!-\! \rvx^*)^T]\rve_i/\beta_n$ for all $i \in \{1,\cdots,d_1\}$, where $\rve_i$ is the $i$-th canonical vector. Then, the MSE of the iterate error $\rvx_n - \rvx^*$ can be estimated as $\E[\|\rvx_n - \rvx^*\|^2] = \sum_{i=1}^{d_1} \rve_i^T\E[(\rvx_n \!-\! \rvx^*)(\rvx_n \!-\! \rvx^*)^T]\rve_i \approx \beta_n\sum_{i=1}^{d_1} \rve_i^T \rmV_{\rvx}\rve_i$. This implies that $\E[\|\rvx_n - \rvx^*\|^2]$ resembles the trace\footnote{Sum of diagonal entries of a matrix.} of $\rmV_{\rvx}$, and decreases at a rate of $\beta_n$. Similar arguments also hold for $\E[\|\rvy_n - \rvy^*\|^2]$ and $\rmV_{\rvy}$.
\section{APPLICATIONS}\label{section:3}
\subsection{Performance Ordering in TTSA}\label{section:efficiency ordering}
The limiting covariance matrices $\rmV_{\rvx}, \rmV_{\rvy}$ described in \eqref{eqn:limiting_cov_matrix} for nonlinear TTSA with Markovian noise inherently incorporate the properties of the underlying Markov chain \textit{completely} in terms of matrices $\rmU_{\rvx}$ and $\rmU_{22}$, as defined in \eqref{eqn:alt_form_of_matrix_U_x} and \eqref{eqn:U_ij}. This raises an intuitive question: \textit{If we can control the stochastic input sequence $\{\xi_n\}_{n\geq 0}$, how does it influence the performance of the TTSA algorithm?} 

\begin{figure}[t]
    \centering
    \includegraphics[width=0.99\columnwidth]{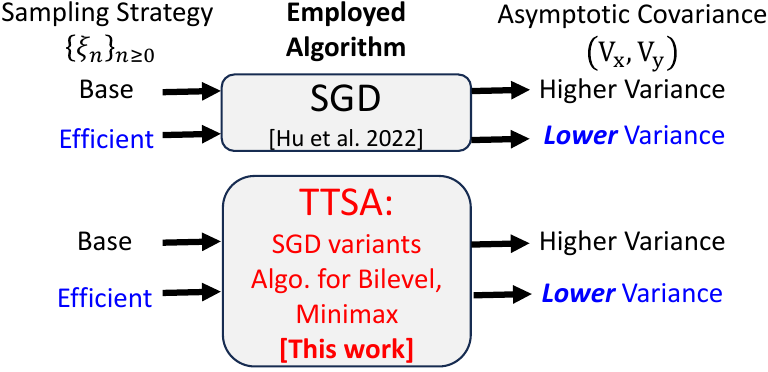}
    \vspace{-3mm}
    \caption{Efficiency Ordering: From SGD to TTSA.}
    \label{fig:diagram_efficiency_ordering}
    \vspace{-0mm}
\end{figure}

This question was originally studied by \citet{hu2022efficiency}, which introduces the notion of efficiency ordering of Markov chains, a metric prevalent in the MCMC literature, in the context of SGD algorithms, and proves that the presence of `better', more efficient sampling strategy leads to improved SGD performance. Broadening this concept, we show that such performance improvements are applicable to the general TTSA framework, beyond mere SGD algorithms, as depicted in Figure \ref{fig:diagram_efficiency_ordering}. To better understand this, let $\rmU^{Z}(g) \triangleq \lim_{s\to\infty} \frac{1}{s}\E[\Delta_s\Delta_s^T]$ be the sampling covariance matrix for a vector-valued function $g:\Xi \!\to\! \sR^d$ and stochastic process $\{Z_n\}$, where $\Delta_s \!=\! \sum_{n=1}^s (g(Z_n) - \E_{\vmu}[g])$ and $\E_{\vmu}[g] \!=\! \sum_{i\in\Xi}g(i)\mu_{i}$.

\begin{definition}[Efficiency Ordering, \citep{mira2001ordering,hu2022efficiency}]\label{def:efficiency_ordering}
    For two Markov chains $\{W_n\}$ and $\{Z_n\}$ with identical stationary distribution $\vmu$, we say $\{Z_n\}$ is more sampling-efficient than $\{W_n\}$, denoted as $W \preceq Z$, if and only if $\rmU^W(g) \geq_L \rmU^Z(g)$ for any vector-valued function $g$.
\end{definition}
Examples of sampling strategies following Definition \ref{def:efficiency_ordering} include random and single shuffling paradigms \citep{ahn2020sgd,safran2020good}, which are shown to be more sampling-efficient when compared to \textit{i.i.d.} sampling. Another example, relevant in the context of token algorithms in distributed learning, is the so-called non-backtracking random walk (NBRW) \citep{alon2007non,lee2012beyond,ben2018comparing}, which is more sampling-efficient than simple random walk (SRW). We point the reader to \citet[Section 4]{hu2022efficiency} for more detailed discussions, where more efficient sampling strategies employed in SGD algorithms lead to reduced asymptotic covariance of iterate errors. With two efficiency-ordered sampling strategies,  we now extend the same performance ordering to TTSA, the proof of which can be found in Appendix $A.3.1$.

\begin{proposition}\label{proposition:performance_ordering}
    For the TTSA algorithm \eqref{eqn:general_two_timescale_SA}, given two different underlying Markov chains $\{W_n\}_{n\geq 0}$ and $\{Z_n\}_{n\geq 0}$ that are efficiency ordered, i.e., $W \preceq Z$, we have $\rmV_{\rvx}^{(W)} \geq_L \rmV_{\rvx}^{(Z)}$ and $\rmV_{\rvy}^{(W)} \geq_L \rmV_{\rvy}^{(Z)}$.
\end{proposition}
Proposition \ref{proposition:performance_ordering} enables us to expand the scope of \citet{hu2022efficiency} by employing sampling-efficient strategies to a wider class of optimization problems within the TTSA framework. Specifically, our scope extends existing results as follows:

(i) \textit{From vanilla SGD to its variants:} The TTSA structure accommodates many SGD variants for finite-sum minimization, including the Polyak-Ruppert averaging \citep{ruppert1988efficient,polyak1992acceleration} and momentum SGD \citep{gadat2018stochastic,li2022revisiting}. Other variants of SGD in the TTSA framework, e.g., signSGD and normalized SGD, are provided in \citet[Section 4.3]{xiao2023convergence} with detailed expressions. 

(ii) \textit{From finite-sum minimization to bilievel and minimax problems:} Many algorithms within the TTSA framework can handle bilevel and minimax problems. For instance, \citet[Algorithm 1]{hong2023two}  effectively deals with both inner and outer objectives in bilevel optimization, while the stochastic gradient descent ascent algorithm \citep[Algorithm 1]{lin2020gradient} seeks saddle points in the minimax problem.

From Proposition \ref{proposition:performance_ordering}, all the above algorithms enjoy improved asymptotic performance when driven by more efficient samples. For instance, in the token algorithm setting \citep{hu2022efficiency,triastcyn2022decentralized,hendrikx2023principled,even2023stochastic}, a token can employ NBRW over SRW to solve various optimization problems with these TTSA algorithms. When random access of each data point is possible, \citet[Lemma 4.2]{hu2022efficiency} highlights that through a state-space augmentation, shuffling algorithms -- conceptualized as Markov chains -- outperform \textit{i.i.d.} sampling, achieving \textit{zero} sampling covariance. Using Proposition \ref{proposition:performance_ordering}, we can show that this leads to \textit{zero} limiting covariance $\rmV_{\rvx}, \rmV_{\rvy}$ for all algorithms represented as TTSA. The superiority of shuffling techniques over \textit{i.i.d.} sampling has indeed been studied for specific stochastic optimization settings, such as minimax optimization \citep{das2022sampling,cho2022sgda} and SGD with momentum \citep{tran2021smg}. However, Proposition \ref{proposition:performance_ordering} firmly establishes this at a much broader scope as described in (i) and (ii), such as bilevel optimization with shuffling methods, whose finite-time analysis remains an open problem.

\begin{figure}[t]
    \centering
    \begin{subfigure}[b]{0.48\columnwidth}
        \centering
        \includegraphics[width=\textwidth]{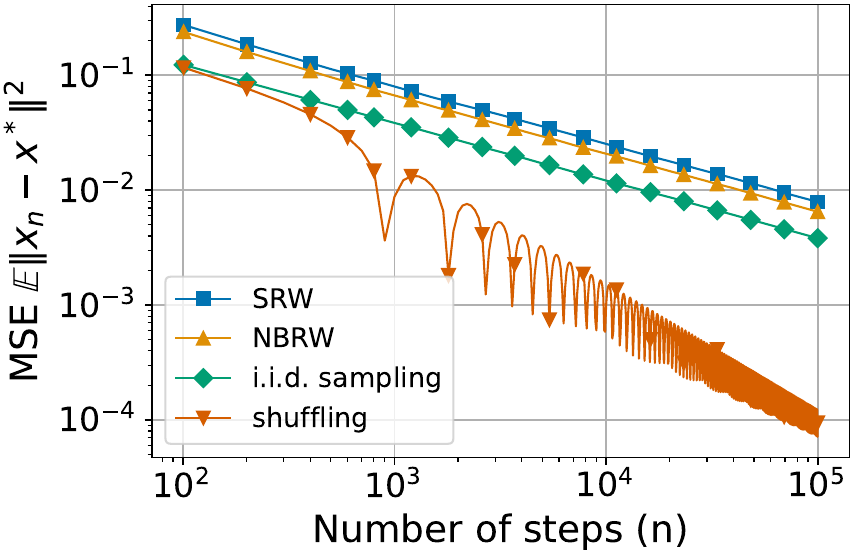}
        \caption{MSE}
        \label{fig:2a}
    \end{subfigure}
    \hfill
    \begin{subfigure}[b]{0.48\columnwidth}
        \centering
        \includegraphics[width=\textwidth]{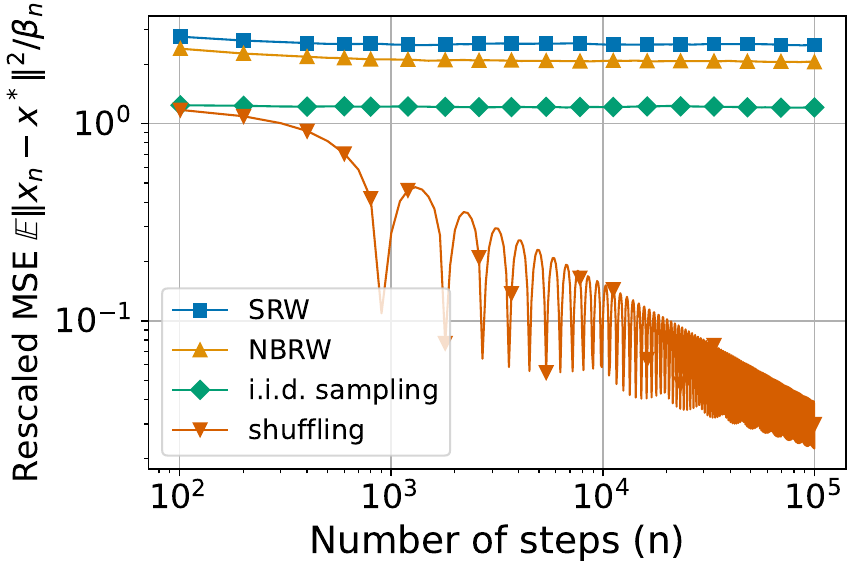}
        \caption{Rescaled MSE}
        \label{fig:2b}
    \end{subfigure}
    \hfill
    \vspace{-0mm}
    \caption{Comparison of the performance among different sampling strategies in momentum SGD.}
    \label{fig:2}
    \vspace{-0mm}
\end{figure}

\textbf{Simulations.} We present numerical experiments for different sampling strategies employed in the momentum SGD algorithm to solve the $L_2$-regularized binary classification problem using the dataset \textit{a9a} (with $123$ features) from LIBSVM \citep{chang2011libsvm}. Specifically, to simulate the token algorithm in distributed learning, we employ NBRW and SRW as the stochastic input to the momentum SGD on the wikiVote graph \citep{leskovec2014snap}, comprising $889$ nodes and $2914$ edges.\footnote{We incorporate both NBRW and SRW with importance reweighting to achieve a uniform target distribution.} Each node on the wikiVote graph is assigned with one data point from the dataset \textit{a9a}, thus $889$ data points in total. We also assess the momentum SGD's performance under \textit{i.i.d.} sampling and single shuffling using the same dataset of size $889$. In Figure \ref{fig:2a}, we observe that NBRW has a smaller MSE than SRW across all time $n$, with a similar trend for single shuffling over \textit{i.i.d.} sampling. Figure \ref{fig:2b} demonstrates that the rescaled MSEs of NBRW, SRW and \textit{i.i.d.} sampling approach some constants, while the curve for single shuffling still decreases in linear rate because eventually the limiting covariance matrix therein will be zero. We defer the detailed simulation settings and more simulation results to Appendix $A.5$.

\subsection{Asymptotic Behavior of Nonlinear GTD Algorithms}\label{section:TDC algortihm}
The CLT result not only allows comparison of limiting covariance matrices of two efficiency-ordered stochastic inputs in distributed learning, but also offers insights into an algorithm's asymptotic performance, as showcased in Table \ref{tab:related_works_TTSA}. This is particularly relevant in RL where the stochastic sequence $\{\xi_n\}$ is generated by a given policy and thus uncontrollable. An important aspect in RL is policy evaluation in MDP with the primary goal of estimating the value function of a given policy, which is essential for further policy improvement \citep{sutton2018reinforcement}. In this part, we focus on a family of gradient-based TD learning (GTD) algorithms, which are instances of TTSA \citep{maei2009convergent,wang2021non}. We leverage Theorem \ref{theorem:CLT_general_TTSA} to derive the pioneering statistical properties of these algorithms when using \textit{nonlinear} value function approximation and Markovian samples for policy evaluation.

Tabular methods for estimating the value function, such as SARSA, have been widely used, but can be problematic when the state-action space is large \citep{sutton2018reinforcement}. TD learning with linear function approximation has been extensively studied \citep{srikant2019finite,doan2019finite,wang2020decentralized,li2023sharp}. In contrast to linear function approximation, nonlinear approaches, e.g. neural networks, are more practical choices known for their strong representation capabilities and eliminating the need for feature mapping \citep{wai2020provably,wang2021non}. However, \citet{tsitsiklis1997an} notes the potential divergence of TD learning with nonlinear function approximation. Addressing the divergence, \citet{maei2009convergent} introduces nonlinear GTD2 and TDC algorithms with almost sure convergence guarantees. These methods iterate over gradients of the mean-square projected Bellman error (MSPBE) in order to obtain the best estimate of the nonlinear value function that minimizes MSPBE \citep{maei2009convergent,xu2021sample,wang2021non}.

While non-asymptotic analyses of GTD2 and TDC algorithms have been established for both \textit{i.i.d.} and Markovian settings with linear approximations \citep{karmakar2018two,dalal2018finite,dalal2020tale,kaledin2020finite,li2023sharp}, results for the nonlinear function approximation remain scarce since MSEPBE becomes nonconvex and the two-timescale update rule is nonlinear. For asymptotic analysis, \citet{karmakar2018two} studies the almost sure convergence of general TTSA and applies it to nonlinear TDC algorithm, extending from \textit{i.i.d.} \citep{maei2009convergent} to Markovian samples. This analysis can also be applied to nonlinear GTD2 algorithm. Only a few works \citep{xu2021sample,wang2021non} provide non-asymptotic analysis specifically for nonlinear TDC algorithm with Markovian samples and constant step sizes while the results cannot be extrapolated to nonlinear GTD2 algorithm. Therefore, a comprehensive analysis of these algorithms with Markovian samples under decreasing step sizes remains lacking in RL. 

We now summarize nonlinear GTD2 and TDC algorithms, followed by their asymptotic results in Proposition \ref{proposition:asymptotic_result_TDC_algorithm}. An MDP is defined as a $5$-tuple $(\gS,\gA,P,r,\alpha)$, where $\gS$ and $\gA$ are the finite state and action spaces, and $P$ and $r$ are transition kernel and reward function, with $\alpha$ being a discount factor. A policy $\vpi$ maps each state $s \!\in\! \gS$ onto an action probability distribution $\vpi(\cdot|s)$, with $\vmu^{\vpi}$ being the corresponding stationary distribution. The Markovian samples $\{s_n\}$ then follow the transition probability $\sP(s, s') \!=\! \sum_{a\in\gA}\! P(s'|s,a)\vpi(a|s)$. The value function for policy $\vpi$ from initial state $s$ is $W^{\vpi}(s) \!=\! \E_{\vpi}[\sum_{n=0}^{\infty} \alpha^n r_n | s_0 = s]$, where $r_n \!\triangleq\! r(s_n, a_n, s_{n+1})$. The GTD2 and TDC algorithms estimate $W^{\vpi}(s)$ via nonlinear functions $W_{\rvx}(s)$ and its feature function $\phi_{\rvx}(s) = \nabla_{\rvx}W_{\rvx}(s)$ parameterized by $\rvx$. For linear approximation $W_{\rvx}(s) = \phi(s)^T\rvx$, $\phi(s)$ is independent of $\rvx$. However, with nonlinear $W_{\rvx}(s)$, $\phi_{\rvx}(s)$ depends on $\rvx$. Defining TD error as $\delta_n = r_n + \alpha W_{\rvx_n}(s_{n+1}) - W_{\rvx_n}(s_{n})$, the iterates $(\rvx_n, \rvy_n)$ of the GTD2 and TDC algorithms admit an equilibrium $(\rvx^*, \rvy^*)$, with $\rvx^*$ ensuring $\E_{s_n \sim \vmu}[\delta_n(\rvx^*)\phi_{\rvx^*}(s_n)] = \vzero$, and $\rvy^* = \vzero$. Details of these algorithms and conditions for the following CLT results are in Appendix $A.4.1$.
\begin{proposition}\label{proposition:asymptotic_result_TDC_algorithm}
    For both nonlinear GTD2 and TDC algorithms under Markovian samples, we have 
    \begin{align*} 
        &\lim_{n\to\infty} \rvx_n = \rvx^* \quad \text{a.s.} ~~\text{and} \quad \lim_{n\to\infty} \rvy_n = \vzero \quad \text{a.s.} \\ 
        &\frac{1}{\sqrt{\beta_n}}(\rvx_n - \rvx^*)\! \xrightarrow{~d~} N(\vzero,\rmV_{\rvx}), ~ \frac{1}{\sqrt{\gamma_n}}\rvy_n\xrightarrow{~d~} N(\vzero,\rmV_{\rvy}), 
    \end{align*}
    where $\rmV_{\rvx}, \rmV_{\rvy}$ are identical for both algorithms. 
\end{proposition}
The proof of Proposition \ref{proposition:asymptotic_result_TDC_algorithm} and the exact forms of $\rmV_{\rvx}, \rmV_{\rvy}$ are in Appendix $A.4.2$. This proposition offers a state-of-the-art performance analysis of nonlinear GTD2 and TDC algorithms in RL, employing Markovian samples and general decreasing step sizes. While \citet{doan2021finite,zeng2021two} provide finite-time bounds within the general TTSA framework, their applicability to nonlinear GTD2 and TDC algorithms is restricted by specific choice of the step sizes, as explained in Section \ref{section:related work}. The usefulness of these finite-time results are further limited due to the lack of any definitive indication regarding the tightness of the bounds associated with these two algorithms. Moreover, empirical studies \citep{dann2014policy,ghiassian2020gradient} have not consistently favored either one of the two algorithms when compared across all tasks, leading to a lack of consensus regarding which one is the better performing overall. Proposition \ref{proposition:asymptotic_result_TDC_algorithm} clarifies that, in the long run, both GTD2 and TDC algorithms exhibit identical behaviors under the CLT scaling. 

\begin{figure}[t]
    \centering
    \begin{subfigure}[b]{0.48\columnwidth}
        \centering
        \includegraphics[width=\textwidth]{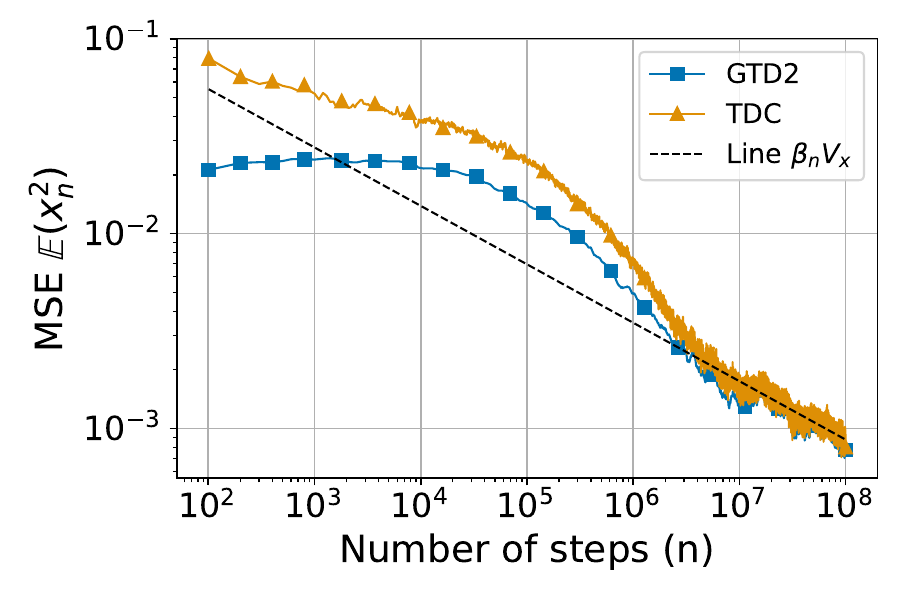}
        \caption{MSE}
        \label{fig:3a}
    \end{subfigure}
    \hfill
    \begin{subfigure}[b]{0.48\columnwidth}
        \centering
        \includegraphics[width=\textwidth]{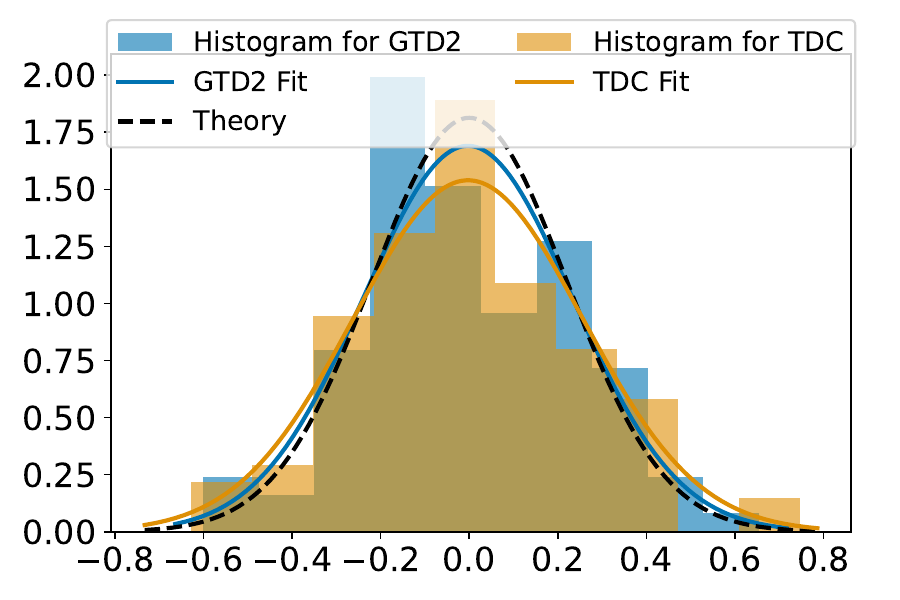}
        \caption{Histograms of $\beta_n^{-1/2}x_n$}
        \label{fig:3b}
    \end{subfigure}
    \hfill
    \vspace{-0mm}
    \caption{Comparison of nonlinear GTD2 and TDC algorithms in the $5$-state random walk task.}
    \label{fig:3}
    \vspace{-0mm}
\end{figure}
\textbf{Simulations.} We consider a $5$-state random walk task \citep{dann2014policy,sutton2018reinforcement} for nonlinear GTD2 and TDC algorithms. Each state can transit to the right or left next state with probability $0.5$, with reward $+0.5$ if turning right or $-0.5$ otherwise. Let discount factor $\alpha \!=\! 0.9$, we consider the nonlinear value function $W_{x}(s) \!=\! a(s)(e^{\kappa x}\!-\!1)$ for a scalar parameter $x$, where $a = [-2, -6, -3, -4, -5], \kappa \!=\! 0.1$. The ground truth $W(s) \!=\! 0$ for $s \!\in\! [5]$ such that $x^* \!=\! 0$ for $W_{\rvx}(s)$ achieves the accurate approximation. Figure \ref{fig:3a} illustrates the long-term performance of GTD2 and TDC algorithms. Starting from $n = 10^7$, they align with the line $\beta_n \rmV_{\rvx}$, with $\rmV_{\rvx}$ being a scalar from Proposition \ref{proposition:asymptotic_result_TDC_algorithm}. This reaffirms the relationship between MSE and CLT, as detailed in Section \ref{section:main_results}. Figure \ref{fig:3b} displays a histogram of $\beta_n^{-1/2} x_n$, generated from $100$ independent trials at $n = 10^8$ for both algorithms. We show that their experimental density curves are close to the theoretical Gaussian density with zero mean and variance $\rmV_{\rvx}$.  We defer the detailed simulation settings, calculations related to Figure \ref{fig:3}, and additional simulation results to Appendix $A.5$.

\section{CONCLUSION}
In this paper, we present the first CLT analysis of nonlinear TTSA in the context of controlled Markovian noise with general forms of decreasing step sizes. Our result greatly extends the scope of existing literature by allowing most general settings and performance ordering across nonlinear TTSA algorithms, notably in distributed optimization and RL. Our work highlights the potential of asymptotic analysis for algorithmic improvement and comparison, addressing areas where conventional finite-time analysis fall short, thus suggesting that more attention should be given to asymptotic statistics. 

\section{ACKNOWLEDGMENTS}
We thank the anonymous reviewers for their constructive comments. This work was supported in part by National Science Foundation under Grant Nos. CNS-2007423 and IIS-1910749.

\bibliographystyle{apalike}
\bibliography{jie_machine_learning_mega_references}

\section*{Checklist}

 \begin{enumerate}

 \item For all models and algorithms presented, check if you include:
 \begin{enumerate}
   \item A clear description of the mathematical setting, assumptions, algorithm, and/or model. [Yes]
   \item An analysis of the properties and complexity (time, space, sample size) of any algorithm. [Yes]
   \item (Optional) Anonymized source code, with specification of all dependencies, including external libraries. [No]
 \end{enumerate}

 \item For any theoretical claim, check if you include:
 \begin{enumerate}
   \item Statements of the full set of assumptions of all theoretical results. [Yes]
   \item Complete proofs of all theoretical results. [Yes]
   \item Clear explanations of any assumptions. [Yes]     
 \end{enumerate}

 \item For all figures and tables that present empirical results, check if you include:
 \begin{enumerate}
    \item The code, data, and instructions needed to reproduce the main experimental results (either in the supplemental material or as a URL). [No]
    \item All the training details (e.g., data splits, hyperparameters, how they were chosen). [Yes]
     \item A clear definition of the specific measure or statistics and error bars (e.g., with respect to the random seed after running experiments multiple times). [Yes]
     \item A description of the computing infrastructure used. (e.g., type of GPUs, internal cluster, or cloud provider). [Yes]
 \end{enumerate}

 \item If you are using existing assets (e.g., code, data, models) or curating/releasing new assets, check if you include:
 \begin{enumerate}
   \item Citations of the creator If your work uses existing assets. [Yes]
   \item The license information of the assets, if applicable. [Not Applicable]
   \item New assets either in the supplemental material or as a URL, if applicable. [Not Applicable]
   \item Information about consent from data providers/curators. [Not Applicable]
   \item Discussion of sensible content if applicable, e.g., personally identifiable information or offensive content. [Not Applicable]
 \end{enumerate}

 \item If you used crowdsourcing or conducted research with human subjects, check if you include:
 \begin{enumerate}
   \item The full text of instructions given to participants and screenshots. [Not Applicable]
   \item Descriptions of potential participant risks, with links to Institutional Review Board (IRB) approvals if applicable. [Not Applicable]
   \item The estimated hourly wage paid to participants and the total amount spent on participant compensation. [Not Applicable]
 \end{enumerate}

 \end{enumerate}

\newpage

 \renewcommand{\thesection}{A.\arabic{section}}
\onecolumn
\aistatstitle{Central Limit Theorem for Two-Timescale Stochastic Approximation with Markovian Noise: Theory and Applications \\
Supplementary Materials}

\setcounter{section}{0}
\section{Proof of Lemma 2.1}\label{appendix:proof_of_a.s_convergence}
The two-timescale SA form with iterate-dependent Markov chain we consider in this paper is as follows:
\begin{subequations}\label{eqn:general_TTSA_form}
\begin{equation}
    \rvx_{n+1} = \rvx_n + \beta_{n+1} h_1(\rvx_n, \rvy_n, \xi_{n+1}),
\end{equation}
\begin{equation}
    \rvy_{n+1} = \rvy_n + \gamma_{n+1} h_2(\rvx_n, \rvy_n, \xi_{n+1}),
\end{equation}
\end{subequations}
with the goal of finding the root $(\rvx^*,\rvy^*)$ such that
\begin{equation}\label{eqn:def_h1_h2}
    \bar h_1(\rvx^*,\rvy^*) = \E_{\xi \sim \vmu}[h_1(\rvx^*,\rvy^*,\xi)] = 0, \quad \bar h_2(\rvx^*,\rvy^*) = \E_{\xi \sim \vmu}[h_2(\rvx^*,\rvy^*,\xi)] = 0.
\end{equation}
For self-contained purposes, we reproduce assumptions \ref{assump:1} -- \ref{assump:5} for the TTSA algorithm \eqref{eqn:general_TTSA_form} below.
\begin{enumerate}[label=A\arabic*., ref=(A\arabic*)]
  \item The step sizes $\beta_n \triangleq (n+1)^{-b}$ and $\gamma_n \triangleq (n+1)^{-a}$, where $0.5 < a < b \leq 1$. \label{assump:1}
  \item For the $C^1$ function $h_1 : \sR^{d_1} \times \sR^{d_2} \times \Xi \to \sR^{d_1}$, there exists a positive constant $L_1$ such that $\|h_1(\rvx,\rvy,\xi)\| \leq L_1(1 + \|\rvx\| + \|\rvy\|)$ for every $\rvx \in \sR^{d_1}, \rvy \in \sR^{d_2}, \xi \in \Xi$. The same condition holds for the $C^1$ function $h_2$ as well. \label{assump:2}
  \item Consider a $C^1$ function $\lambda : \sR^{d_1} \to \sR^{d_2}$. For every $\rvx \in \sR^{d_1}$, the following three properties hold: (i) $\lambda(\rvx)$ is the globally attracting point of the related ODE $\dot \rvy = \bar h_2(\rvx,\rvy)$; (ii) $\nabla_{\rvy}\bar{h}_2(\rvx,\lambda(\rvx))$ is Hurwitz; (iii) $\|\lambda(\rvx)\| \leq L_2(1+\|\rvx\|)$ for some positive constant $L_2$. Additionally, let $\hat{h}_1(\rvx) \triangleq \bar{h}_1(\rvx,\lambda(\rvx))$, there exists a set of disjoint roots $\Lambda \triangleq \{\rvx^*: \hat{h}_1(\rvx^*)  = 0, \nabla_{\rvx} \hat{h}_1(\rvx^*) + \frac{\mathds{1}_{\{b=1\}}}{2}\rmI \text{ is Hurwitz}\}$, which is also the globally attracting set for trajectories of the related ODE $\dot \rvx = \hat h_1(\rvx)$. \label{assump:3} 
  \item $\{\xi_n\}_{n\geq 0}$ is an iterate-dependent Markov chain on finite state space $\Xi$. For every $n\geq 0$, $P(\xi_{n+1} = j | \rvx_m,\rvy_m, \xi_m, 0 \leq m \leq n) = P(\xi_{n+1} = j | \rvx_n,\rvy_n, \xi_n=i) = \rmP_{i,j}[\rvx_n,\rvy_n]$, where the transition kernel $\rmP[\rvx,\rvy]$ is continuous in $\rvx, \rvy$, and the Markov chain generated by $\rmP[\rvx,\rvy]$ is ergodic so that it admits a stationary distribution $\vpi(\rvx,\rvy)$, and $\vpi(\rvx^*,\lambda(\rvx^*)) = \vmu$. \label{assump:4} 
  \item $\sup_{n\geq 0} (\|\rvx_n\| + \|\rvy_n\|) < \infty$ a.s. \label{assump:5}
\end{enumerate}

Now, we translate the assumptions in \citet{yaji2020stochastic} below in our notations and TTSA algorithm \eqref{eqn:general_TTSA_form} in order to apply the almost sure convergence result therein.
\begin{enumerate}[label=B\arabic*., ref=(B\arabic*)]
  \item The step sizes $\beta_n \triangleq n^{-b}$ and $\gamma_n \triangleq n^{-a}$, where $0.5 < a < b \leq 1$. \label{assump:Bone}
  \item Assume the function $h_1(\rvx,\rvy,\xi)$ is continuous and differentiable with respect to $\rvx, \rvy$. There exists a positive constant $L_1$ such that $\|h_1(\rvx,\rvy,\xi)\| \leq L_1(1 + \|\rvx\| + \|\rvy\|)$ for every $\rvx \in \sR^{d_1}, \rvy \in \sR^{d_2}, \xi \in \Xi$. The same condition holds for the function $h_2$ as well. \label{assump:Btwo}
  \item Assume there exists a continuous function $\lambda : \sR^{d_1} \to \sR^{d_2}$ such that the following two properties hold for any $\rvx \in \sR^{d_1}$: (i) $\|\lambda(\rvx)\| \leq L_2(1+\|\rvx\|)$ for some positive constant $L_2$; (ii) the ODE $\dot \rvy = \bar h_2(\rvx, \rvy)$ has a globally asymptotically stable equilibrium $\lambda(\rvx)$ such that $\bar h_2(\rvx,\lambda(\rvx)) = 0$. Additionally, let $\hat{h}_1(\rvx) \triangleq \bar h_1(\rvx,\lambda(\rvx))$, there exists a set of disjoint roots $\Lambda \triangleq \{\rvx^*: \hat{h}_1(\rvx^*)  = 0\}$, which is the set of globally asymptotically stable equilibria of the ODE $\dot\rvx = \hat h_1(\rvx)$. \label{assump:Bthree}
  \item $\{\xi_n\}_{n\geq 0}$ is an iterate-dependent Markov process in finite state space $\Xi$. For every $n\geq 0$, $P(X_{n+1} = j | \rvx_m,\rvy_m, \xi_m, 0 \leq m \leq n) = P(X_{n+1} = j | \rvx_n,\rvy_n, \xi_n=i) = \rmP_{i,j}[\rvx_n,\rvy_n]$, where the transition kernel $\rmP[\rvx,\rvy]$ is continuous in $\rvx, \rvy$, and the Markov chain generated by $\rmP[\rvx,\rvy]$ is ergodic so that it admits a stationary distribution $\vpi(\rvx,\rvy)$, and $\vpi(\rvx^*,\lambda(\rvx^*)) = \vmu$. \label{assump:Bfour}
  \item $\sup_{n\geq 0} (\|\rvx_n\| + \|\rvy_n\|) < \infty$ a.s. \label{assump:Bfive}
\end{enumerate}
Assumption \ref{assump:Bone} is the standard condition on step sizes corresponding to \citet[Assumption A5]{yaji2020stochastic}. Assumption \ref{assump:Btwo} translates \citet[Assumptions A1, A2]{yaji2020stochastic} from functions $h_1, h_2$ with set values to those with single values. Assumption \ref{assump:Bthree} is the condition on the relevant ODEs of the TTSA algorithm \eqref{eqn:general_TTSA_form}, which is derived from \citet[Assumption A9 -- A11]{yaji2020stochastic}. Assumption \ref{assump:Bfour} simplifies \citet[Assumptions A3, A4]{yaji2020stochastic} by using a single Markov sequence $\{\xi_n\}$ for both iterations in \eqref{eqn:general_TTSA_form}. This has a wide range of applications, such as constrained convex optimization in \citet[Section 7]{yaji2020stochastic}, performance ordering in distributed learning and reinforcement learning algorithms using Markovian samples, which is discussed in Section 3 of our paper. Assumption \ref{assump:Bfive} corresponds to \citet[Assumption A8]{yaji2020stochastic}. Besides, \citet[Assumptions A6, A7]{yaji2020stochastic} are automatically satisfied since the noise terms therein are set to zero in \eqref{eqn:general_TTSA_form}. In the following, we provide the existing almost sure convergence result.
\begin{theorem}[\citet{yaji2020stochastic} Theorem 4]
    Under Assumptions \ref{assump:Bone} -- \ref{assump:Bfive}, for TTSA algorithm \eqref{eqn:general_TTSA_form}, we have
    \begin{equation*}
        \begin{pmatrix}
            \rvx_n \\
            \rvy_n
        \end{pmatrix} \xrightarrow[n\to\infty]{a.s.} \bigcup_{\rvx^* \in \Lambda} \begin{pmatrix}
            \rvx^* \\
            \lambda(\rvx^*)
        \end{pmatrix}.
    \end{equation*}
\end{theorem}
Our Assumptions \ref{assump:1} -- \ref{assump:5} correspond to Assumptions \ref{assump:Bone} -- \ref{assump:Bfive}. Consequently, Lemma 2.1 is a direct application of \citet[Theorem 4]{yaji2020stochastic}. Compared to assumption \ref{assump:Bthree}, the additional conditions in our Assumption \ref{assump:3}, i.e., $\nabla_{\rvy}\bar{h}_2(\rvx,\lambda(\rvx))$ is Hurwitz for every $\rvx \in \sR^{d_1}$ and $\hat h_1(\rvx^*)$ is Hurwitz for $\rvx^* \in \Lambda$, are necessary for the proof of our CLT results in Theorem 2.2 and will be utilized in the following section.

\section{Proof of Theorem 2.2}\label{appendix:proof_CLT}
Without loss of generality, the proof in this part is conditioned on the event that $\{\rvx_n \to \rvx^*, \rvy_n \to \rvy^* \triangleq \lambda(\rvx^*)\}$ for some $\rvx^* \in \Lambda$.
The proof of Theorem 2.2 includes two parts: First, in Appendix \ref{appendix:2.1}, we decompose the functions $h_1(\rvx,\rvy,\xi)$ and $h_2(\rvx,\rvy,\xi)$ of \eqref{eqn:general_TTSA_form} into several terms and quantify the asymptotic behavior of each term. Second, in Appendix \ref{appendix:2.2}, we partition those terms in each iteration into three parts (six sequences in total) and show that there is a sequence $L_n^{(\rvx)}$ ($L_n^{(\rvy)}$ resp.) in each iteration that contributes to weak convergence, while the remaining sequences diminish to zero when multiplied by the CLT scale $1/\sqrt{\beta_n}$ for iterates $\rvx_n$ ($1/\sqrt{\gamma_n}$ for iterates $\rvy_n$ resp.) so that they do not play a role in the final CLT result. The proofs of all technical lemmas used in Appendix \ref{appendix:2.1} and Appendix \ref{appendix:2.2} are deferred to Appendix \ref{appendix:2.3} -- \ref{appendix:2.6} for better readability.

\subsection{Decomposition of Markovian Noise in the TTSA Algorithm}\label{appendix:2.1}
Throughout Appendix \ref{appendix:proof_CLT}, we define an operator $(\rmP m)(\rvx,\rvy,i)$ for any function $m: \sR^{d_1} \times \sR^{d_2} \times \Xi \to \sR^d$ as follows:
\begin{equation}\label{eqn:def_poisson_operator}
    (\rmP m)(\rvx,\rvy,i) \triangleq \sum_{j\in\Xi} \rmP_{i,j}[\rvx,\rvy] m(\rvx,\rvy,j). 
\end{equation}

The ultimate goal in this subsection is to decompose \eqref{eqn:general_TTSA_form} into  
\begin{subequations}\label{eqn:TTSA_decomposition_3}
    \begin{equation}
        \rvx_{n+1} = \rvx_n + \beta_{n+1}\bar h_1(\rvx_n,\rvy_n) +\beta_{n+1} M_{n+1}^{(\rvx)} + \beta_{n+1} r^{(\rvx,1)}_n + \beta_{n+1} r^{(\rvx,2)}_n,
    \end{equation}
    \begin{equation}
        \rvy_{n+1} = \rvy_n + \gamma_{n+1} \bar h_2(\rvx_n,\rvy_n) + \gamma_{n+1}M_{n+1}^{(\rvy)} + \gamma_{n+1} r^{(\rvy,1)}_n + \gamma_{n+1} r^{(\rvy,2)}_n,
    \end{equation}
\end{subequations}
where $M_{n+1}^{(\rvx)}, M_{n+1}^{(\rvy)}$ are two Martingale difference noise terms adapted to the filtration $\gF_n \triangleq \{\rvx_0,\rvy_0, \xi_1, \cdots, \xi_n\}$. For $i = 1, 2$, $r^{(\rvx,i)}_n, r^{(\rvy,i)}_n$ are additional noise terms derived from Markovian noise and do not exist for \textit{i.i.d.} sequence $\{\xi_n\}$. Therefore, these terms are absent from the previous CLT results for TTSA with Martingale difference noise \citep{konda2004convergence, mokkadem2006convergence}, where the \textit{i.i.d.} sequence $\{\xi_n\}$ is their main focus.

We first rewrite \eqref{eqn:general_TTSA_form} as
\begin{subequations}\label{eqn:TTSA_decomposition_1}
    \begin{equation}\label{eqn:TTSA_decomposition_1_x}
        \rvx_{n+1} = \rvx_n + \beta_{n+1} \Bar{h}_1(\rvx_n,\rvy_n) + \beta_{n+1} (h_1(\rvx_n, \rvy_n,\xi_{n+1}) - \Bar{h}_1(\rvx_n,\rvy_n)),
    \end{equation}
    \begin{equation}\label{eqn:TTSA_decomposition_1_y}
        \rvy_{n+1} = \rvy_n + \gamma_{n+1} \Bar{h}_2(\rvx_n,\rvy_n) + \gamma_{n+1} (h_2(\rvx_n, \rvy_n,\xi_{n+1}) - \Bar{h}_2(\rvx_n,\rvy_n)).
    \end{equation}
\end{subequations}
Then, given the underlying state-dependent Markov chain $\{\xi_n\}_{n\geq 0}$ with transition kernel $\rmP[\rvx,\rvy]$ that satisfies Assumption \ref{assump:4}, there exists a solution $m_1( \rvx,\rvy,\cdot): \Xi \to \sR^{d_1}$ to the following Poisson equation:
\begin{equation}\label{eqn:poisson_eq_h1}
    h_1(\rvx,\rvy,\xi) - \bar{h}_1(\rvx,\rvy) = m_1( \rvx,\rvy,\xi) - (\rmP m_1)(\rvx,\rvy,\xi).
\end{equation}
Similarly, there exists a solution $m_2( \rvx,\rvy,\cdot): \Xi \to \sR^{d_2}$ to the following:
\begin{equation}\label{eqn:poisson_eq_h2}
    h_2(\rvx,\rvy,\xi) - \bar{h}_2(\rvx,\rvy) = m_2( \rvx,\rvy,\xi) - (\rmP m_2)(\rvx,\rvy,\xi).
\end{equation}
This Poisson equation technique has been well discussed in \citet[Section 2]{chen2020explicit} and \citet{benveniste2012adaptive,meyn2022control}. For $l \in \{1, 2\}$, the explicit form of the solution $m_l$ to the corresponding Poisson equation \eqref{eqn:poisson_eq_h1} or \eqref{eqn:poisson_eq_h2} is given by
\begin{equation}\label{eqn:sol_Poisson_eq_closed_form}
 m_l(\rvx,\rvy,i) = \sum_{j\in\Xi}\sum_{k=0}^{\infty} \left(\rmP[\rvx,\rvy] - \vone\vpi[\rvx,\rvy]^T\right)_{(i,j)}^k h_l(\rvx,\rvy,j) = \sum_{j\in\Xi}\left(\rmI - \rmP[\rvx,\rvy] + \vone\vpi[\rvx,\rvy]^T\right)^{-1}_{(i,j)} h_l(\rvx,\rvy,j),
\end{equation}   
where $\vpi[\rvx,\rvy]$ is the stationary distribution of $\rmP[\rvx,\rvy]$, and $(i,j)$ represents the $(i,j)$-th entry of the corresponding matrix. 
The proof of \eqref{eqn:sol_Poisson_eq_closed_form} can be found in \citet[Appendix B.3.1 (B.16)]{delyon2000stochastic} and \citet[Appendix B]{hu2022efficiency}. Now, by \eqref{eqn:poisson_eq_h1}, \eqref{eqn:poisson_eq_h2} and \eqref{eqn:sol_Poisson_eq_closed_form}, we can further rewrite \eqref{eqn:TTSA_decomposition_1} as
\begin{subequations}\label{eqn:TTSA_decomposition_2}
\begin{equation}
\begin{split}
    \rvx_{n+1} =& ~ \rvx_n + \beta_{n+1} \bar h_1(\rvx_n,\rvy_n) + \beta_{n+1}\underbrace{(m_1(\rvx_n,\rvy_n,\xi_{n+1}) - (\rmP m_1)(\rvx_n,\rvy_n,\xi_n))}_{M_{n+1}^{(\rvx)}}\\
    &+ \beta_{n+1}\underbrace{((\rmP m_1)(\rvx_{n+1}, \rvy_{n+1}, \xi_{n+1}) - (\rmP m_1)(\rvx_{n}, \rvy_{n}, \xi_{n+1}))}_{r^{(\rvx,1)}_n}  \\
    &+ \beta_{n+1}\underbrace{((\rmP m_1)(\rvx_n, \rvy_n, \xi_n) - (\rmP m_1)(\rvx_{n+1}, \rvy_{n+1}, \xi_{n+1}))}_{r^{(\rvx,2)}_n},
\end{split}
\end{equation} 
and similarly,
\begin{equation}
    \begin{split}
    \rvy_{n+1} =& ~ \rvy_n + \gamma_{n+1} \bar h_2(\rvx_n,\rvy_n) + \gamma_{n+1}\underbrace{(m_2(\rvx_n,\rvy_n,\xi_{n+1}) - (\rmP m_2)(\rvx_n,\rvy_n,\xi_n))}_{M_{n+1}^{(\rvy)}}\\
    &+ \gamma_{n+1}\underbrace{((\rmP m_2)(\rvx_{n+1}, \rvy_{n+1}, \xi_{n+1}) - (\rmP m_2)(\rvx_{n}, \rvy_{n}, \xi_{n+1}))}_{r^{(\rvy,1)}_n}  \\
    &+ \gamma_{n+1}\underbrace{((\rmP m_2)(\rvx_n, \rvy_n, \xi_n) - (\rmP m_2)(\rvx_{n+1}, \rvy_{n+1}, \xi_{n+1}))}_{r^{(\rvy,2)}_n},
    \end{split}
\end{equation}   
\end{subequations}
which becomes \eqref{eqn:TTSA_decomposition_3}. This kind of decomposition is well known for single-timescale SA with Markovian noise \citep{delyon2000stochastic, benveniste2012adaptive, fort2015central,fort2016convergence}, but now we need to deal with coupled variables $(\rvx_n,\rvy_n)$ for each iteration in \eqref{eqn:TTSA_decomposition_2}. 

Now, we further decompose the covariance of $M_{n+1}^{(\vtheta)},M_{n+1}^{(\rvx)}$ in Lemma \ref{lemma:decomposition_Mn} and characterize the asymptotic behavior of each decomposed term in Lemma \ref{lemma:characteristics_decomposition}, which will be used later in the proof of Lemma \ref{lemma:weak_convergence_Ln} in the next subsection and is critical in quantifying the limiting covariance matrix in our main CLT result.
\begin{lemma}\label{lemma:decomposition_Mn}
For $M_{n+1}^{(\vtheta)},M_{n+1}^{(\rvx)}$ defined in \eqref{eqn:TTSA_decomposition_2}, their covariance can be decomposed into the following forms:
\begin{equation*}
    \E\left[\left.M^{(\rvx)}_{n+1}(M^{(\rvx)}_{n+1})^T\right|\gF_n\right] = \rmU_{11} + \rmD_n^{(11)} + \rmJ_n^{(11)},
\end{equation*}
\begin{equation*}
    \E\left[\left.M^{(\rvx)}_{n+1}(M^{(\rvy)}_{n+1})^T\right|\gF_n\right] = \rmU_{12} + \rmD_n^{(12)} + \rmJ_n^{(12)}.
\end{equation*}
\begin{equation*}
    \E\left[\left.M^{(\rvy)}_{n+1}(M^{(\rvx)}_{n+1})^T\right|\gF_n\right] = \rmU_{21} + \rmD_n^{(21)} + \rmJ_n^{(21)}.
\end{equation*}
\begin{equation*}
    \E\left[\left.M^{(\rvy)}_{n+1}(M^{(\rvy)}_{n+1})^T\right|\gF_n\right] = \rmU_{22} + \rmD_n^{(22)} + \rmJ_n^{(22)}.
\end{equation*}
where $\rmU_{12} = \rmU_{21}^T$, $\rmD_n^{(12)} = (\rmD_n^{(21)})^T$, $\rmJ_n^{(12)} = (\rmJ_n^{(21)})^T$.
\end{lemma}

\begin{lemma}\label{lemma:characteristics_decomposition}
For $i,j \in \{1, 2\}$, and with $\rmU_n^{(ij)}, \rmD_n^{(ij)}$ and $\rmJ_n^{(ij)}$ as stated in Lemma \ref{lemma:decomposition_Mn}, we have
\begin{equation*}
\rmU_{ij} = \lim_{s\to\infty}\frac{1}{s}\E\left[\left(\sum_{n=1}^s h_i(\rvx^*,\rvy^*,\xi_{n})\right)\left(\sum_{n=1}^s h_j(\rvx^*,\rvy^*,\xi_{n})\right)^T\right],
\end{equation*}
\begin{equation*}
    \lim_{n\to\infty} \rmD_n^{(ij)} = 0 ~~ \text{a.s.}
\end{equation*}
\begin{equation*}
\lim_{n\to\infty} \gamma_n \E\left[\left\|\sum_{k=1}^n \rmJ^{(ij)}_k\right\|\right] = 0 ~~ \text{a.s.}
\end{equation*}
\end{lemma}
The proof of Lemmas \ref{lemma:decomposition_Mn} and \ref{lemma:characteristics_decomposition} are located in Appendices \ref{appendix:2.3} and \ref{appendix:2.4}.

\vspace{1mm}
\begin{remark}\label{remark:1}
    We note that for \textit{i.i.d.} sequence $\{\xi_n\}$ with marginal $\vmu$, matrix $\rmJ_n^{(ij)}$ in Lemma \ref{lemma:decomposition_Mn} is \textit{zero} for all time $n$, as indicated in the proof of Lemma \ref{lemma:characteristics_decomposition}. Thus, the condition on $\rmJ_n^{(ij)}$ in Lemma \ref{lemma:characteristics_decomposition} is automatically satisfied. Furthermore, the matrix $\rmU_{ij}$ in Lemma \ref{lemma:characteristics_decomposition} reduces to the marginal covariance $\E_{\xi\sim\vmu}[h_i(\rvx^*,\rvy^*,\xi)h_j(\rvx^*,\rvy^*,\xi)^T]$. Thus, Lemma \ref{lemma:decomposition_Mn} and Lemma \ref{lemma:characteristics_decomposition} analyze the general Markovian sequence $\{\xi_n\}$ and include the \textit{i.i.d.} sequence $\{\xi_n\}$ as a special case, whose resulting terms have been mainly analyzed in \citet{konda2004convergence,mokkadem2006convergence}.
\end{remark}

\vspace{1mm}
\begin{remark}\label{remark:2}
    In Section 2.2, we originally obtain the matrix $\rmU_{\rvx}$ in the following form: 
    \begin{equation*}
        \rmU_{\rvx} = \lim_{s \to \infty}\frac{1}{s} \E\left[\left(\sum_{n=1}^s \tilde\Delta_n^{\rvx}\right)\left(\sum_{n=1}^s\tilde\Delta_n^{\rvx}\right)^T\right],
    \end{equation*}
\end{remark}
where $$\tilde\Delta_n^{\rvx} = h_1(\rvx^*, \rvy^*, \xi_n) - \rmQ_{12}\rmQ_{22}^{-1} h_2(\rvx^*,\rvy^*,\xi_n) = \begin{bmatrix}
    \rmI & -\rmQ_{12}\rmQ_{22}^{-1}
\end{bmatrix} \begin{bmatrix}
    h_1(\rvx^*, \rvy^*, \xi_n) \\ h_2(\rvx^*,\rvy^*,\xi_n)
\end{bmatrix}.$$
Then, $\rmU_{\rvx}$ can be rewritten as
\begin{equation}\label{eqn:alt_matrix_U_x_form}
\begin{split}
    \rmU_{\rvx} &= \lim_{s \to \infty}\frac{1}{s} \E\left[\left(\sum_{n=1}^s \begin{bmatrix}
    \rmI & -\rmQ_{12}\rmQ_{22}^{-1}
\end{bmatrix} \begin{bmatrix}
    h_1(\rvx^*, \rvy^*, \xi_n) \\ h_2(\rvx^*,\rvy^*,\xi_n)
\end{bmatrix}\right)\left(\sum_{n=1}^s\begin{bmatrix}
    \rmI & -\rmQ_{12}\rmQ_{22}^{-1}
\end{bmatrix} \begin{bmatrix}
    h_1(\rvx^*, \rvy^*, \xi_n) \\ h_2(\rvx^*,\rvy^*,\xi_n)
\end{bmatrix}\right)^T\right] \\
&= \lim_{s \to \infty}\frac{1}{s} \begin{bmatrix}
    \rmI & -\rmQ_{12}\rmQ_{22}^{-1}
\end{bmatrix} \E\left[\left(\sum_{n=1}^s \begin{bmatrix}
    h_1(\rvx^*, \rvy^*, \xi_n) \\ h_2(\rvx^*,\rvy^*,\xi_n)
\end{bmatrix}\right)\left(\sum_{n=1}^s \begin{bmatrix}
    h_1(\rvx^*, \rvy^*, \xi_n) \\ h_2(\rvx^*,\rvy^*,\xi_n)
\end{bmatrix}\right)^T\right] \begin{bmatrix}
    \rmI & -\rmQ_{12}\rmQ_{22}^{-1}
\end{bmatrix}^T \\
&= \begin{bmatrix}
    \rmI & -\rmQ_{12}\rmQ_{22}^{-1}
\end{bmatrix} \begin{bmatrix}
    \rmU_{11} & \rmU_{12} \\ \rmU_{21} & \rmU_{22}
\end{bmatrix} \begin{bmatrix}
    \rmI & -\rmQ_{12}\rmQ_{22}^{-1}
\end{bmatrix}^T,
\end{split}
\end{equation}
where the last equality comes from the expression of $\rmU_{ij}$ in Lemma \ref{lemma:characteristics_decomposition}.

\subsection{Analysis of Coupled Iterates $(\rvx_n, \rvy_n)$}\label{appendix:2.2}
In view of Lemma 2.1 (almost sure convergence) in Section 2.1, for large enough $n$, both iterates $\rvx_n, \rvy_n$ are close to the equilibrium $(\rvx^*, \rvy^*)$. In this part, we further decompose \eqref{eqn:TTSA_decomposition_3} in relation to the equilibrium. To do so, we first apply the Taylor expansion to functions $\bar h_1(\rvx_n, \rvy_n)$ and $\bar h_2(\rvx_n, \rvy_n)$ in \eqref{eqn:TTSA_decomposition_3} at $(\rvx^*, \rvy^*)$, which results in
\begin{equation*}
    \bar h_1(\rvx_n, \rvy_n) = \nabla_{\rvx} \bar h_1(\rvx^*,\rvy^*)(\rvx_n - \rvx^*) + \nabla_{\rvy} \bar h_1(\rvx^*,\rvy^*)(\rvy_n - \rvy^*) + O(\|\rvx_n - \rvx^*\|^2 + \|\rvy_n - \rvy^*\|^2),
\end{equation*}
\begin{equation*}
    \bar h_2(\rvx_n, \rvy_n) = \nabla_{\rvx} \bar h_2(\rvx^*,\rvy^*)(\rvx_n - \rvx^*) + \nabla_{\rvy} \bar h_2(\rvx^*,\rvy^*)(\rvy_n - \rvy^*) + O(\|\rvx_n - \rvx^*\|^2 + \|\rvy_n - \rvy^*\|^2).
\end{equation*}
Denote by
\begin{align*}
\rmQ_{11} \triangleq \nabla_{\rvx} \bar h_1(\rvx^*,\rvy^*), ~~~~\rmQ_{12} \triangleq \nabla_{\rvy} \bar h_1(\rvx^*,\rvy^*), ~~~~\rmQ_{21} \triangleq \nabla_{\rvx} \bar h_2(\rvx^*,\rvy^*), ~~~~\rmQ_{22} \triangleq \nabla_{\rvy} \bar h_2(\rvx^*,\rvy^*),
\end{align*}
we can rewrite \eqref{eqn:TTSA_decomposition_3} as
\begin{subequations}
    \begin{equation}\label{eqn:two_timescale_SA_2a}
        \rvx_{n+1} = \rvx_n + \beta_{n+1}\left(\rmQ_{11} (\rvx_n - \rvx^*) + \rmQ_{12}(\rvy_n - \rvy^*) + M^{(\rvx)}_{n+1} + r^{(\rvx,1)}_n + r^{(\rvx,2)}_n +  \rho_n^{(\rvx)}\right),
    \end{equation}
    \begin{equation}\label{eqn:two_timescale_SA_2b}
        \rvy_{n+1} = \rvy_n + \gamma_{n+1}\left(\rmQ_{21}(\rvx_n - \rvx^*) + \rmQ_{22}(\rvy_n - \rvy^*) + M^{(\rvy)}_{n+1} + r^{(\rvy,1)}_n + r^{(\rvy,2)}_n + \rho_n^{(\rvy)}\right),
    \end{equation}
\end{subequations}
where $\rho_n^{(\rvx)} = O(\|\rvx_n - \rvx^*\|^2 + \|\rvy_n - \rvy^*\|^2)$, $\rho_n^{(\rvy)} = O(\|\rvx_n - \rvx^*\|^2 + \|\rvy_n - \rvy^*\|^2)$ are the error terms coming from Taylor expansion. Then, we rewrite iterates $\rvy_n - \rvy^*$ in \eqref{eqn:two_timescale_SA_2b} as
\begin{equation*}
    \rvy_n - \rvy^* = \gamma_{n+1}^{-1}\rmQ_{22}^{-1}(\rvy_{n+1} - \rvy_n) - \rmQ_{22}^{-1}\rmQ_{21}(\rvx_n - \rvx^*) - \rmQ_{22}^{-1}\left(M^{(\rvy)}_{n+1} + r^{(\rvy,1)}_n + r^{(\rvy,2)}_n + \rho_n^{(\rvy)}\right).
\end{equation*}
Substituting the above into the $(\rvy_n - \rvy^*)$ term in \eqref{eqn:two_timescale_SA_2a} yields the following:
\begin{equation}\label{eqn:decompose_rvx_n}
\begin{split}
    \rvx_{n+1} &- \rvx^* =~ \rvx_n - \rvx^* + \beta_{n+1}\bigg(\rmQ_{11} (\rvx_n-\rvx^*) + \gamma_{n+1}^{-1}\rmQ_{12}\rmQ_{22}^{-1}(\rvy_{n+1} - \rvy_n) - \rmQ_{12}\rmQ_{22}^{-1}\rmQ_{21}(\rvx_n - \rvx^*) \\
    &- \rmQ_{12}\rmQ_{22}^{-1}\left(M^{(\rvy)}_{n+1} + r^{(\rvy,1)}_n + r^{(\rvy,2)}_n + \rho_n^{(\rvy)}\right) + M^{(\rvx)}_{n+1} + r^{(\rvx,1)}_n + r^{(\rvx,2)}_n +  \rho_n^{(\rvx)}\bigg) \\
    =&~ \left(\rmI + \beta_{n+1}\rmK_{\rvx}\right) (\rvx_n - \rvx^*) + \beta_{n+1} \left(\gamma_{n+1}^{-1}\rmQ_{12}\rmQ_{22}^{-1}(\rvy_{n+1}-\rvy_n)\right) + \beta_{n+1}\left(M^{(\rvx)}_{n+1} - \rmQ_{12}\rmQ_{22}^{-1}M^{(\rvy)}_{n+1}\right) \\
    &+ \beta_{n+1} \left( r^{(\rvx,1)}_n + r^{(\rvx,2)}_n + \rho_n^{(\rvx)} - \rmQ_{12}\rmQ_{22}^{-1}(r^{(\rvy,1)}_n + r^{(\rvy,2)}_n + \rho_n^{(\rvy)})\right),
\end{split}
\end{equation}
where $\rmK_{\rvx} = \rmQ_{11} - \rmQ_{12}\rmQ_{22}^{-1}\rmQ_{21}$. As we observe from \eqref{eqn:decompose_rvx_n}, iterates $\rvx_{n}$ naturally embed two sequences: 

(i) $\beta_{n+1}\gamma_{n+1}^{-1}\rmQ_{12}\rmQ_{22}^{-1}(\rvy_{n+1}-\rvy_n)$; (ii) $\beta_{n+1}(M_{n+1}^{(\rvx)}-\rmQ_{12}\rmQ_{22}^{-1}M^{(\rvy)}_{n+1})$.

These sequences can be expressed recursively by following similar steps as in \citet{delyon2000stochastic,mokkadem2006convergence,fort2015central}. Specifically, let
\begin{equation}
    u_n \triangleq \sum_{k=1}^n \beta_k, \quad s_n \triangleq \sum_{k=1}^n \gamma_k.
\end{equation}
Then, the sequences (i) and (ii) can be used to drive the following iterations respectively:
\begin{equation}\label{eqn:R_n_rvx}
    R_{n}^{(\rvx)} \triangleq e^{\beta_{n}\rmK_{\rvx}} R_{n-1}^{(\rvx)} + \beta_{n}\gamma_{n}^{-1} \rmQ_{12}\rmQ_{22}^{-1}(\rvy_{n}-\rvy_{n-1}) = \sum_{k=1}^n e^{(u_{n}-u_k)\rmK_{\rvx}}\beta_{k}\gamma_{k}^{-1}\rmQ_{12}\rmQ_{22}^{-1}(\rvy_{k}-\rvy_{k-1}).
\end{equation} 
\begin{equation}\label{eqn:L_n_rvx}
    L_{n}^{(\rvx)} \triangleq e^{\beta_{n}\rmK_{\rvx}} L_{n-1}^{(\rvx)} + \beta_{n} (M^{(\rvx)}_{n} - \rmQ_{12}\rmQ_{22}^{-1}M^{(\rvy)}_{n}) = \sum_{k=1}^n e^{(u_{n}-u_k)\rmK_{\rvx}}\beta_{k}(M^{(\rvx)}_{k} - \rmQ_{12}\rmQ_{22}^{-1}M^{(\rvx)}_{k}),
\end{equation}
The remaining noise in the iterates $\rvx_n$ is defined as $\Delta_{n+1}^{(\rvx)} \triangleq \rvx_{n+1} - \rvx^* - L_{n+1}^{(\rvx)} - R_{n+1}^{(\rvx)}$.

Similarly, for iterates $\rvy_n$ in \eqref{eqn:two_timescale_SA_2b}, we define the following sequences:
\begin{equation}\label{eqn:R_n_rvy}
    R_{n}^{(\rvy)} = e^{\gamma_{n}\rmQ_{22}} R_{n-1}^{(\rvy)} + \gamma_{n} \rmQ_{21}(L_{n-1}^{(\rvx)} + R_{n-1}^{(\rvx)}) = \sum_{k=1}^n e^{(s_n - s_k)\rmQ_{22}} \gamma_k \rmQ_{21} (L_{k-1}^{(\rvx)} + R_{k-1}^{(\rvx)}).
\end{equation}
\begin{equation}\label{eqn:L_n_rvy}
    L_{n}^{(\rvy)} = e^{\gamma_{n}\rmQ_{22}} L_{n-1}^{(\rvy)} + \gamma_{n}M^{(\rvy)}_{n} = \sum_{k=1}^n e^{(s_{n}-s_k)\rmQ_{22}}\gamma_{k}M^{(\rvy)}_{k},
\end{equation}
The remaining noise in iterates $\rvy_n$ is denoted as $\Delta_{n+1}^{(\rvy)} \triangleq \rvy_{n+1} - \rvy^* - L_{n+1}^{(\rvy)} - R_{n+1}^{(\rvy)}$.

In view of Lemmas \ref{lemma:decomposition_Mn} and \ref{lemma:characteristics_decomposition}, we have the following results characterizing the weak convergence of sequences $\beta_n^{-1/2}L_n^{(\rvx)}$ and $\gamma_n^{-1/2}L_n^{(\rvy)}$, which are proved in Appendix \ref{appendix:2.5}.
\begin{lemma}\label{lemma:weak_convergence_Ln}
    For $L_n^{(\rvx)}$ and $L_n^{(\rvy)}$ defined in \eqref{eqn:L_n_rvx} and \eqref{eqn:L_n_rvy}, we have
\begin{equation*}
\begin{pmatrix}
\sqrt{\beta_n^{-1}} L_{n}^{(\rvx)} \\ \sqrt{\gamma_n^{-1}} L_{n}^{(\rvy)} \end{pmatrix} \xrightarrow[n \to \infty]{dist} N\left(0,\begin{pmatrix}
\rmV_{\rvx} & 0 \\ 0 & \rmV_{\rvy}
\end{pmatrix}\right),
\end{equation*}
where \begin{equation*}
\begin{split}
    &\rmV_{\rvx} \!=\! \int_0^{\infty}\! e^{t\left(\rmK_{\rvx}+\frac{\mathds{1}_{\{b=1\}}}{2}\rmI\right)} \rmU_{\rvx} e^{t\left(\rmK_{\rvx}+\frac{\mathds{1}_{\{b=1\}}}{2}\rmI\right)^T} dt,\\
    &\rmV_{\rvy} \!=\! \int_0^{\infty} e^{t\rmQ_{22}} \rmU_{22} e^{t\rmQ_{22}^T} dt,
\end{split}
\end{equation*}
with $\rmU_{\rvx}$ of the form \eqref{eqn:alt_matrix_U_x_form} and $\rmU_{22}$ defined in Lemma \ref{lemma:characteristics_decomposition}.
\end{lemma}
Lemma \ref{lemma:weak_convergence_Ln} confirms that $\beta_n^{-1/2}L_n^{(\rvx)}$ and $\gamma_n^{-1/2}L_n^{(\rvy)}$ weakly converge to Gaussian multivariate distributions, mirroring the weak convergence of $\beta_n^{-1/2}(\rvx_n-\rvx^*)$ and $\gamma_n^{-1/2}(\rvy_n-\rvy^*)$ as described in Theorem 2.2 (Section 2.3). Furthermore, we establish the following results that the sequences $R_n^{(\rvx)}, R_n^{(\rvy)}, \Delta_n^{(\rvx)}$, and $\Delta_n^{(\rvy)}$ decay to zero at rates faster than their respective CLT scales. This is achieved by examining a bounded deterministic sequence that iteratively refines the upper bounds of $ \rvy_n - \rvy^* $ and $\Delta_n^{(\rvy)}$. Detailed insights are provided in Appendix \ref{appendix:2.6}.
\begin{lemma}\label{lemma:final_result_R_n_Delta_n}
We have
\begin{enumerate}
    \item for some constant $c> b/2$, $\|R_n^{(\rvx)}\| = O(n^{-c})$ a.s
    \item $\|R_n^{(\rvy)}\| = O(\sqrt{\beta_n\log u_n})$ a.s.
    \item $\|\Delta_n^{(\rvx)}\| = o(\sqrt{\beta_n})$ a.s.
    \item $\|\Delta_n^{(\rvy)}\| = o(\sqrt{\beta_n})$ a.s.
\end{enumerate}
\end{lemma}
Note that $\sqrt{\beta_n/\gamma_n \log u_n} = O(\sqrt{n^{a-b}\log n}) = o(1)$ since $a - b < 0$ by Assumption \ref{assump:1}. Therefore, from Lemma \ref{lemma:final_result_R_n_Delta_n}, we have $\beta_n^{-1/2}(R_n^{(\rvx)}+\Delta_n^{(\rvx)}) \to 0$ and $\gamma_n^{-1/2}(R_n^{(\rvy)}+\Delta_n^{(\rvy)}) \to 0$ almost surely. By Lemma \ref{lemma:weak_convergence_Ln}, we establish the weak convergence for $\beta_{n}^{-1/2}L_n^{(\rvx)}$ and $\gamma_{n}^{-1/2}L_n^{(\rvy)}$. Together with $\rvx_n - \rvx^* = L_n^{(\rvx)} + R_n^{(\rvx)} + \Delta_n^{(\rvx)}$ and $\rvy_n - \rvy^* = L_n^{(\rvy)} + R_n^{(\rvy)} + \Delta_n^{(\rvy)}$, we show that
\begin{align*}
\beta_{n}^{-1/2}(\rvx_n-\rvx^*) \xrightarrow[n \to \infty]{dist} N(0,\rmV_{\rvx}), \quad \gamma_{n}^{-1/2}(\rvy_n-\rvy^*) \xrightarrow[n \to \infty]{dist} N(0,\rmV_{\rvy}),
\end{align*}
which completes the proof.

\subsection{Proof of Lemma \ref{lemma:decomposition_Mn}}\label{appendix:2.3}
We decompose the covariance form of $M_{n+1}^{(\rvx)}$ and $M_{n+1}^{(\rvy)}$ into several terms using the Poisson equation method \citep{benveniste2012adaptive,fort2015central,chen2020explicit}. The detailed steps are as follows.
\begin{equation}\label{eqn:covariance_rvx}
\begin{split}
        &\E\left[\left.M^{(\rvx)}_{n+1}(M^{(\rvx)}_{n+1})^T\right|\gF_n\right] \\
        =&~ \E[m_1(\rvx_n, \rvy_n, \xi_{n+1})m_1(\rvx_n, \rvy_n, \xi_{n+1})^T|\gF_n] + (\rmP m_1)(\rvx_n, \rvy_n, \xi_n) \left[(\rmP m_1)(\rvx_n, \rvy_n, \xi_n)\right]^T \\
        &- \E[m_1(\rvx_n, \rvy_n, \xi_{n+1})|\gF_n] \left[(\rmP m_1)(\rvx_n, \rvy_n, \xi_n)\right]^T - (\rmP m_1)(\rvx_n, \rvy_n, \xi_n) \E[m_1(\rvx_n, \rvy_n, \xi_{n+1})|\gF_n]^T \\
        =&~\E[m_1(\rvx_n, \rvy_n, \xi_{n+1})m_1(\rvx_n, \rvy_n, \xi_{n+1})^T|\gF_n] -  (\rmP m_1)(\rvx_n, \rvy_n, \xi_n) \left[(\rmP m_1)(\rvx_n, \rvy_n, \xi_n)\right]^T.
\end{split}
\end{equation}
where the second equality is because $(\rmP m_1)(\rvx_n,\rvy_n,\xi_n)$, as defined in \eqref{eqn:def_poisson_operator}, can also be written as 
\begin{equation}
    (\rmP m_1)(\rvx_n,\rvy_n,\xi_{n}) = \E[m_1(\rvx_n,\rvy_n,\xi_{n+1})|\gF_n].
\end{equation}
Similarly, we have
\begin{equation*}
\E\left[\left.M^{(\rvy)}_{n+1}(M^{(\rvy)}_{n+1})^T\right|\gF_n\right] = \E[m_2(\rvx_n, \rvy_n, \xi_{n+1})m_2(\rvx_n, \rvy_n, \xi_{n+1})^T|\gF_n] -  (\rmP m_2)(\rvx_n, \rvy_n, \xi_n) \left[(\rmP m_2)(\rvx_n, \rvy_n, \xi_n)\right]^T,
\end{equation*}
and
\begin{equation*}
\E\left[\left.M^{(\rvx)}_{n+1}(M^{(\rvy)}_{n+1})^T\right|\gF_n\right] = \E[m_1(\rvx_n, \rvy_n, \xi_{n+1})m_2(\rvx_n, \rvy_n, \xi_{n+1})^T|\gF_n] -  (\rmP m_1)(\rvx_n, \rvy_n, \xi_n) \left[(\rmP m_2)(\rvx_n, \rvy_n, \xi_n)\right]^T.
\end{equation*}

We now focus on $\E\left[\left.M^{(\rvx)}_{n+1}(M^{(\rvx)}_{n+1})^T\right|\gF_n\right]$ as an example. The same steps of decomposition can be extrapolated to $\E\left[\left.M^{(\rvy)}_{n+1}(M^{(\rvy)}_{n+1})^T\right|\gF_n\right]$ and $\E\left[\left.M^{(\rvx)}_{n+1}(M^{(\rvy)}_{n+1})^T\right|\gF_n\right]$, and their proofs are omitted in this part to avoid repetition and maintain brevity. Define 
\begin{equation}\label{eqn:def_Gi}
    \phi(\rvx,\rvy,i) \triangleq \sum_{j\in\Xi}\rmP_{i,j}[\rvx,\rvy] m_1(\rvx,\rvy,j)m_1(\rvx,\rvy,j)^T -  (\rmP m_1)(\rvx,\rvy,i) \left[(\rmP m_1)(\rvx,\rvy,i)\right]^T,
\end{equation}
and let its expectation w.r.t the stationary distribution $\vpi[\rvx,\rvy]$ be $\bar \phi(\rvx,\rvy) \triangleq \E_{i\sim \vpi[\rvx,\rvy]}[\phi(\rvx,\rvy,i)]$.
We can construct \textit{another Poisson equation}, i.e.,
\begin{equation*}
\begin{split}
&\E\left[\left.M^{(\rvx)}_{n+1}(M^{(\rvx)}_{n+1})^T\right|\gF_n\right] - \sum_{\xi_n \in \Xi}\pi_{\xi_n}[\rvx_n,\rvy_n]\E\left[\left.M^{(\rvx)}_{n+1}(M^{(\rvx)}_{n+1})^T\right|\gF_n\right] \\
=&~ \phi(\rvx_n,\rvy_n,\xi_{n+1}) - \bar\phi(\rvx_n,\rvy_n) \\
=&~ \varphi(\rvx_n,\rvy_n, \xi_{n+1}) - (\rmP \varphi)(\rvx_n, \rvy_n, \xi_{n+1}),
\end{split}
\end{equation*}
with the matrix-valued function $\varphi: \sR^{d_1} \times \sR^{d_2} \times \Xi \to \R^{d_1 \times d_1}$ as its solution. Then, we have
\begin{equation}\label{eqn:V_1}
\begin{split}
   \phi(\rvx_n,\rvy_n,\xi_{n+1}) =& \underbrace{\bar\phi(\rvx^*,\rvy^*)}_{\rmU_{11}} + \underbrace{\bar\phi(\rvx_n,\rvy_n) - \bar\phi(\rvx^*,\rvy^*)}_{\rmD^{(11)}_{n}} + \underbrace{\varphi(\rvx_n,\rvy_n,\xi_{n+1}) - (\rmP \varphi)(\rvx_n,\rvy_n,\xi_{n})}_{\rmJ^{(11,A)}_{n}} \\
   &+ \underbrace{(\rmP \varphi)(\rvx_n,\rvy_n,\xi_{n}) - (\rmP\varphi)(\rvx_n,\rvy_n,\xi_{n+1})}_{\rmJ^{(11,B)}_{n}},
\end{split}
\end{equation}
where matrix $\rmJ^{(11)}_n$ defined in Lemma \ref{lemma:decomposition_Mn} given by $\rmJ^{(11)}_n \triangleq \rmJ^{(11,A)}_n + \rmJ^{(11,B)}_n$. This completes the proof.    

\subsection{Proof of Lemma \ref{lemma:characteristics_decomposition}}\label{appendix:2.4}
\textbf{Expression of matrices $\rmU_{ij}$.} We first give the exact expression of $\bar\phi(\rvx^*,\rvy^*)$ defined in Appendix \ref{appendix:2.3} in order to derive matrix $\rmU_{ij}$ for $i,j\in\{1,2\}$. Recall the explicit form of the solution to the Poisson equation as studied in \citet[Appendix B.3.1 (B.16)]{delyon2000stochastic} and \citet[Appendix B]{hu2022efficiency}, with $\vpi[\rvx^*,\rvy^*] = \vmu$, we have
\begin{equation*}
\begin{split}
    m_1(\rvx^*,\rvy^*,i) &=  \sum_{j\in\Xi}\sum_{k=0}^{\infty} \left(\rmP[\rvx^*,\rvy^*] - \vone\vmu^T\right)_{(i,j)}^k h_1(\rvx^*,\rvy^*,j) \\
    &= \sum_{j\in\Xi}\sum_{k=0}^{\infty} \left(\rmP^k_{i,j}[\rvx^*,\rvy^*] h_1(\rvx^*,\rvy^*,j) - \underbrace{\bar h_1(\rvx^*,\rvy^*)}_{= \vzero}\right) \\
    &= \E\left[\left.\sum_{k=0}^{\infty} h_1(\rvx^*,\rvy^*,\xi_k)\right|\xi_0 = i\right],
\end{split}
\end{equation*}
where the second equality comes from $(\rmP[\rvx^*,\rvy^*] - \vone\vmu^T)^k = \rmP[\rvx^*,\rvy^*]^k - \vone\vmu^T$ for $k\geq 1$ by induction. The last equality is by rewriting $m_1(\rvx^*,\rvy^*,i)$ into a conditional expectation form on the Markov chain $\{\xi_n\}$ conditioned on $\xi_0 = i$. Similarly, for $(\rmP m_1)(\rvx^*,\rvy^*,i)$, we have
\begin{equation*}
    (\rmP m_1)(\rvx^*,\rvy^*,i) = \E\left[\left.\sum_{k=1}^{\infty} h_1(\rvx^*,\rvy^*,\xi_k)\right|\xi_0 = i\right].
\end{equation*}
Hence, for function $\phi(\rvx,\rvy,i)$ defined in \eqref{eqn:def_Gi}, taking the expectation over $\xi_0 \sim \vmu$, i.e., the underlying Markov chain is in the stationary regime from the beginning, we have
\begin{equation*}
\begin{split}
    \bar\phi(\rvx^*,\rvy^*) &= \sum_{i\in\Xi} \left[\mu_i m_1(\rvx,\rvy,i)m_1(\rvx,\rvy,i)^T - \mu_i(\rmP m_1)(\rvx,\rvy,i) \left[(\rmP m_1)(\rvx,\rvy,i)\right]^T\right], \\
    &= \E_{\vmu}\!\left[\left(\sum_{k=0}^{\infty} h_1(\rvx^*,\rvy^*,\xi_k)\right)\!\!\left(\sum_{k=0}^{\infty} h_1(\rvx^*,\rvy^*,\xi_k)\right)^T\right] \!-\! \E_{\vmu}\!\left[\left(\sum_{k=1}^{\infty} h_1(\rvx^*,\rvy^*,\xi_k)\right)\!\!\left(\sum_{k=1}^{\infty} h_1(\rvx^*,\rvy^*,\xi_k)\right)^T\right] \\
    &= \E_{\vmu}[h_1(\rvx^*,\rvy^*,\xi_0)h_1(\rvx^*,\rvy^*,\xi_0)^T] + \E_{\vmu}\left[h_1(\rvx^*,\rvy^*,\xi_0) \left(\sum_{k=1}^{\infty} h_1(\rvx^*,\rvy^*,\xi_k)\right)^T\right] \\
    &~~~+ \E_{\vmu}\left[ \left(\sum_{k=1}^{\infty} h_1(\rvx^*,\rvy^*,\xi_k)\right)h_1(\rvx^*,\rvy^*,\xi_0)^T\right] \\
    &= \sum_{k=1}^{\infty}\left[\text{Cov}(h_1(\rvx^*,\rvy^*,\xi_0),h_1(\rvx^*,\rvy^*,\xi_k)) \!+\! \text{Cov}(h_1(\rvx^*,\rvy^*,\xi_k),h_1(\rvx^*,\rvy^*,\xi_0))\right] \\
    &~~~+ \text{Cov}(h_1(\rvx^*,\rvy^*,\xi_0),h_1(\rvx^*,\rvy^*,\xi_0)),
\end{split}
\end{equation*}
where
\begin{equation*}
\begin{split}
\text{Cov}(h_1(\rvx^*,\rvy^*,\xi_0), h_1(\rvx^*,\rvy^*,\xi_k)) &\triangleq \E_{\vmu}[h_1(\rvx^*,\rvy^*,\xi_0)h_1(\rvx^*,\rvy^*,\xi_k)^T] - \E_{\vmu}[h_1(\rvx^*,\rvy^*,\xi_0)]\E_{\vmu}[h_1(\rvx^*,\rvy^*,\xi_0)]^T \\
&=\E_{\vmu}[h_1(\rvx^*,\rvy^*,\xi_0)h_1(\rvx^*,\rvy^*,\xi_k)^T] - \underbrace{\bar h_1(\rvx^*,\rvy^*)}_{\vzero}\bar h_1(\rvx^*,\rvy^*)^T \\
&= \E_{\vmu}[h_1(\rvx^*,\rvy^*,\xi_0)h_1(\rvx^*,\rvy^*,\xi_k)^T]
\end{split}
\end{equation*}
is the covariance between $h_1(\rvx^*,\rvy^*,\xi_0)$ and $h_1(\rvx^*,\rvy^*,\xi_k)$ for the Markov chain $\{\xi_n\}$ in the stationary regime. By following steps similar to \citet[Proof of Theorem 6.3.7]{bremaud2013markov} using $h_1(\rvx^*,\rvy^*,\xi)$ as the test function, we get
\begin{equation}
    \bar\phi(\rvx^*,\rvy^*) = \lim_{n\to\infty}\frac{1}{n}\E\left[\left(\sum_{k=0}^{n} h_1(\rvx^*,\rvy^*,\xi_k)\right)\left(\sum_{k=0}^{n} h_1(\rvx^*,\rvy^*,\xi_k)\right)^T\right] = \rmU_{11}.
\end{equation}
We can follow the same procedures as above to derive $\rmU_{12}, \rmU_{21}$ and $\rmU_{22}$.

\textbf{Analysis of matrices $\rmD_n^{(ij)}$.} To analyze $\rmD_n^{(ij)}$, we first discuss the continuity of functions $m_1, m_2$ in \eqref{eqn:sol_Poisson_eq_closed_form}, and $(\rmP m_1)(\rvx,\rvy, \xi), (\rmP m_2)(\rvx,\rvy, \xi)$ defined in \eqref{eqn:def_poisson_operator}. By Assumption \ref{assump:4}, the transition kernel $\rmP[\rvx,\rvy]$ and its corresponding stationary distribution $\vpi[\rvx,\rvy]$ are continuous in $\rvx, \rvy$, and the inverse $(\rmI - \rmP[\rvx,\rvy] + \vone\vpi[\rvx,\rvy]^T)^{-1}$ is well defined and continuous in $\rvx, \rvy$. This, together with the continuous functions $h_1, h_2$ assumed in Assumption \ref{assump:2}, leads to the continuity of functions $m_1(\rvx,\rvy,\xi), m_2(\rvx,\rvy, \xi)$ and $(\rmP m_1)(\rvx,\rvy, \xi), (\rmP m_2)(\rvx,\rvy, \xi)$ in $\rvx \in \sR^{d_1}, \rvy \in \sR^{d_2}$ for any $\xi \in \Xi$. Furthermore, this results in the conclusion that the function $\phi(\rvx,\rvy,i)$ defined in \eqref{eqn:def_Gi} is also continuous in $\rvx \in \sR^{d_1}, \rvy \in \sR^{d_2}$ for any $\xi \in \Xi$. Thus, its mean field $\bar \phi(\rvx,\rvy)$ is continuous in $\rvx \in \sR^{d_1}, \rvy \in \sR^{d_2}$ as well. By the almost sure convergence $\rvx_n \to \rvx^*$ and $\rvy_n \to \rvy^*$, along with the continuity of function $\bar \phi(\rvx,\rvy)$, we have $$\lim_{n\to\infty} \rmD_n^{(11)} = \lim_{n\to\infty} \bar \phi(\rvx_n,\rvy_n) - \bar \phi(\rvx^*,\rvy^*) = 0 \quad a.s.$$
Similarly, we draw the same conclusion for $\rmD_n^{(12)}, \rmD_n^{(21)}$ and $\rmD_n^{(22)}$.

\textbf{Analysis of matrices $\rmJ_n^{(ij)}$.} We still focus on the case where $i=j = 1$, with other cases following the same steps. As demonstrated in \eqref{eqn:V_1}, we can decompose $\rmJ_n^{(11)} = \rmJ_n^{(11,A)} + \rmJ_n^{(11,B)}$, where 
$$\rmJ_n^{(11,A)} = \varphi(\rvx_n,\rvy_n,\xi_{n+1}) - (\rmP \varphi)(\rvx_n,\rvy_n,\xi_{n}), \quad \rmJ_n^{(11,B)} = (\rmP \varphi)(\rvx_n,\rvy_n,\xi_{n}) - (\rmP\varphi)(\rvx_n,\rvy_n,\xi_{n+1}).$$
$\rmJ_n^{(11,A)}$ is a Martingale difference term adapted to $\gF_n$, i.e., $$\E[\rmJ_n^{(11,A)}| \gF_n] = \E[\varphi(\rvx_n,\rvy_n,\xi_{n+1})| \gF_n] - (\rmP \varphi)(\rvx_n,\rvy_n,\xi_{n}) = 0$$
due to the definition of $(\rmP \varphi)(\rvx_n,\rvy_n,\xi_{n})$ as stated in \eqref{eqn:def_poisson_operator}.
Using the Burkholder inequality from Lemma \ref{lemma:Burkholder Inequality} with $p = 1$, for some constant $C_1 > 0$ we get 
\begin{equation}\label{eqn:29}
    \E\left[\left\|\sum_{k=1}^n \rmJ_k^{(11,A)}\right\|\right] \leq C_1 \E\left[\sqrt{\left(\sum_{k=1}^n \left\|\rmJ_k^{(11,A)}\right\|^2\right)}\right].
\end{equation}
By Assumption \ref{assump:5}, iterates $\rvx_n, \rvy_n$ are always within some compact set $\Omega$ such that $\sup_n \left\|\rmJ_n^{(11,A)}\right\| \leq C_{\Omega} < \infty$ for a set-dependent constant $C_{\Omega}$, and thus
\begin{equation}\label{eqn:J_k_11_A}
    \gamma_n C_1 \sqrt{\left(\sum_{k=1}^n \left\|\rmJ_k^{(11,A)}\right\|^2\right)} \leq  C_1 C_{\Omega} \gamma_n \sqrt{n}.
\end{equation}
The last term of \eqref{eqn:J_k_11_A} decreases to zero in $n$ due to $a > 1/2$ in Assumption \ref{assump:1} and is therefore uniformly bounded with respect to $n$.

For $\rmJ_n^{(11,B)}$, we use the Abel transformation in Lemma \ref{lemma:abel_transformation} to obtain
\begin{equation*}
\sum_{k=1}^{n} \rmJ_k^{(11,B)} = \sum_{k=1}^n \left[(\rmP\varphi)(\rvx_k,\rvy_k,\xi_{k-1}) - (\rmP\varphi)(\rvx_{k-1},\rvy_{k-1},\xi_{k-1})\right] + (\rmP \varphi)(\rvx_0,\rvy_0,\xi_{0}) - (\rmP \varphi)(\rvx_n,\rvy_n,\xi_{n}).
\end{equation*}
Following same steps to derive the continuity of function $m_1$ in the previous paragraph, we have that the matrix-valued function $(\rmP \varphi)(\rvx,\rvy,\xi)$ is continuous in $\rvx \in \sR^{d_1},\rvy\in \sR^{d_2}$. Thus, by Assumption \ref{assump:5} that $(\rvx_n,\rvy_n)$ are within some compact set $\Omega$, there exists a constant $L_{\Omega}$ such that 
\begin{equation*}
    \|(\rmP\varphi)(\rvx_k,\rvy_k,\xi_{k-1}) - (\rmP\varphi)(\rvx_{k-1},\rvy_{k-1},\xi_{k-1})\| \leq L_{\Omega} (\| \rvx_k - \rvx_{k-1}\| + \| \rvy_k - \rvy_{k-1}\|) \leq C_{\Omega}' L_{\Omega}(\beta_k + \gamma_k),
\end{equation*}
where $C_{\Omega}' = \max_{(\rvx,\rvy)\in \Omega, \xi \in \Xi} \{h_1(\rvx,\rvy,\xi),h_2(\rvx,\rvy,\xi)\}$. Also, $\|(\rmP \varphi)(\rvx_0,\rvy_0,\xi_{0})\| + \|(\rmP \varphi)(\rvx_n,\rvy_n,\xi_{n})\|$ are upper-bounded by some positive constant $C_{\Omega}''$.
From the above, we have 
\begin{equation}
   \left\|\sum_{k=1}^n \rmJ_k^{(11,B)}\right\| \leq  C_{\Omega}'' + C_{\Omega}'L_{\Omega}\sum_{k=1}^n(\beta_k+\gamma_k) \leq C_{\Omega}'' + C'''_{\Omega}\sum_{k=1}^n \gamma_k
\end{equation}
for some set-dependent constant $C'''_{\Omega} > 0$. Note that 
\begin{equation}\label{eqn:J_k_11_B}
\gamma_n\left\|\sum_{k=1}^n \rmJ_k^{(11,B)}\right\| \leq \gamma_n C_{\Omega}'' + C_{\Omega}''' \gamma_n\sum_{k=1}^n\gamma_k <\gamma_n C_{\Omega}'' + \frac{C_{\Omega}'''}{a} n^{1-2a}
\end{equation}
where the last inequality is from $\sum_{k=1}^n \gamma_k < \frac{1}{a} n^{1-a}$. We observe that \eqref{eqn:J_k_11_B} is decreasing to zero in $n$ due to $a > 1/2$ and is thus uniformly bounded with respect to $n$.

Note that $\rmJ_k^{(11)} = \rmJ_k^{(11,A)} + \rmJ_k^{(11,B)}$, by triangular inequality we have
\begin{equation}\label{eqn:32}
\begin{split}
\gamma_n\E\left[\left\|\sum_{k=1}^n \rmJ_k^{(11)}\right\|\right] &\leq \gamma_n\E\left[\left\|\sum_{k=1}^n \rmJ_k^{(11,A)}\right\|\right] + \gamma_n\E\left[\left\|\sum_{k=1}^n \rmJ_k^{(11,B)}\right\|\right] \\
&\leq \gamma_n C_1 \E\left[\sqrt{\left(\sum_{k=1}^n \left\|\rmJ_k^{(11,A)}\right\|^2\right)}\right] + \gamma_n \E\left[\left\|\sum_{k=1}^n \rmJ_k^{(11,B)}\right\|\right] \\
&= \E\left[\gamma_n C_1\sqrt{\left(\sum_{k=1}^n \left\|\rmJ_k^{(11,A)}\right\|^2\right)} + \gamma_n\left\|\sum_{k=1}^n \rmJ_k^{(11,B)}\right\|\right],
\end{split}
\end{equation}
where the second inequality comes from \eqref{eqn:29}. By \eqref{eqn:J_k_11_A} and \eqref{eqn:J_k_11_B} we know that both terms in the last line of \eqref{eqn:32} are bounded by constants that depend on the set $\Omega$. Therefore, by dominated convergence theorem, taking the limit over the last line of \eqref{eqn:32} gives
\begin{equation*}
\lim_{n\to\infty} \E\!\left[\gamma_n C_1\sqrt{\left(\sum_{k=1}^n \left\|\rmJ_k^{(11,A)}\right\|^2\right)} \!+\! \gamma_n\left\|\sum_{k=1}^n \rmJ_k^{(11,B)}\right\|\right] \!=\! \E\!\left[\lim_{n\to\infty} \gamma_n C_1\sqrt{\left(\sum_{k=1}^n \left\|\rmJ_k^{(11,A)}\right\|^2\right)} \!+\! \gamma_n\left\|\sum_{k=1}^n \rmJ_k^{(11,B)}\right\|\right] \!=\! 0.
\end{equation*}
Therefore, we have 
\begin{equation*}
    \lim_{n\to\infty} \gamma_n \E\left[\left\|\sum_{k=1}^n \rmJ_k^{(11)}\right\|\right] = 0,
\end{equation*}
which completes the proof.    

\subsection{Proof of Lemma \ref{lemma:weak_convergence_Ln}}\label{appendix:2.5}
Define a Martingale $Z^{(n)} = \{Z^{(n)}_k\}_{k\geq 1}$ such that
\begin{equation*}
    Z^{(n)}_k \triangleq \begin{pmatrix}
        \sqrt{\beta_{n}^{-1}}L_n^{(\rvx)} \\ \sqrt{\gamma_{n}^{-1}}L_n^{(\rvy)}
    \end{pmatrix} = \begin{pmatrix}
        \beta_n^{-1/2} e^{u_n \rmK_{\rvx}} & 0 \\ 0 & \gamma_n^{-1/2} e^{s_n\rmQ_{22}}
    \end{pmatrix} \times  \sum_{j=1}^k \begin{pmatrix}
        e^{-u_k\rmK_{\rvx}}\beta_k(M^{(\rvx)}_{k} - \rmQ_{12}\rmQ_{22}^{-1}M^{(\rvy)}_{k}) \\
        e^{-s_k\rmQ_{22}}\gamma_{k}M^{(\rvy)}_{k}
    \end{pmatrix}.
\end{equation*}
Then, the Martingale difference array $Z_k^{(n)} - Z_{k-1}^{(n)}$ becomes
\begin{equation*}
    Z_k^{(n)} - Z_{k-1}^{(n)} = \begin{pmatrix}
        \beta_n^{-1/2} e^{(u_n-u_k) \rmK_{\rvx}}\beta_k(M^{(\rvx)}_{k} - \rmQ_{12}\rmQ_{22}^{-1}M^{(\rvy)}_{k}) \\
        \gamma_n^{-1/2} e^{(s_n-s_k) \rmQ_{22}}\gamma_k M^{(\rvy)}_{k}
    \end{pmatrix}
\end{equation*}
such that
\begin{equation*}
        \sum_{k=1}^n \E\left[(Z_k^{(n)} - Z_{k-1}^{(n)})(Z_k^{(n)} - Z_{k-1}^{(n)})^T | \gF_{k-1}\right]=\begin{pmatrix}
            S^{(11)}_{n} & S^{(12)}_{n} \\ S^{(21)}_{n} & S^{(22)}_{n}
        \end{pmatrix},
\end{equation*}
where, in view of decomposition of $M^{(\rvx)}_{n}$ and $M^{(\rvy)}_{n}$ in Lemma \ref{lemma:decomposition_Mn}, we have
\begin{subequations}
    \begin{equation}
    \begin{split}
        S^{(11)}_{n} = \beta_n^{-1} \sum_{k=1}^n \beta_k^2 e^{(u_n - u_k)\rmK_{\rvx}}&\bigg(\rmU_{11} \!+\! \rmD_k^{(11)} \!+\! \rmJ_{k}^{(11)} \!-\! (\rmU_{12} \!+\! \rmD_k^{(12)} \!+\! \rmJ_{k}^{(12)})(\rmQ_{12}\rmQ_{22}^{-1})^T \\
        &+ \rmQ_{12}\rmQ_{22}^{-1}(\rmU_{22} + \rmD_k^{(22)} + \rmJ_{k}^{(22)})(\rmQ_{12}\rmQ_{22}^{-1})^T \\
        &- \rmQ_{12}\rmQ_{22}^{-1}(\rmU_{21} + \rmD_k^{(21)} + \rmJ_{k}^{(21)}) \bigg) e^{(u_n - u_k)\rmK_{\rvx}^T},
    \end{split}
    \end{equation}
    \begin{equation}
        S^{(12)}_{n} = \beta_n^{-1/2}\gamma_n^{-1/2}\sum_{k=1}^n \beta_k\gamma_k e^{(u_n-u_k)\rmK_{\rvx}}(\rmU_{12} - \rmQ_{12}\rmQ_{22}^{-1}\rmU_{22})e^{(s_n-s_k)\rmQ_{22}^T},
    \end{equation}
    \begin{equation}
        S_{n}^{(22)} = \gamma_n^{-1} \sum_{k=1}^n \gamma_k^2 e^{(s_n - s_k)\rmQ_{22}}(\rmU_{22} + \rmD_k^{(22)} + \rmJ_{k}^{(22)}) e^{(s_n - s_k)\rmQ_{22}^T},
    \end{equation}
\end{subequations}
and $S_n^{(21)} = (S_n^{(12)})^T$. 

We now focus on $S_{n}^{(11)}$, whose property is given by the following lemma.
\begin{lemma}\label{lemma:S_n_11}
    $\lim_{n\to\infty} S^{(11)}_{n} = \rmV_{\rvx}$, where $\rmV_{\rvx}$ is of the form in Lemma \ref{lemma:weak_convergence_Ln}.
\end{lemma}
\begin{proof}
We rewrite $S_{n}^{(11)}$ into three parts:
\begin{equation}\label{eqn:A_1,n}
\begin{split}
    &S_{n}^{(11)} \\
    &=  \underbrace{\beta_n^{-1} \sum_{k=1}^n\! \bigg(\beta_k^2 e^{(u_n \!-\! u_k)\rmK_{\rvx}}(\rmU_{11} \!+\! \rmQ_{12}\rmQ_{22}^{-1} \rmU_{22} (\rmQ_{12}\rmQ_{22}^{-1})^T \!-\! \rmU_{12}(\rmQ_{12}\rmQ_{22}^{-1})^T \!-\! \rmQ_{12}\rmQ_{22}^{-1}\rmU_{21})e^{(u_n \!-\! u_k)\rmK_{\rvx}^T}\bigg)}_{S_{n}^{(11,A)}} \\
    &+ \underbrace{\beta_n^{-1} \sum_{k=1}^n\! \bigg(\beta_k^2 e^{(u_n \!-\! u_k)\rmK_{\rvx}} (\rmD_k^{(11)} \!+\! \rmQ_{12}\rmQ_{22}^{-1} \rmD_k^{(22)} (\rmQ_{12}\rmQ_{22}^{-1})^T \!-\! \rmD_k^{(12)}(\rmQ_{12}\rmQ_{22}^{-1})^T \!-\! \rmQ_{12}\rmQ_{22}^{-1}\rmD_k^{(21)})e^{(u_n \!-\! u_k)\rmK_{\rvx}^T}\bigg)}_{S_{n}^{(11,B)}} \\
    &+ \underbrace{\beta_n^{-1} \sum_{k=1}^n\! \bigg(\beta_k^2 e^{(u_n \!-\! u_k)\rmK_{\rvx}} (\rmJ_k^{(11)} \!+\! \rmQ_{12}\rmQ_{22}^{-1} \rmJ_k^{(22)} (\rmQ_{12}\rmQ_{22}^{-1})^T \!-\! \rmJ_k^{(12)}(\rmQ_{12}\rmQ_{22}^{-1})^T \!-\! \rmQ_{12}\rmQ_{22}^{-1}\rmJ_k^{(21)})e^{(u_n \!-\! u_k)\rmK_{\rvx}^T}\bigg).}_{S_{n}^{(11,C)}}
\end{split}
\end{equation}

We aim to demonstrate that $$\lim_{n\to\infty} S^{(11,A)}_{n} = \rmV_{\rvx}, \quad \lim_{n\to\infty} \|S^{(11,B)}_{n}\| = 0, \quad \lim_{n\to\infty} \|S^{(11,C)}_{n}\| = 0.$$

From Lemma \ref{lemma:matrix_norm_inequality}, we have for some $c, T>0$ such that
\begin{equation*}
\begin{split}
    \|S^{(11,B)}_{n}\| \leq \beta_n^{-1} \sum_{k=1}^n& \bigg\|\rmD_k^{(11)} + \rmQ_{12}\rmQ_{22}^{-1} \rmD_k^{(22)} (\rmQ_{12}\rmQ_{22}^{-1})^T - \rmD_k^{(12)}(\rmQ_{12}\rmQ_{22}^{-1})^T - \rmQ_{12}\rmQ_{22}^{-1}\rmD_k^{(21)}\bigg\| \beta_k^2 c^2e^{-2T(u_n - u_k)}.
\end{split}
\end{equation*}
Applying Lemma \ref{lemma:upperbound_of_sequence_x}, together with $\rmD^{(ij)}_n \to 0$ a.s. in Lemma \ref{lemma:characteristics_decomposition}, gives
\begin{equation*}
\begin{split}
\limsup_n \|S^{(11,B)}_{n}\| \leq \frac{1}{C(b,p)} \limsup_n \|\rmD_k^{(11)} \!+\! \rmQ_{12}\rmQ_{22}^{-1} \rmD_k^{(22)} (\rmQ_{12}\rmQ_{22}^{-1})^T \!-\! \rmD_k^{(12)}(\rmQ_{12}\rmQ_{22}^{-1})^T \!-\! \rmQ_{12}\rmQ_{22}^{-1}\rmD_k^{(21)}\| = 0,
\end{split}
\end{equation*}
for some constant $C(b,p)>0$ defined in Lemma \ref{lemma:upperbound_of_sequence_x}.

We now consider $S^{(11,C)}_{n}$. Set 
\begin{align*}
    \psi_n \triangleq \sum_{k=1}^n \left( \rmJ_k^{(11)} \!+\! \rmQ_{12}\rmQ_{22}^{-1} \rmJ_k^{(22)} (\rmQ_{12}\rmQ_{22}^{-1})^T \!-\! \rmJ_k^{(12)}(\rmQ_{12}\rmQ_{22}^{-1})^T \!-\! \rmQ_{12}\rmQ_{22}^{-1}\rmJ_k^{(21)}\right),
\end{align*}
we can rewrite $S^{(11,C)}_{n}$ as 
\begin{equation*}
    S^{(11,C)}_{n} = \beta_n^{-1}\sum_{k=1}^n \beta_k^2 e^{(u_n-u_k) \rmK_{\rvx}} (\psi_k - \psi_{k-1})e^{(u_n-u_k) (\rmK_{\rvx})^T}.
\end{equation*}
By the Abel transformation in Lemma \ref{lemma:abel_transformation}, we have
\begin{equation}\label{eqn:94}
\begin{split}
    S^{(11,C)}_{n} = \beta_n \psi_n +  \beta_n^{-1}\sum_{k=1}^{n-1} &\left[\beta_k^2 e^{(u_n - u_k)\rmK_{\rvx}} \psi_k e^{(u_n - u_k)\rmK_{\rvx}^T} - \beta_{k+1}^2 e^{(u_n - u_{k+1})\rmK_{\rvx}} \psi_k e^{(u_n - u_{k+1})\rmK_{\rvx}^T}\right].
\end{split}
\end{equation}
We know from Lemma \ref{lemma:characteristics_decomposition} that $\beta_n \psi_n \to 0$ a.s. because $\psi_n = o(\gamma_n^{-1})$ such that $\beta_n \psi_n = o(\beta_n/\gamma_n)$. Besides,
\begin{equation*}
\begin{split}
    \| \beta_k e^{(u_n-u_k)\rmK_{\rvx}} - \beta_{k+1} e^{(u_n-u_{k+1})\rmK_{\rvx}}\| &= \|(\beta_k - \beta_{k+1})e^{(u_n-u_k)\rmK_{\rvx}} + \beta_{k+1}e^{(u_n-u_k)\rmK_{\rvx}}(\rmI - e^{-\beta_{k+1}\rmK_{\rvx}})\| \\
    &\leq  C_1 \beta_k^2 e^{-(u_n - u_k)T},
\end{split}
\end{equation*}
for some constant $C_1 > 0$ because $\beta_n - \beta_{n+1} \leq C_2\beta_n^2$ and $\|\rmI - e^{-\beta_{k+1}\rmK_{\rvx}}\| \leq C_3\beta_{k+1}$ for some $C_2, C_3 > 0$. Moreover, for some $C_4 > 0$,
\begin{equation*}
\begin{split}
    \| \beta_k e^{(u_n-u_k)\rmK_{\rvx}}\| + \|\beta_{k+1} e^{(u_n-u_{k+1})\rmK_{\rvx}}\| &\leq \beta_k \|e^{(u_n-u_k)\rmK_{\rvx}}\| + \beta_{k}\|e^{(u_n-u_{k})\rmK_{\rvx}}\|\cdot\|e^{-\beta_{k+1}\rmK_{\rvx}}\| \\ 
    &\leq C_4 \beta_k e^{-(u_n-u_k)T}.
\end{split}
\end{equation*}
Using Lemma \ref{lemma:matrix_inequality} on \eqref{eqn:94} gives 
\begin{equation*}
    \left\|S^{(11,C)}_{n}\right\| \leq C_1C_4 \beta_n^{-1}\sum_{k=1}^{n-1} \beta_k^2 e^{-2(u_n-u_k)T} \|\beta_k \psi_k\| + \| \beta_n \psi_n\|.
\end{equation*}
Applying Lemma \ref{lemma:upperbound_of_sequence_x} again gives
\begin{equation*}
    \limsup_{n} \left\|S^{(11,C)}_{n}\right\| \leq C_5\limsup_n \|\beta_n \psi_n\| = 0,
\end{equation*}
for some constant $C_5 > 0$. 

We provide an existing lemma below for the term $S^{(11,A)}_{n}$.
\begin{lemma}[\cite{mokkadem2005compact} Lemma 4]\label{lemma:deterministic_convergence}
For a sequence with decreasing step size $\beta_n = (n+1)^{-b}$ for $b \in (1/2,1]$, $u_n = \sum_{k=1}^n \beta_k$, a positive semi-definite matrix $\rmU$ and a Hurwitz matrix $\rmQ$, which is given by
$$\beta_n^{-1} \sum_{k=1}^n \beta_n^2 e^{(u_n - u_k)\rmQ} \rmU e^{(u_n - u_k)\rmQ^T},$$
we have 
\begin{equation*}
    \lim_{n\to\infty} \beta_n^{-1} \sum_{k=1}^n \beta_n^2 e^{(u_n - u_k)\rmQ} \rmU e^{(u_n - u_k)\rmQ^T} = \rmV
\end{equation*}
where $\rmV$ is the solution of the Lyapunov equation 
$$\left(\rmQ + \frac{\mathds{1}_{\{b = 1\}}}{2}\rmI\right)\rmV + \rmV\left(\rmQ^T + \frac{\mathds{1}_{\{b = 1\}}}{2}\rmI\right) + \rmU = 0.$$
\end{lemma}
Recall in \eqref{eqn:alt_matrix_U_x_form} that $\rmU_{\rvx} = \rmU_{11} \!+\! \rmQ_{12}\rmQ_{22}^{-1} \rmU_{22} (\rmQ_{12}\rmQ_{22}^{-1})^T \!-\! \rmU_{12}(\rmQ_{12}\rmQ_{22}^{-1})^T \!-\! \rmQ_{12}\rmQ_{22}^{-1}\rmU_{21}$. Then, we rewrite $S^{(11,A)}_{n}$, defined in \eqref{eqn:A_1,n}, as $$S^{(11,A)}_{n} = \beta_n^{-1} \sum_{k=1}^n\! \beta_k^2 e^{(u_n \!-\! u_k)\rmK_{\rvx}}\rmU_{\rvx}e^{(u_n \!-\! u_k)\rmK_{\rvx}^T}$$ 
Therefore, $\lim_{n\to\infty} S^{(11,A)}_{n} = \rmV_{\rvx}$ is a direct application of Lemma \ref{lemma:deterministic_convergence}, where $\rmV_{\rvx}$ is the solution to the following Lyapunov equation
$$\left(\rmK_{\rvx} + \frac{\mathds{1}_{\{b = 1\}}}{2}\rmI\right)\rmV_{\rvx} + \rmV_{\rvx}\left(\rmK_{\rvx}^T + \frac{\mathds{1}_{\{b = 1\}}}{2}\rmI\right) + \rmU_{\rvx} = 0.$$
Together with Lemma \ref{lemma:closed_form_solution_lyapunov_equation}, we show the closed form $\rmV_{\rvx}$ in Lemma \ref{lemma:weak_convergence_Ln}.
\end{proof}
By repeating the same process as in the demonstration of Lemma \ref{lemma:S_n_11}, we conclude that $\lim_{n\to\infty} S^{(22)}_{n} = \rmV_{\rvy}$.

Moreover, the property of $S^{(12)}_{n}$ is given as follows.
\begin{lemma}
    $\lim_{n\to\infty} S^{(12)}_{n} = 0$.
\end{lemma}
\begin{proof}
Note that
\begin{equation*}
\begin{split}
    \left\| S_{n}^{(12)}\right\| &= O\left(\beta_{n}^{-1/2}\gamma_n^{-1/2}\sum_{k=1}^n \beta_k\gamma_k \|e^{(u_n-u_k)\rmK_{\rvx}}\|\|e^{(s_n-s_k)\rmQ_{22}^T}\|\right) \\
    &= O\left(\beta_{n}^{-1/2}\gamma_n^{-1/2} \sum_{k=1}^n \beta_k\gamma_k e^{-(u_n-u_k)T}e^{-(s_n-s_k)T'}\right) \\
    &= O\left(\beta_{n}^{-1/2}\gamma_n^{-1/2} \sum_{k=1}^n \beta_k\gamma_k e^{-(s_n-s_k)T'}\right)
\end{split}
\end{equation*}
for some $T, T' > 0$, where the second equality is from Lemma \ref{lemma:matrix_norm_inequality} and the third equality comes from $e^{-(u_n-u_k)T} \leq 1$. Then, we use Lemma \ref{lemma:upperbound_of_sequence_x} with $p=0$ to obtain 
\begin{equation}\label{eqn:74}
    \sum_{k=1}^n \beta_k\gamma_k e^{-(s_n-s_k)T'} = O(\beta_n)
\end{equation}
where $\beta_n /\beta_{n+1} = (1+1/n)^{b} = 1 + O(1/n) = 1 + o(\gamma_n)$ satisfies the condition in Lemma \ref{lemma:upperbound_of_sequence_x}.
Additionally, since $\beta_n = o(\gamma_n)$, we have
\begin{equation*}
    \beta_n^{-1/2}\gamma_{n}^{-1/2}\sum_{k=1}^n \beta_k\gamma_k^{-1/2}\gamma_k^{3/2} e^{-(s_n-s_k)T'} =O(\beta_n^{1/2}\gamma_n^{-1/2}) = o(1).
\end{equation*}
Then, it follows that $\lim_{n\to\infty} S^{(12)}_{n} = 0$.
\end{proof}

Consequently, we obtain 
\begin{equation*}
\lim_{n\to\infty}\sum_{k=1}^n \E\left[(Z_k^{(n)} - Z_{k-1}^{(n)})(Z_k^{(n)} - Z_{k-1}^{(n)})^T | \gF_{k-1}\right] = \begin{pmatrix}
    \rmV_{\rvx} & 0 \\ 0 & \rmV_{\rvy}
\end{pmatrix}.
\end{equation*}

The last part of this proof is to verify the conditions of the Martingale CLT in Theorem \ref{theorem:clt_martingale}. For some $\tau > 0$, we have
\begin{equation}\label{eqn:quadratic_variance_rate}
\begin{split}
    &\sum_{k=1}^n \E\left[\|Z_{k}^{(n)} - Z_{k-1}^{(n)}\|^{2+\tau} | \gF_{k-1}\right] \\
    &= O\left(\beta_n^{-(1+\frac{\tau}{2})}\sum_{k=1}^n \beta_k^{2+\frac{\tau}{2}} \beta_k^{\frac{\tau}{2}} e^{-(2+\tau)(u_n-u_k)T} + \gamma_n^{-(1+\frac{\tau}{2})}\sum_{k=1}^n \gamma_k^{2+\frac{\tau}{2}}\gamma_k^{\frac{\tau}{2}} e^{-(2+\tau)(s_n-s_k)T'}\right) \\
    &= O\left(\beta_n^{\frac{\tau}{2}} + \gamma_n^{\frac{\tau}{2}}\right)
\end{split}
\end{equation}
where the last equality comes from Lemma \ref{lemma:upperbound_of_sequence_x}. Since \eqref{eqn:quadratic_variance_rate} also holds for $\tau = 0$, we have
\begin{equation*}
    \sum_{k=1}^n \E\left[\|Z_{k}^{(n)} - Z_{k-1}^{(n)}\|^{2} | \gF_{k-1}\right] = O(1) < \infty.
\end{equation*}
Therefore, all the conditions in Theorem \ref{theorem:clt_martingale} are satisfied and its application then gives
\begin{equation}\label{eqn:CLT_Z_n}
    Z^{(n)} = \begin{pmatrix}
        \sqrt{\beta_n^{-1}} L_{n}^{(\vtheta)} \\ \sqrt{\gamma_n^{-1}} L_{n}^{(\rvx)} \end{pmatrix} \xrightarrow[n \to \infty]{dist} N\left(0,\begin{pmatrix}
    \rmV_{\rvx} & 0 \\ 0 & \rmV_{\rvy}
\end{pmatrix}\right).
\end{equation}
This completes the proof.

\subsection{Upper Bounds of $R_n^{(\rvx)}, R_n^{(\rvy)}, \Delta_n^{(\rvx)}$, and $\Delta_n^{(\rvy)}$ Towards Lemma \ref{lemma:final_result_R_n_Delta_n}}\label{appendix:2.6}
In this part, we aim to show that $R_n^{(\rvx)}, R_n^{(\rvy)}, \Delta_n^{(\rvx)}$, and $\Delta_n^{(\rvy)}$ decrease to zero faster than the CLT scaling factor, and are thus not present in the final CLT results. To proceed with the analysis, we provide the additional lemma to get the tighter upper bounds of $\|L_n^{(\rvx)}\|$ and $\|L_n^{(\rvy)}\|$ as follows, which is useful in deriving the tight upper bounds of $R_n^{(\rvx)}, R_n^{(\rvy)}, \Delta_n^{(\rvx)}$, and $\Delta_n^{(\rvy)}$ later in Lemmas \ref{lemma:upper_bound_R_n} and \ref{lemma:upper_bound_Delta_n}.
\begin{lemma}\label{lemma:tighter_upper_bound_Ln}
For $L_n^{(\rvx)}$ and $L_n^{(\rvy)}$ defined in \eqref{eqn:L_n_rvx} and \eqref{eqn:L_n_rvy}, we further have
    \begin{equation*}
        \|L^{(\rvx)}_n \| = O\left(\sqrt{\beta_n \log(u_n)}\right) \quad a.s.
    \end{equation*}
    \begin{equation*}
        \|L^{(\rvy)}_n \| = O\left(\sqrt{\gamma_n \log(s_n)}\right) \quad a.s.
    \end{equation*}
\end{lemma}
\begin{proof}
 This proof follows from \citet[Lemma 1]{pelletier1998almost}. We only need the special case of \citet[Lemma 1]{pelletier1998almost} that fits our scenario, i.e., we let the two types of step sizes therein be the same. For self-contained purposes, we attach the following lemma.
\begin{lemma}[\cite{pelletier1998almost} Lemma 1]\label{lemma:tight_upperbound_Ln}
    Consider a sequence 
    $$L_{n+1} = e^{u_n\rmQ} \sum_{k=1}^n e^{-u_k\rmQ}\beta_k M_{k+1},$$
    where $\beta_n = (n+1)^{-b}$, $1/2<b\leq 1$, $u_n = \sum_{k=1}^n \beta_k$, matrix $\rmQ$ is Hurwitz, and $\{M_{n}\}$ is a Martingale difference sequence adapted to the filtration $\gF$. Almost surely, $\limsup_n \E[\|M_{n+1}\|^2 | \gF_n] \leq M^2$ and there exists $\tau \in (0,2)$, $b(2+\tau) > 2$, such that $\sup_n \E[\|M_{n+1}\|^{2+\tau}|\gF_n] < \infty$. Then, almost surely,
    \begin{equation}
        \limsup_n \frac{\|L_n\|}{\sqrt{\beta_n \log(u_n)}} \leq C_M,
    \end{equation}
    where $C_M$ is a constant dependent on $M$.
\end{lemma}
By Assumption \ref{assump:5}, the iterates $(\vtheta_n,\rvx_n)$ are bounded within a compact subset $\Omega$. Recall the forms of $M^{(\rvx)}_{n+1}, M^{(\rvy)}_{n+1}$ defined in \eqref{eqn:TTSA_decomposition_2}, they comprise the functions $m_1(\rvx_n,\rvy_n,i), m_2(\rvx_n,\rvy_n,i)$ and $(\rmP m_1)(\rvx_n,\rvy_n,i), (\rmP m_2)(\rvx_n,\rvy_n,i)$, which in turn include the function $h_1(\rvx_n,\rvy_n,i), h_2(\rvx_n,\rvy_n,i)$. We know that $h_1(\rvx_n,\rvy_n,i), h_2(\rvx_n,\rvy_n,i)$ are bounded for $(\rvx_n,\rvy_n)$ within some compact set $\Omega$. Thus, $M^{(\rvx)}_{n+1}, M^{(\rvy)}_{n+1}$ are bounded (because of finite state space $\Xi$) and we denote by
$C^{(\rvx)}_{\Omega}$ and $C^{(\rvy)}_{\Omega}$ as their corresponding upper bounds, i.e., $\E[\|M^{(\rvx)}_{n+1}\|^2| \gF_n] \leq C^{(\rvx)}_{\Omega}$ and $\E[\|M^{(\rvx)}_{n+1}\|^2| \gF_n] \leq C^{(\rvy)}_{\Omega}$. Thus, by the application of Lemma \ref{lemma:tight_upperbound_Ln}, we have
\begin{equation}
    \limsup_n \frac{\|L_n^{(\rvx)}\|}{\sqrt{\beta_n \log(u_n)}} \leq C^{(\rvx)}_{\Omega}, \quad \limsup_n \frac{\|L_n^{(\rvy)}\|}{\sqrt{\gamma_n \log(s_n)}} \leq C^{(\rvy)}_{\Omega},
\end{equation}
such that almost surely, $\|L_n^{(\rvx)}\| = O(\sqrt{\beta_n\log(u_n)})$ and $\|L_n^{(\rvy)}\| = O(\sqrt{\gamma_n\log(s_n)})$, which completes the proof.   
\end{proof}

Now, we present the following condition on any given real-valued deterministic sequence $\{\omega_n\}$ in a similar vein as in \citet[Definition 2]{mokkadem2006convergence}.
\begin{enumerate}[label=(C)., ref=C]
        \item Let $\{\omega_n\}$ be a positive real-valued and uniformly bounded deterministic sequence. Moreover, $\{\omega_n\}$ satisfies $\displaystyle \frac{\omega_n}{\omega_{n+1}} = 1 + o(\gamma_n).$ \label{definition:omega_n}
    \end{enumerate}
In what follows, choices of sequences satisfying Condition \ref{definition:omega_n} will be employed towards proving Lemma \ref{lemma:final_result_R_n_Delta_n}. We explain how to derive the upper bounds for $R_n^{(\rvx)}, R_n^{(\rvy)}, \Delta_n^{(\rvx)}$, and $\Delta_n^{(\rvy)}$, while the main difficulty in the procedure is to derive the upper bounds of $\Delta_n^{(\rvx)}, \Delta_n^{(\rvy)}$ because we have to deal with the additional noise terms $r_n^{(\rvx,2)}, r_n^{(\rvy,2)}$ therein, which arise from the decomposition of Markovian noise. 

First of all, by almost sure convergence, we have $\|\rvy_n - \rvy^*\| = o(1)$, the only upper bound of $\rvy_n - \rvy^*$ for us. Letting $\omega_n \equiv 1$ is obviously one of the choices of $\{\omega_n\}$ that satisfy Condition \ref{definition:omega_n}. Thus, setting $\| \rvy_n - \rvy^*\| = O(\omega_n)$ allows us to present the initial upper bounds for $R_n^{(\rvx)}, R_n^{(\rvy)}$, which involve $\omega_n$, as indicated in Lemma \ref{lemma:upper_bound_R_n}. 
\begin{lemma}\label{lemma:upper_bound_R_n}
    Suppose there exists a nonrandom sequence $\{\omega_n\}$ satisfying Condition \ref{definition:omega_n} such that $\|\rvy_n - \rvy^*\| = O(\omega_n)$ a.s. Then, with $R_n^{(\rvx)}$, $R_n^{(\rvy)}$ defined in \eqref{eqn:R_n_rvx} and \eqref{eqn:R_n_rvy}, for some $s > 1/2$, we have 
    \begin{equation*}
        \|R_n^{(\rvx)}\| = O(\beta_n\gamma_n^{-1}w_n + n^{-s}) \quad a.s.
    \end{equation*}
    \begin{equation*}
        \|R_n^{(\rvy)}\| = O\left(\beta_n\gamma_n^{-1}w_n + \sqrt{\beta_n\log(u_n)}\right) \quad a.s.
    \end{equation*}
\end{lemma}
For the proof of Lemma \ref{lemma:upper_bound_R_n} we refer the reader to \citet[Lemma 5]{mokkadem2006convergence}, which applies directly since both sequences $R_n^{(\rvx)}$ and $R_n^{(\rvy)}$, as defined in \eqref{eqn:R_n_rvx} and \eqref{eqn:R_n_rvy}, have the same form as in \citet{mokkadem2006convergence} and do not include additional terms $r_n^{(\rvx,1)}, r_n^{(\rvx,2)}$, $r_n^{(\rvy,1)}, r_n^{(\rvy,2)}$ arising from the Markovian noise.

Lemma \ref{lemma:upper_bound_R_n} implies that $\|R_n^{(\rvx)}\| = o(1)$ and $\|R_n^{(\rvy)}\| = o(1)$ since $\beta_n\gamma_n^{-1} \to 0$ and $\omega_n$ is uniformly bounded in Condition \ref{definition:omega_n}. Since $\|\rvy_n - \rvy^*\| = o(1)$ by almost sure convergence, Lemma \ref{lemma:tighter_upper_bound_Ln} and Lemma \ref{lemma:upper_bound_R_n} indicate that $L_n^{(\rvy)} = o(1), R_n^{(\rvy)} = o(1)$, we thus have $\Delta_n^{(\rvy)} = o(1)$. So, in addition to set $\|\rvy_n - \rvy^*\| = O(\omega_n)$, we also let $\Delta_n^{(\rvy)} = O(\eta_n)$ for some sequence $\{\eta_n\}$ satisfying Condition \ref{definition:omega_n}. Then, we need to characterize the exact forms of $\Delta_n^{(\rvx)}$, $\Delta_n^{(\rvy)}$.

By substituting  \eqref{eqn:R_n_rvx} and \eqref{eqn:L_n_rvx} in \eqref{eqn:decompose_rvx_n}, along with the definition $\Delta_{n+1}^{(\rvx)} \triangleq \rvx_{n+1} - \rvx^* - L_{n+1}^{(\rvx)} - R_{n+1}^{(\rvx)}$, we obtain $\Delta_{n+1}^{(\rvx)}$ as follows.
\begin{equation}\label{eqn:delta_n_rvx}
\begin{split}
\Delta_{n+1}^{(\rvx)} &= (\rmI + \beta_{n+1}\rmK_{\rvx})(\rvx_n - \rvx^*) + \beta_{n+1} \left( r^{(\rvx,1)}_n + r^{(\rvx,2)}_n + \rho_n^{(\rvx)} - \rmQ_{12}\rmQ_{22}^{-1}(r^{(\rvy,1)}_n + r^{(\rvy,2)}_n + \rho_n^{(\rvy)})\right) \\
&~~~- e^{\beta_{n+1}\rmK_{\rvx}}L_n^{(\rvx)} - e^{\beta_{n+1}\rmK_{\rvx}}R_n^{(\rvx)} \\
&= (\rmI + \beta_{n+1}\rmK_{\rvx})(\rvx_n - \rvx^*) + \beta_{n+1} \left( r^{(\rvx,1)}_n + r^{(\rvx,2)}_n + \rho_n^{(\rvx)} - \rmQ_{12}\rmQ_{22}^{-1}(r^{(\rvy,1)}_n + r^{(\rvy,2)}_n + \rho_n^{(\rvy)})\right) \\
&~~~- (\rmI + \beta_{n+1}\rmK_{\rvx} + O(\beta_{n+1}^2))L_n^{(\rvx)} - (\rmI + \beta_{n+1}\rmK_{\rvx} + O(\beta_{n+1}^2))R_n^{(\rvx)} \\
&= (\rmI + \beta_{n+1}\rmK_{\rvx})\Delta_{n}^{(\rvx)} + \beta_{n+1} \left( r^{(\rvx,1)}_n + r^{(\rvx,2)}_n + \rho_n^{(\rvx)} - \rmQ_{12}\rmQ_{22}^{-1}(r^{(\rvy,1)}_n + r^{(\rvy,2)}_n + \rho_n^{(\rvy)})\right) \\
&~~~+ O(\beta_{n+1}^2)\left(L_n^{(\rvx)} + R_n^{(\rvx)}\right),
\end{split}
\end{equation}
where the third equality is by using the Taylor expansion $e^{\beta_{n+1}\rmK_{\rvx}} = \rmI + \beta_{n+1}\rmK_{\rvx} + O(\beta_{n+1}^2)$, and the fourth equality stems from the definition $\Delta_n^{(\rvx)} = \rvx_n - \rvx^* - L_n^{(\rvx)} - R_n^{(\rvx)}$. 

Similarly, for $\Delta_{n+1}^{(\rvy)}$, we have
\begin{equation}\label{eqn:delta_n_rvy}
\begin{split}
    \Delta_{n+1}^{(\rvy)} &\triangleq \rvy_{n+1} - \rvy^* - L_{n+1}^{(\rvy)} - R_{n+1}^{(\rvy)} \\
    &= (\rmI + \gamma_{n+1}\rmQ_{22})(\rvy_n - \rvy^*) + \gamma_{n+1}\left(r^{(\rvy,1)}_n + r^{(\rvy,2)}_n + \rho_n^{(\rvy)}\right) + \gamma_{n+1}\rmQ_{21}(\rvx_{n} - \rvx^*) \\
    &~~~- e^{\gamma_{n+1}\rmQ_{22}}L_{n}^{(\rvy)} - e^{\gamma_{n+1}\rmQ_{22}}R_{n}^{(\rvy)} - \gamma_{n+1}\rmQ_{21}(L_n^{(\rvx)}+ R_n^{(\rvx)}) \\
    &= (\rmI + \gamma_{n+1}\rmQ_{22})(\rvy_n - \rvy^*) + \gamma_{n+1}\left(r^{(\rvy,1)}_n + r^{(\rvy,2)}_n + \rho_n^{(\rvy)}\right) + \gamma_{n+1}\rmQ_{21}\Delta_{n}^{(\rvx)} \\
    &~~~- (\rmI + \gamma_{n+1}\rmQ_{22} + O(\gamma_{n+1}^2))L_{n}^{(\rvy)} - (\rmI + \gamma_{n+1}\rmQ_{22} + O(\gamma_{n+1}^2))R_{n}^{(\rvy)} \\ 
    &= (\rmI + \gamma_{n+1}\rmQ_{22})\Delta_{n}^{(\rvy)} + \gamma_{n+1}\left(r^{(\rvy,1)}_n + r^{(\rvy,2)}_n + \rho_n^{(\rvy)}\right) + \gamma_{n+1}\rmQ_{21}\Delta_{n}^{(\rvx)} + O(\gamma_{n+1}^2)\left(L_{n}^{(\rvy)} + R_{n}^{(\rvy)}\right),
\end{split}
\end{equation}
where the third equality is from $e^{\gamma_{n+1}\rmQ_{22}} = \rmI + \gamma_{n+1}\rmQ_{22} + O(\gamma_{n+1}^2)$ and $\Delta_n^{(\rvx)} = \rvx_n - \rvx^* - L_n^{(\rvx)} - R_n^{(\rvx)}$, and the fourth equality is because the definition $\Delta_n^{(\rvy)} = \rvy_n - \rvy^* - L_n^{(\rvy)} - R_n^{(\rvy)}$.

In what follows, we investigate the asymptotic behavior of the terms $r^{(\rvx,1)}_n, r^{(\rvy,2)}_n$ and $r^{(\rvy,1)}_n, r^{(\rvy,2)}_n$ that are part of the sequences $\Delta_n^{(\rvx)}$ and $\Delta_n^{(\rvy)}$, respectively. The results of this analysis will be used in Lemma \ref{lemma:upper_bound_Delta_n} to show the upper bounds of $\Delta_n^{(\rvx)}, \Delta_n^{(\rvy)}$.
\begin{lemma}\label{lemma:property_r_n}
 For $r^{(\rvx,1)}_n,r^{(\rvx,2)}_n, r^{(\rvy,1)}_n, r^{(\rvy,2)}_n$ defined in \eqref{eqn:TTSA_decomposition_2}, the following holds almost surely:
\begin{equation*}
    \|r^{(\rvx,1)}_n\| = O(\gamma_n) = o(\sqrt{\beta_n}), \quad \sup_{n}\left\|\sum_{k=1}^n r^{(\rvx,2)}_k\right\| < \infty,
\end{equation*}
\begin{equation*}
    \|r^{(\rvy,1)}_n\| = O(\gamma_n) = o(\sqrt{\beta_n}), \quad \sup_n\left\|\sum_{k=1}^n r^{(\rvy,2)}_k\right\| <\infty.
\end{equation*}   
\end{lemma}
\begin{proof}
We only prove the result for $r^{(\rvx,1)}_n$ and $r^{(\rvx,2)}_n$, since the result for $r^{(\rvy,1)}_n$ and $r^{(\rvy,2)}_n$ follows similar arguments. Observe that for any compact set $\Omega$ satisfying Assumption \ref{assump:5}, we have
\begin{equation*}
\begin{split}
   r^{(\rvx,1)}_n = &~(\rmP m_1)(\rvx_{n+1}, \rvy_{n+1}, \xi_{n+1}) - (\rmP m_1)(\rvx_{n}, \rvy_{n}, \xi_{n+1}) \\
    \leq&  ~\sum_{j\in\Xi} L_{\Omega} (\|\rvx_{n+1} - \rvx_n\| + \|\rvy_{n+1} - \rvy_n\|) \\
    \leq& ~|\Xi| L_{\Omega} C_{\Omega}( \beta_{n+1}  + \gamma_{n+1})
\end{split}
\end{equation*}
where the first inequality is because $(\rmP m_1)(\rvx, \rvy, \xi)$ is continuous in $\rvx, \rvy$ for any $\xi \in \Xi$, and the fact that any continuous function is locally Lipschitz with a set-dependent Lipschitz constant $L_{\Omega}$. The second inequality is from update rule \eqref{eqn:general_TTSA_form}, Assumption \ref{assump:2}, and $(\rvx_n,\rvy_n) \in \Omega$ for some compact subset $\Omega$ (by Assumption \ref{assump:5}) such that 
$$\max_{(\rvx,\rvy) \in \Omega, \xi \in \Xi}\{\|h_1(\rvx,\rvy,\xi)\|, \|h_2(\rvx,\rvy,\xi)\|\} \leq C_{\Omega},$$
Then, because $a > 1/2 \geq b/2$ by Assumption \ref{assump:1}, we have $\|r^{(\rvx,1)}_n\| = O(\gamma_n) = o(\sqrt{\beta_n})$.

Now, let $\nu_n \triangleq (\rmP m_1)(\rvx_n, \rvy_n, \xi_n)$ such that $r^{(\rvx,2)}_n = \nu_n - \nu_{n+1}$. Note that 
$$\sum_{k=1}^n r^{(\rvx,2)}_k = \nu_1 - \nu_{n+1},$$
and by Assumption \ref{assump:5}, $\|\nu_n\|$ is upper bounded by a constant dependent on the compact set $\Omega$, which leads to
\begin{equation*}
    \sup_n\left\|\sum_{k=1}^n r^{(\rvx,2)}_k\right\| = \sup_n\|\nu_1 - \nu_{n+1}\|  < \infty \quad \text{a.s.}
\end{equation*}
This completes the proof.
\end{proof}

We are now ready to state the lemma for the sequences $\Delta_n^{(\rvx)}, \Delta_n^{(\rvy)}$.
\begin{lemma}\label{lemma:upper_bound_Delta_n}
Suppose that there exist two sequences $\{\omega_n\}$ and $\{\eta_n\}$ satisfying Condition \ref{definition:omega_n} such that $\|\rvy_n - \rvy^*\| = O(\omega_n)$ a.s. and $\|\Delta_n^{(\rvy)}\| = O(\eta_n)$ a.s. We have
\begin{equation*}
    \|\Delta_n^{(\rvx)}\| = O(\beta_n^2\gamma_n^{-2}w_n^2 + \beta_n\gamma_n^{-1}\eta_n) + o(\sqrt{\beta_n}) \quad a.s.
\end{equation*}
\begin{equation*}
    \|\Delta_n^{(\rvy)}\| = O(\beta_n^2\gamma_n^{-2}w_n^2 + \beta_n\gamma_n^{-1}\eta_n) + o(\sqrt{\beta_n}) \quad a.s.
\end{equation*}
\end{lemma}
\begin{proof}
The sequences in $\Delta_n^{(\rvx)}, \Delta_n^{(\rvy)}$ involve additional noise terms $r_n^{(\rvx,2)}, r_n^{(\rvy,2)}$ arising from Markovian noise, causing the challenge of deriving their upper bounds. In this proof we separate out these terms $r_n^{(\rvx,2)}, r_n^{(\rvy,2)}$ from other noise terms, and specifically analyze their asymptotic rates, which contribute to the $o(\sqrt{\beta_n})$ term.

Recall from \eqref{eqn:delta_n_rvx}, \eqref{eqn:delta_n_rvy}, we have
\begin{align*}
&\Delta_{n+1}^{(\rvx)} \!=\! (\rmI \!+\! \beta_{n+1}\rmK_{\rvx})\Delta_{n}^{(\rvx)} \!+\! \beta_{n+1} \left( r^{(\rvx,1)}_n \!+\! r^{(\rvx,2)}_n \!+\! \rho_n^{(\rvx)} \!-\! \rmQ_{12}\rmQ_{22}^{-1}(r^{(\rvy,1)}_n \!+\! r^{(\rvy,2)}_n \!+\!\rho_n^{(\rvy)})\right) \!+\! O(\beta_{n+1}^2)\left(L_n^{(\rvx)} \!+\! R_n^{(\rvx)}\right), \\
&\Delta_{n+1}^{(\rvy)} = (\rmI + \gamma_{n+1}\rmQ_{22})\Delta_{n}^{(\rvy)} + \gamma_{n+1}\left(r^{(\rvy,1)}_n + r^{(\rvy,2)}_n + \rho_n^{(\rvy)}\right) + \gamma_{n+1}\rmQ_{21}\Delta_{n}^{(\rvx)} + O(\gamma_{n+1}^2)\left(L_{n}^{(\rvy)} + R_{n}^{(\rvy)}\right).
\end{align*}
We observe that they include the additional terms $r^{(\rvx,2)}_n, r^{(\rvy,2)}_n$ arising from the decomposition of the Markovian noise, which are missing in \citet[equations (20), (21)]{mokkadem2006convergence}. To deal with this issue, we can decompose $\Delta_{n+1}^{(\rvx)}$ into two parts, i.e., $\Delta_{n+1}^{(\rvx)} = \Delta_{n+1}^{(\rvx,1)} + \Delta_{n+1}^{(\rvx,2)}$, where
\begin{align}
\Delta_{n+1}^{(\rvx,1)} &\triangleq (\rmI + \beta_{n+1}\rmK_{\rvx})\Delta_{n}^{(\rvx,1)} + \beta_{n+1} \left( r^{(\rvx,1)}_n + \rho_n^{(\rvx)} - \rmQ_{12}\rmQ_{22}^{-1}(r^{(\rvy,1)}_n +\rho_n^{(\rvy)})\right) + O(\beta_{n+1}^2)\left(L_n^{(\rvx)} + R_n^{(\rvx)}\right), \label{eqn:Delta_n_x_1}\\
\Delta_{n+1}^{(\rvx,2)} &\triangleq (\rmI + \beta_{n+1}\rmK_{\rvx})\Delta_{n}^{(\rvx,2)} + \beta_{n+1} \left(r^{(\rvx,2)}_n - \rmQ_{12}\rmQ_{22}^{-1}r^{(\rvy,2)}_n\right). \label{eqn:Delta_n_x_2}
\end{align}
Similarly, we can decompose $\Delta_{n+1}^{(\rvy)}$ into two parts, i.e., $\Delta_{n+1}^{(\rvy)} = \Delta_{n+1}^{(\rvy,1)} + \Delta_{n+1}^{(\rvy,2)}$, where
\begin{align}
\Delta_{n+1}^{(\rvy,1)} &\triangleq (\rmI + \gamma_{n+1}\rmQ_{22})\Delta_{n}^{(\rvy,1)} + \gamma_{n+1} \left( r^{(\rvy,1)}_n + \rho_n^{(\rvy)} \right) + \gamma_{n+1}\rmQ_{21}\Delta_{n}^{(\rvx)} + O(\gamma_{n+1}^2)\left(L_{n}^{(\rvy)} + R_{n}^{(\rvy)}\right), \label{eqn:Delta_n_y_1}\\
\Delta_{n+1}^{(\rvy,2)} &\triangleq (\rmI + \gamma_{n+1}\rmQ_{22})\Delta_{n}^{(\rvy,2)} + \gamma_{n+1} r^{(\rvy,2)}_n. \label{eqn:Delta_n_y_2} 
\end{align}

Let us first focus on the terms $\Delta_{n+1}^{(\rvx,2)}$ and $\Delta_{n+1}^{(\rvy,2)}$. In the following, we show that 
\begin{equation}\label{eqn:delta_rvx_2}
    \|\Delta_{n+1}^{(\rvx,2)}\| = o(\sqrt{\beta_n}), \quad \|\Delta_{n+1}^{(\rvy,2)}\| = o(\sqrt{\beta_n}).
\end{equation}
Denote by $$\Phi_{k,n} \triangleq \prod_{j={k+1}}^n (\rmI + \beta_{j}\rmK_{\rvx}), \quad \theta_n \triangleq  \sum_{k=1}^n r^{(\rvx,2)}_k - \rmQ_{12}\rmQ_{22}^{-1}r^{(\rvy,2)}_k,$$
and by convention $\Phi_{n+1,n} = \rmI$, we can rewrite $\Delta_{n+1}^{(\rvx,2)}$ in \eqref{eqn:Delta_n_x_2} as
\begin{equation*}
\Delta_{n+1}^{(\rvx,2)} = \sum_{k=1}^n \Phi_{k,n}\beta_{k+1}\left(\theta_k - \theta_{k-1}\right)
\end{equation*}
because $\theta_k - \theta_{k-1} = r^{(\rvx,2)}_k - \rmQ_{12}\rmQ_{22}^{-1}r^{(\rvy,2)}_k$. By Abel transformation in Lemma \ref{lemma:abel_transformation}, we have
\begin{equation*}
\Delta_{n+1}^{(\rvx,2)} = \beta_{n+1}\theta_n + \sum_{k=1}^{n-1}(\beta_k\Phi_{k,n} - \beta_{k+1}\Phi_{k+1,n})\theta_k.
\end{equation*}
Note that $\beta_{n+1}\theta_n = \beta_{n+1} \left( \sum_{k=1}^n r^{(\rvx,2)}_k\right) - \rmQ_{12}\rmQ_{22}^{-1}\beta_{n+1}\left( \sum_{k=1}^n r^{(\rvy,2)}_k\right)$. By Lemma \ref{lemma:property_r_n}, we have
\begin{equation*}
\|\beta_{n+1}\theta_n\| = O(\beta_{n}) = o(\sqrt{\beta_n}) \quad a.s.
\end{equation*}
such that $\beta_{n+1}\theta_n \to 0$ almost surely. Furthermore,
\begin{equation*}
\begin{split}
\|\beta_k\Phi_{k,n} - \beta_{k+1}\Phi_{k+1,n}\| &\leq \beta_{k+1}\|\Phi_{k,n} - \Phi_{k+1,n}\| + (\beta_k - \beta_{k+1}) \|\Phi_{k,n}\| \\
&\leq \beta_{k+1} \|\Phi_{k+1,n}\|\beta_k\|\rmK_{\rvx}\| + C_6\beta_k^2\|\Phi_{k,n}\| \\
&\leq C_7\beta_k^2 e^{-(u_n - u_k)T}
\end{split}
\end{equation*}
for some constant $T, C_6, C_7 > 0$, where the last inequality is from Lemma \ref{lemma:matrix_norm_inequality} and $\|\Phi_{k+1,n}\| \leq C_8\|\Phi_{k,n}\|$ for some constant $C_8> 0$ that depends on $e^{T}$. Then,
\begin{equation*}
\sum_{k=1}^{n}\|\beta_k\Phi_{k,n} - \beta_{k+1}\Phi_{k+1,n}\| \|\theta_k\| \leq C_8 \sum_{k=1}^{n} \beta_k e^{-(u_n - u_k)T} \|\beta_k\Psi_k\| = O(\|\beta_n\theta_n\|),
\end{equation*}
where the last equality is the application of Lemma \ref{lemma:upperbound_of_sequence_x}. Thus, we have $\|\Delta_{n+1}^{(\rvx,2)}\| = o(\sqrt{\beta_n})$. Repeating the same steps, we get $\|\Delta_{n+1}^{(\rvy,2)}\| = o(\sqrt{\beta_n})$.

We now turn to $\Delta_{n+1}^{(\rvx,1)}$ and $\Delta_{n+1}^{(\rvy,1)}$. As shown in \citet[p.11]{mokkadem2006convergence}, there exist two matrix norms $\|\cdot\|_T$ and $\|\cdot\|_M$ such that for large enough $n$, 
\begin{equation*}
    \| \rmI + \beta_{n+1}\rmK_{\rvx}\|_T \leq 1 - \beta_{n+1}T, \quad \| \rmI + \gamma_{n+1}\rmQ_{22}\|_M \leq 1 - \gamma_{n+1}M,
\end{equation*}
for some $T, M > 0$. The corresponding vector norm $\|\rvv\|_T \triangleq \|[\rvv \cdots \rvv]\|_T$, where $[\rvv \cdots \rvv] \in \sR^{d_1 \times d_1}$, and $\|\rvu\|_M \triangleq \|[\rvu \cdots \rvu]\|_M$, where $[\rvu \cdots \rvu] \in \sR^{d_2 \times d_2}$.

Then, we have
\begin{align*}
\|\Delta_{n+1}^{(\rvx,1)}\|_T &\leq (1-\beta_{n+1}T)\|\Delta_{n}^{(\rvx,1)}\|_T \!+\! \beta_{n+1} \left( \|r^{(\rvx,1)}_n\|_T \!+\! \|\rho_n^{(\rvx)}\|_T \!+\! \|\rmQ_{12}\rmQ_{22}^{-1}(r^{(\rvy,1)}_n \!+\!\rho_n^{(\rvy)})\|_T\right) \\
&~~~+ O(\beta_{n+1}^2)\left(\|L_n^{(\rvx)}\|_T \!+\! \|R_n^{(\rvx)}\|_T\right), \\  
\|\Delta_{n+1}^{(\rvy,1)}\|_M &\leq (1-\gamma_{n+1}M)\|\Delta_{n}^{(\rvy,1)}\|_M \!+\! \gamma_{n+1} \left(\|r^{(\rvy,1)}_n\|_M \!+\! \|\rho_n^{(\rvy)}\|_M\right) + \gamma_{n+1}\|\rmQ_{21}\Delta_n^{(\rvx)}\|_M \\
&~~~+ O(\gamma_{n+1}^2)\left(\|L_n^{(\rvy)}\|_M \!+\! \|R_n^{(\rvy)}\|_M\right).   
\end{align*}

We note that 
\begin{align*}
    \|\rho_n^{(\rvx)}\|_T + \|\rmQ_{12}\rmQ_{22}^{-1} \rho_n^{(\rvy)}\|_T &= O(\|\rho_n^{(\rvx)}\| + \|\rho_n^{(\rvy)}\|) \\
    &= O(\|\rvx_n - \rvx^*\|^2 + \|\rvy_n - \rvy^*\|^2) \\
    &= O(\|L_n^{(\rvx)}\|^2+\|R_n^{(\rvx)}\|^2 + \|\Delta_n^{(\rvx)}\|^2_T + \|L_n^{(\rvy)}\|^2+\|R_n^{(\rvy)}\|^2 + \|\Delta_n^{(\rvy)}\|^2_M).
\end{align*}
where the first and the last equalities are from the equivalence of norms. Similarly,
\begin{align*}
\|\rho_n^{(\rvy)}\|_M &= O(\|\rho_n^{(\rvx)}\| + \|\rho_n^{(\rvy)}\|) = O(\|L_n^{(\rvx)}\|^2+\|R_n^{(\rvx)}\|^2 + \|\Delta_n^{(\rvx)}\|^2_T + \|L_n^{(\rvy)}\|^2+\|R_n^{(\rvy)}\|^2 + \|\Delta_n^{(\rvy)}\|^2_M).
\end{align*}

Also note that $\|\Delta_n^{(\rvx)}\|_T^2 = O(\|\Delta_n^{(\rvx,1)}\|_T^2) + o(\beta_n)$ and $\|\Delta_n^{(\rvy)}\|_M^2 = O(\|\Delta_n^{(\rvy,1)}\|_M^2) + o(\beta_n)$ by \eqref{eqn:delta_rvx_2}. It then follows that
\begin{align}
\|\Delta_{n+1}^{(\rvx,1)}\|_T &\leq (1-\beta_{n+1}T)\|\Delta_{n}^{(\rvx,1)}\|_T + \beta_{n+1} O(\beta_{n+1}\|L_n^{(\rvx)}\| + \beta_{n+1}\|R_n^{(\rvx)}\| + \|r^{(\rvx,1)}_n\|) \notag \\
&~~~~ + \beta_{n+1} O(\|L_n^{(\rvx)}\|^2+\|R_n^{(\rvx)}\|^2 + \|\Delta_n^{(\rvx,1)}\|^2_T + \|L_n^{(\rvy)}\|^2+\|R_n^{(\rvy)}\|^2 + \|\Delta_n^{(\rvy,1)}\|^2_M + o(\beta_n)), \label{eqn:Delta_n_x_T} \\
\|\Delta_{n+1}^{(\rvy,1)}\|_M &\leq (1-\gamma_{n+1}M)\|\Delta_{n}^{(\rvy,1)}\|_M + \gamma_{n+1} O(\gamma_{n+1}\|L_n^{(\rvy)}\| + \gamma_{n+1}\|R_n^{(\rvy)}\| + \|r^{(\rvy,1)}_n\|) \notag\\
&~~~~ + \gamma_{n+1} O(\|L_n^{(\rvx)}\|^2+\|R_n^{(\rvx)}\|^2 + \|\Delta_n^{(\rvx,1)}\|^2_T + \|\Delta_n^{(\rvx,1)}\|_T + \|L_n^{(\rvy)}\|^2+\|R_n^{(\rvy)}\|^2 + \|\Delta_n^{(\rvy,1)}\|^2_M + o(\sqrt{\beta_n})). \label{eqn:Delta_n_y_M}
\end{align}
From Lemma \ref{lemma:property_r_n} we know that $\|r^{(\rvx,1)}_n\| = o(\sqrt{\beta_n})$ and $\|r^{(\rvy,1)}_n\| = o(\sqrt{\beta_n})$, thus we can omit $o(\beta_n)$ and $o(\sqrt{\beta_n})$ terms in \eqref{eqn:Delta_n_x_T}, \eqref{eqn:Delta_n_y_M}. Lemma \ref{lemma:tighter_upper_bound_Ln} suggests that $\lim_{n\to\infty} L_n^{(\rvx)} = 0$ and $\lim_{n\to\infty} L_n^{(\rvy)} = 0$. Moreover, Lemma \ref{lemma:upper_bound_R_n} implies that $\lim_{n\to\infty} R_n^{(\rvx)} = 0$ and $\lim_{n\to\infty} R_n^{(\rvy)} = 0$ since $\omega_n$ is a bounded sequence by Condition \ref{definition:omega_n} and $\beta_n\gamma_n^{-1} = o(1)$. By almost sure convergence $\lim_{n\to\infty} \rvx_n = \rvx^*$ and $\lim_{n\to\infty} \rvy_n = \rvy^*$, we have $\lim_{n\to\infty} \Delta_n^{(\rvx)} = 0$ and $\lim_{n\to\infty} \Delta_n^{(\rvy)} = 0$. This implies that for large enough $n$, there exists some $0 < T'< T, 0 < M' < M$ such that
\begin{align*}
-T \|\Delta_{n}^{(\rvx,1)}\|_T + O(\|\Delta_{n}^{(\rvx,1)}\|_T^2) \leq T' \|\Delta_{n}^{(\rvx,1)}\|_T, \quad -M \|\Delta_{n}^{(\rvy,1)}\|_M + O(\|\Delta_{n}^{(\rvy,1)}\|_M^2) \leq -M' \|\Delta_{n}^{(\rvy,1)}\|_T.  
\end{align*}
Bringing the above inequalities back to \eqref{eqn:Delta_n_x_T} and \eqref{eqn:Delta_n_y_M} (and omit $o(\beta_n)$ and $o(\sqrt{\beta_n})$ terms therein) leads to 
\begin{align}
\|\Delta_{n+1}^{(\rvx,1)}\|_T &\leq (1-\beta_{n+1}T')\|\Delta_{n}^{(\rvx,1)}\|_T + \beta_{n+1} O(\beta_{n+1}\|L_n^{(\rvx)}\| + \beta_{n+1}\|R_n^{(\rvx)}\| + \|r^{(\rvx,1)}_n\|) \notag \\
&~~~~ + \beta_{n+1} O(\|L_n^{(\rvx)}\|^2+\|R_n^{(\rvx)}\|^2 + \|L_n^{(\rvy)}\|^2+\|R_n^{(\rvy)}\|^2 + \|\Delta_n^{(\rvy,1)}\|^2_M), \label{eqn:Delta_n_x_T2} \\
\|\Delta_{n+1}^{(\rvy,1)}\|_M &\leq (1-\gamma_{n+1}M')\|\Delta_{n}^{(\rvy,1)}\|_M + \gamma_{n+1} O(\gamma_{n+1}\|L_n^{(\rvy)}\| + \gamma_{n+1}\|R_n^{(\rvy)}\| + \|r^{(\rvy,1)}_n\|) \notag\\
&~~~~ + \gamma_{n+1} C_9(\|L_n^{(\rvx)}\|^2+\|R_n^{(\rvx)}\|^2 + \|\Delta_n^{(\rvx,1)}\|_T + \|L_n^{(\rvy)}\|^2+\|R_n^{(\rvy)}\|^2). \label{eqn:Delta_n_y_M2}
\end{align}
for some $C_9 > 0$. Since $\lim_{n\to\infty} \Delta_n^{(\rvy)} = 0$, we have $O(\|\Delta_n^{(\rvy,1)}\|_M^2) \leq C_{10} \|\Delta_n^{(\rvy,1)}\|_M$ for some $0< C_{10} < T' / C_9$ for large enough $n$ so that we can further modify \eqref{eqn:Delta_n_x_T2} as
\begin{equation}\label{eqn:Delta_n_x_T3}
\begin{split}
\|\Delta_{n+1}^{(\rvx,1)}\|_T &\leq (1-\beta_{n+1}T')\|\Delta_{n}^{(\rvx,1)}\|_T + \beta_{n+1} O(\beta_{n+1}\|L_n^{(\rvx)}\| + \beta_{n+1}\|R_n^{(\rvx)}\| + \|r^{(\rvx,1)}_n\|) \\
&~~~~ + \beta_{n+1} O(\|L_n^{(\rvx)}\|^2+\|R_n^{(\rvx)}\|^2 + \|L_n^{(\rvy)}\|^2+\|R_n^{(\rvy)}\|^2) + \beta_{n+1}C_{10}\|\Delta_n^{(\rvy,1)}\|_M.
\end{split}
\end{equation}

Rewriting \eqref{eqn:Delta_n_y_M2} gives
\begin{equation*}
\begin{split}
\|\Delta_{n+1}^{(\rvy,1)}\|_M \leq& \frac{1}{\gamma_{n+1}M'}\left[\|\Delta_{n}^{(\rvy,1)}\|_M - \|\Delta_{n+1}^{(\rvy,1)}\|_M\right] + O(\gamma_{n+1}\|L_n^{(\rvy)}\| + \gamma_{n+1}\|R_n^{(\rvy)}\| + \|r^{(\rvy,1)}_n\|) \\
&~+ C_9(\|L_n^{(\rvx)}\|^2+\|R_n^{(\rvx)}\|^2 + \|\Delta_n^{(\rvx,1)}\|_T + \|L_n^{(\rvy)}\|^2+\|R_n^{(\rvy)}\|^2). 
\end{split}
\end{equation*}
Taking it back to \eqref{eqn:Delta_n_x_T3} induces
\begin{equation}\label{eqn:Delta_n_x_T4}
\begin{split}
\|\Delta_{n+1}^{(\rvx,1)}\|_T &\leq (1-\beta_{n+1}T')\|\Delta_{n}^{(\rvx,1)}\|_T + \beta_{n+1} O(\beta_{n+1}\|L_n^{(\rvx)}\| + \beta_{n+1}\|R_n^{(\rvx)}\| + \|r^{(\rvx,1)}_n\|) \\
&~~~~ + \beta_{n+1} O(\|L_n^{(\rvx)}\|^2+\|R_n^{(\rvx)}\|^2 + \|L_n^{(\rvy)}\|^2+\|R_n^{(\rvy)}\|^2) \\
&~~~~ + \frac{\beta_{n+1}C_{10}}{\gamma_{n+1}M'}\left[\|\Delta_{n}^{(\rvy,1)}\|_M - \|\Delta_{n+1}^{(\rvy,1)}\|_M\right] + \beta_{n+1} O(\gamma_{n+1}\|L_n^{(\rvy)}\| + \gamma_{n+1}\|R_n^{(\rvy)}\| + \|r^{(\rvy,1)}_n\|) \\
&~~~~ + \beta_{n+1} C_9 C_{10}(\|L_n^{(\rvx)}\|^2+\|R_n^{(\rvx)}\|^2 + \|\Delta_n^{(\rvx,1)}\|_T + \|L_n^{(\rvy)}\|^2+\|R_n^{(\rvy)}\|^2) \\
&\leq (1-\beta_{n+1}T'')\|\Delta_{n}^{(\rvx,1)}\|_T + \beta_{n+1} O(\|L_n^{(\rvx)}\|^2+\|R_n^{(\rvx)}\|^2 + \|L_n^{(\rvy)}\|^2+\|R_n^{(\rvy)}\|^2) \\
&~~~~ + \beta_{n+1} O(\beta_{n+1}\|L_n^{(\rvx)}\| + \beta_{n+1}\|R_n^{(\rvx)}\| + \|r^{(\rvx,1)}_n\| + \gamma_{n+1}\|L_n^{(\rvy)}\| + \gamma_{n+1}\|R_n^{(\rvy)}\| + \|r^{(\rvy,1)}_n\|) \\
&~~~~ + \frac{\beta_{n+1}C_{10}}{\gamma_{n+1}M'}\left[\|\Delta_{n}^{(\rvy,1)}\|_M - \|\Delta_{n+1}^{(\rvy,1)}\|_M\right],
\end{split}
\end{equation}
where the last inequality is by setting $0 < T'' < T' - C_9C_{10}$. Now, \eqref{eqn:Delta_n_y_M2} and \eqref{eqn:Delta_n_x_T4} correspond to \citet[equations (27), (28)]{mokkadem2006convergence}. Thus, we can leverage the result therein for $\Delta_{n+1}^{(\rvx,1)}$ and $\Delta_{n+1}^{(\rvy,1)}$, which is given below.
\begin{lemma}[\citet{mokkadem2006convergence} Appendix A.4.2 and Appendix A.4.3]\label{lemma:intermediate_Delta_n}
    For $\Delta_{n+1}^{(\rvx,1)}$ and $\Delta_{n+1}^{(\rvy,1)}$ with inequalities in \eqref{eqn:Delta_n_x_T4} and \eqref{eqn:Delta_n_y_M2}, and assume that $\|\rvy_n - \rvy^*\| = O(\omega_n)$, $\|\Delta_n^{(\rvy,1)}\| = O(\eta'_n)$ where $\{\omega_n\}$ and $\{\eta_n'\}$ satisfy Condition \ref{definition:omega_n}, we have
\begin{equation*}
    \|\Delta_n^{(\rvx,1)}\| = O(\beta_n^2\gamma_n^{-2}w_n^2 + \beta_n\gamma_n^{-1}\eta'_n) + o(\sqrt{\beta_n}) \quad a.s.
\end{equation*}
\begin{equation*}
    \|\Delta_n^{(\rvy,1)}\| = O(\beta_n^2\gamma_n^{-2}w_n^2 + \beta_n\gamma_n^{-1}\eta'_n) + o(\sqrt{\beta_n}) \quad a.s.
\end{equation*}
\end{lemma}
Since $\|\Delta_n^{(\rvy)}\| = O(\eta_n)$ in Lemma \ref{lemma:upper_bound_Delta_n}, and $\|\Delta_n^{(\rvy,1)}\| \leq \|\Delta_n^{(\rvy)}\| + \|\Delta_n^{(\rvy,2)}\| = O(\eta_n + \sqrt{\beta_n})$ by \eqref{eqn:delta_rvx_2}, we set $\eta_n' = \eta_n + \sqrt{\beta_n}$. With Lemma \ref{lemma:intermediate_Delta_n}, it follows that, almost surely,
\begin{align*}
\|\Delta_n^{(\rvx)}\| &\leq \|\Delta_n^{(\rvx,1)}\| + \|\Delta_n^{(\rvx,2)}\| \\
&= O(\beta_n^2\gamma_n^{-2}w_n^2 + \beta_n\gamma_n^{-1}\eta_n) + O(\beta_n\gamma_n^{-1}\sqrt{\beta_n}) + o(\sqrt{\beta_n}) \\
&= O(\beta_n^2\gamma_n^{-2}w_n^2 + \beta_n\gamma_n^{-1}\eta_n) + o(\sqrt{\beta_n}), \\
\end{align*}
where the last equality is from $O(\beta_n\gamma_n^{-1}\sqrt{\beta_n}) = o(\sqrt{\beta_n})$ because $\beta_n\gamma_n^{-1} = o(1)$. Similarly,
\begin{align*}
\|\Delta_n^{(\rvy)}\| &\leq \|\Delta_n^{(\rvy,1)}\| + \|\Delta_n^{(\rvy,2)}\| \\
&= O(\beta_n^2\gamma_n^{-2}w_n^2 + \beta_n\gamma_n^{-1}\eta_n) + O(\beta_n\gamma_n^{-1}\sqrt{\beta_n}) + o(\sqrt{\beta_n}) \\
&= O(\beta_n^2\gamma_n^{-2}w_n^2 + \beta_n\gamma_n^{-1}\eta_n) + o(\sqrt{\beta_n}). \\
\end{align*}
This completes the proof of Lemma \ref{lemma:upper_bound_Delta_n}.
\end{proof}

By Lemma \ref{lemma:upper_bound_R_n} and Lemma \ref{lemma:upper_bound_Delta_n}, we can iteratively fine-tune the expression of $\omega_n$ and $\eta_n$ (in other words, tightening the upper bounds of $\omega_n$ and $\eta_n$). We are now ready to prove lemma \ref{lemma:final_result_R_n_Delta_n}.
\begin{proof}[\textbf{Proof of lemma \ref{lemma:final_result_R_n_Delta_n}}]
Since $\lim_{n\to\infty}\Delta_n^{(\rvy)} = 0$ almost surely by Lemma \ref{lemma:upper_bound_Delta_n}, we can set $\eta_n \equiv 1$ such that 
\begin{equation}\label{eqn:Delta_n_x_3}
    \|\Delta_n^{(\rvy)}\| = O(\beta_n^2\gamma_{n}^{-2}\omega_n^2 + \beta_n\gamma_n^{-1}) + o(\sqrt{\beta_n}).
\end{equation}
According to \eqref{eqn:Delta_n_x_3}, we set $\eta_n = O(\beta_n^2\gamma_{n}^{-2}\omega_n^2 + [\beta_n\gamma_n^{-1}]^k) + o(\sqrt{\beta_n})$ for integer $k \geq 1$. Then, we have
\begin{equation*}
\begin{split}
    \frac{\eta_n}{\eta_{n+1}} &\leq \frac{\beta_n^2\gamma_n^{-2}\omega_n^2 + [\beta_n\gamma_n^{-1}]^k + \sqrt{\beta_n}}{\beta_{n+1}^2\gamma_{n+1}^{-2}\omega_{n+1}^2 + [\beta_{n+1}\gamma_{n+1}^{-1}]^k + \sqrt{\beta_{n+1}}} \\
    &= 1 +  \frac{(\beta_n^2\gamma_n^{-2}\omega_n^2 - \beta_{n+1}^2\gamma_{n+1}^{-2}\omega_{n+1}^2) + ([\beta_n\gamma_n^{-1}]^k - [\beta_{n+1}\gamma_{n+1}^{-1}]^k) + (\sqrt{\beta_n} - \sqrt{\beta_{n+1}})}{\beta_{n+1}^2\gamma_{n+1}^{-2}\omega_{n+1}^2 + [\beta_{n+1}\gamma_{n+1}^{-1}]^k + \sqrt{\beta_{n+1}}}\\
    &\leq 1 + \frac{\beta_n^2\gamma_n^{-2}\omega_n^2 - \beta_{n+1}^2\gamma_{n+1}^{-2}\omega_{n+1}^2}{\beta_{n+1}^2\gamma_{n+1}^{-2}\omega_{n+1}^2} + \frac{[\beta_n\gamma_n^{-1}]^k - [\beta_{n+1}\gamma_{n+1}^{-1}]^k}{[\beta_{n+1}\gamma_{n+1}^{-1}]^k} + \frac{\sqrt{\beta_n} - \sqrt{\beta_{n+1}}}{\sqrt{\beta_{n+1}}}.
\end{split}
\end{equation*}
Since $[\beta_n\gamma_n^{-1}]^k / [\beta_{n+1}\gamma_{n+1}^{-1}]^k = 1 + O(1/n) = 1 + o(\gamma_n)$ for $k \geq 1$ and $\sqrt{\beta_n} / \sqrt{\beta_{n+1}} = 1 + O(1/n) = 1 + o(\gamma_n)$, together with $\omega_n/\omega_{n+1} = 1 + o(\gamma_n)$ in Condition \ref{definition:omega_n}, we have 
$$\frac{\eta_n}{\eta_{n+1}} = 1 + o(\gamma_n).$$
Thus, the new expression of $\eta_n$ also satisfies Condition \ref{definition:omega_n} for \textit{all} $k \geq 1$. There exists some integer $k_0$ such that for all $k \geq k_0$, we have $[\beta_n\gamma_n^{-1}]^k = n^{ka-kb} = o(\sqrt{\beta_n})$ such that
\begin{equation*}
    \|\Delta_n^{(\rvy)}\| = O(\eta_n) = O(\beta_n^2\gamma_{n}^{-2}\omega_n^2) + o(\sqrt{\beta_n}).
\end{equation*}
Similarly, we have
\begin{equation*}
    \|\Delta_n^{(\rvx)}\| = O(\beta_n^2\gamma_{n}^{-2}\omega_n^2) + o(\sqrt{\beta_n}).
\end{equation*}

For $\|\rvy_n - \rvy^*\| = O(\omega_n)$, we set  $\omega_n = \sqrt{\gamma_n \log s_n} + [\beta_n\gamma_n^{-1}]^k$ for integer $k \geq 1$ and check that
\begin{equation*}
    \frac{\omega_n}{\omega_{n+1}} = \frac{\sqrt{\gamma_n \log s_n} + [\beta_n\gamma_n^{-1}]^k}{\sqrt{\gamma_{n+1} \log s_{n+1}} + [\beta_{n+1}\gamma_{n+1}^{-1}]^k} \leq  1 + \frac{\sqrt{\gamma_n \log s_n} - \sqrt{\gamma_{n+1} \log s_{n+1}}}{\sqrt{\gamma_{n+1} \log s_{n+1}}} + \frac{[\beta_n\gamma_n^{-1}]^k - [\beta_{n+1}\gamma_{n+1}^{-1}]^k}{[\beta_{n+1}\gamma_{n+1}^{-1}]^k}.
\end{equation*}
After algebraic calculation, we have $\sqrt{\gamma_n \log s_n} / \sqrt{\gamma_{n+1} \log s_{n+1}} < \sqrt{\gamma_n} / \sqrt{\gamma_{n+1}} = 1 + O(1/n) = 1 + o(\gamma_n)$. Along with $[\beta_n\gamma_n^{-1}]^k / [\beta_{n+1}\gamma_{n+1}^{-1}]^k = 1+ o(\gamma_n)$, it then follows that
$$\frac{\omega_n}{\omega_{n+1}} = 1 + o(\gamma_n).$$
Therefore, $\omega_n = \sqrt{\gamma_n \log s_n} + [\beta_n\gamma_n^{-1}]^k$ satisfies Condition \ref{definition:omega_n} for all $k \geq 1$. There exists some integer $k_1$ such that for all $k \geq k_1$, $[\beta_n\gamma_n^{-1}]^k = o(\sqrt{\gamma_n})$, which in turn implies that $[\beta_n\gamma_n^{-1}]^k = o(\sqrt{\gamma_n \log s_n})$. Thus, $\omega_n = O(\sqrt{\gamma_n\log s_n})$ and $\|\rvy_n - \rvy^*\| = O(\sqrt{\gamma_n\log s_n})$ almost surely.

Consequently, with $\omega_n = O(\sqrt{\gamma_n\log s_n})$, we have
\begin{equation*}
    \|\Delta_n^{(\rvx)}\| = O(\beta_n^2\gamma_n^{-1}\log s_n) + o(\sqrt{\beta_n}) = o(\sqrt{\beta_n}), \quad \|\Delta_n^{(\rvy)}\| = O(\beta_n^2\gamma_n^{-1}\log s_n) + o(\sqrt{\beta_n}) = o(\sqrt{\beta_n}).
\end{equation*}
For $\|R_n^{(\rvx)}\|, \|R_n^{(\rvy)}\|$ of the forms in Lemma \ref{lemma:upper_bound_R_n}, we have the following:
\begin{equation*}
    \|R_n^{(\rvx)}\| = O(\beta_n\gamma_{n}^{-1/2}\log s_n + n^{-s}) = O(n^{a/2-b}\log s_n + n^{-s}) = O(n^{-c}),
\end{equation*}
where $c \triangleq \min\{b - a/2 + \epsilon, s\} > b/2$ for some small enough $\epsilon>0$. In addition, we have
\begin{equation*}
    \|R_n^{(\rvy)}\| = O(\beta_n\gamma_{n}^{-1/2}\log s_n + \sqrt{\beta_n \log u_n}) = O(\sqrt{\beta_n \log u_n}),
\end{equation*}
which completes the proof.
\end{proof}

\section{Performance Ordering in TTSA}\label{appendix:proof_propositon3.1}
\subsection{Proof of Proposition 3.2}
By definition of efficiency ordering in Definition 3.1 (in Section 3.1), we recall that two efficiency-ordered Markov chains $\{W_n\}$ and $\{Z_n\}$ with $W \preceq Z$ and same stationary distribution $\vmu$ obey the following Loewner ordering:
\begin{equation*}
    \rmU^{(W)}(g) \geq_L \rmU^{(Z)}(g)
\end{equation*}
for any vector-valued function $g:\Xi \to \sR^d$, where $$\rmU^{(W)}(g) = \lim_{s\to\infty}\frac{1}{s}\E\left[\left(\sum_{n=1}^s g(W_n) - \E_{\vmu}[g]\right)\left(\sum_{n=1}^s g(W_n) - \E_{\vmu}[g]\right)^T\right]$$
and $\E_{\vmu}[g] = \E_{W \sim \vmu}[g(W)]$.

Now, we turn to the explicit form of $\rmV_{\rvx}$ and $\rmV_{\rvy}$ in our Theorem 2.2 in Section 2.3 and give them below for completeness.
\begin{equation*}
\begin{split}
    &\rmV_{\rvx} \!=\! \int_0^{\infty}\! e^{t\left(\rmK_{\rvx}+\frac{\mathds{1}_{\{b=1\}}}{2}\rmI\right)} \rmU_{\rvx} e^{t\left(\rmK_{\rvx}+\frac{\mathds{1}_{\{b=1\}}}{2}\rmI\right)^T} dt,\\
    &\rmV_{\rvy} \!=\! \int_0^{\infty} e^{t\rmQ_{22}} \rmU_{22} e^{t\rmQ_{22}^T} dt,
\end{split}
\end{equation*}
where the expression of $\rmU_{\rvx}$ and $\rmU_{22}$ can be found in Remark \ref{remark:2} in Appendix \ref{appendix:2.1}. The only components in $\rmV_{\rvx}, \rmV_{\rvy}$ that are associated with the underlying Markov chain are $\rmU_{\rvx}$ and $\rmU_{22}$. Let the function $g(W_n) \equiv h_2(\rvx^*,\rvy^*,W_n)$ for the TTSA algorithm \eqref{eqn:general_TTSA_form} driven by the Markov chain $\{W_n\}$, then replacing $\{W_n\}$ with a more efficient chain $\{Z_n\}$ leads to $\rmU^{(W)}(g) \geq_L \rmU^{(Z)}(g)$, or equivalently, $\rmU_{22}^{(W)} \geq_L \rmU_{22}^{(Z)}$, where the superscript $(W)$ indicates that the TTSA algorithm \eqref{eqn:general_TTSA_form} is driven by the Markov chain $\{W_n\}$. 

Similarly, let $g(W_n) \equiv \begin{bmatrix}
    h_1(\rvx^*,\rvy^*,W_n) \\ h_2(\rvx^*,\rvy^*,W_n)
\end{bmatrix}$, then we have
\begin{equation}\label{eqn:loewner_ordering_U}
\begin{split}
    \begin{bmatrix}
    \rmU_{11}^{(W)} & \rmU_{12}^{(W)} \\ \rmU_{21}^{(W)} & \rmU_{22}^{(W)}
\end{bmatrix} &= \lim_{s\to\infty} \frac{1}{s}\E\left[\left(\sum_{n=1}^s g(W_n)\right)\left(\sum_{n=1}^s g(W_n)\right)^T\right] \\
&\geq_L \lim_{s\to\infty} \frac{1}{s}\E\left[\left(\sum_{n=1}^s g(Z_n)\right)\left(\sum_{n=1}^s g(Z_n)\right)^T\right] = \begin{bmatrix}
    \rmU_{11}^{(Z)} & \rmU_{12}^{(Z)} \\ \rmU_{21}^{(Z)} & \rmU_{22}^{(Z)}
\end{bmatrix}. 
\end{split}
\end{equation}
We then focus on the matrix $\rmU_{\rvx}$ in the form of \eqref{eqn:alt_matrix_U_x_form}. From the definition of Loewner ordering, for any matrix $\rmC$ with suitable dimension, $\rmA \geq \rmB$ leads to $\rmC\rmA\rmC^T \geq_L \rmC\rmB\rmC^T$. Let $\rmC \equiv [\rmI -\rmQ_{12}\rmQ_{22}^{-1}]$ such that $\rmU_{\rvx} = \rmC \begin{bmatrix}
\rmU_{11} & \rmU_{12} \\ \rmU_{21} & \rmU_{22}
\end{bmatrix}\rmC^T$. Then, with \eqref{eqn:loewner_ordering_U}, we have $\rmU_{\rvx}^{(W)}\geq_L \rmU_{\rvx}^{(Z)}$.

Then, for $\rmV_{\rvx}^{(W)}$ and $\rmV_{\rvx}^{(Z)}$, by the fact that $\rmA_1 \geq_L \rmB_1$ and $\rmA_2 \geq \rmB_2$ leads to $\rmA_1 + \rmA_2 \geq \rmB_1 + \rmB_2$, i.e., Loewner ordering is closed under addition, together with $\rmC\rmA\rmC^T \geq_L \rmC\rmB\rmC^T$, we have $\rmV_{\rvx}^{(W)} \geq_L \rmV_{\rvx}^{(Z)}$ since $\rmU_{\rvx}^{(W)}\geq_L \rmU_{\rvx}^{(Z)}$. Following similar steps above gives $\rmV_{\rvy}^{(W)} \geq_L \rmV_{\rvy}^{(Z)}$ because $\rmU_{22}^{(W)}\geq_L \rmU_{22}^{(Z)}$. This completes the proof.

\section{Asymptotic Behavior of Nonlinear GTD
Algorithms}\label{appendix:proof_propositon3.3}
\subsection{Introduction to GTD2 and TDC Algorithms with Nonlinear Function Approximation}\label{appendix:4.1}
Before starting the proof, for self-contained purposes, we here present the GTD2 and TDC algorithm with nonlinear function approximation, first proposed by \citet{maei2009convergent}. In particular, both algorithms can be represented as TTSA in \eqref{eqn:general_TTSA_form}, where $$h_2(\rvx_n,\rvy_n,\xi_{n+1}) \equiv \delta_n(\rvx_n)\phi_{\rvx_n}(s_n) -\phi_{\rvx_n}(s_n)\phi_{\rvx_n}(s_n)^T\rvy_n,$$ 
and 
\begin{enumerate}
    \item[(i)] For TDC algorithm: $$h_1(\rvx_n,\rvy_n,\xi_{n+1}) \equiv \delta_n(\rvx_n)\phi_{\rvx_n}(s_n) - f_n(\rvx_n,\rvy_n) - \alpha \phi_{\rvx_n}(s_{n+1})\phi_{\rvx_n}\!(s_n)^T\rvy_n,$$
    \item[(ii)] For GTD2 algorithm: $$h_1(\rvx_n,\rvy_n,\xi_{n+1}) \equiv (\phi_{\rvx_n}(s_n)-\alpha\phi_{\rvx_n}(s_{n+1}))\phi_{\rvx_n}(s_n)^T\rvy_n - f_n(\rvx_n,\rvy_n),$$
\end{enumerate}
where $\xi_{n+1} \triangleq (s_n, s_{n+1})$, the feature vector $\phi_{\rvx}(s) \triangleq \nabla_{\rvx} V_{\rvx}(s) \in \sR^d$ and the TD error $\delta_n(\rvx) \triangleq r(s_n, a_n, s_{n+1}) + \alpha V_{\rvx}(s_{n+1}) - V_{\rvx}(s_n) \in \sR$. Lastly, we introduce $f_n(\rvx,\rvy) \triangleq (\delta_n(\rvx) - \phi_{\rvx}(s_n)^T\rvy)\nabla_{\rvx}\phi_{\rvx}(s_n)\rvy \in \sR^d$ for $\rvx,\rvy \in \sR^d$. The conditions on step sizes $\beta_n$ and $\gamma_n$ are in Assumption (A1) in Section 2.1.
For both algorithms, the iterates $\rvx_n$ evolve the parameter for $V_{\rvx}$ to accurately estimate the value function $V^{\vpi}$, and iterates $\rvy_n$ aim to approximate $\E_{\vmu^{\vpi}}[\delta_n(\rvx)\phi_{\rvx}(s_n)]$ for each $\rvx$ value from iterates $\rvx_n$. 
As demonstrated in \citet[Corollary 1]{maei2009convergent}, iterates $(\rvx_n, \rvy_n)$ admit a root $(\rvx^*, \rvy^*)$, where $\rvx^*$ satisfies $\E_{\vmu^{\vpi}}[\delta_n(\rvx^*)\phi_{\rvx^*}(s_n)] = \vzero$, and $\rvy^* = \vzero$.

In the following, we list the conditions (C1) - (C4) commonly assumed in RL.
\begin{enumerate}[label=C\arabic*., ref=(C\arabic*)]
    \item  For any $s \!\in\! \gS$ and $\rvx, \rvx' \!\in\! \sR^{d}$, $|V_{\rvx}(s)| \!\leq\! C_{v}$,  $\|\phi_{\rvx}(s)\|\!\leq\! C_{\phi}$, and $\|\nabla_{\rvx}\phi_{\rvx}(s)\| \!\leq\! D_{v}$ for some positive constants $C_{v}, C_{\phi}, D_{v}$. Besides, we assume $f_n(\rvx,\rvy)$ is Lipschitz continuous in $\rvy$ for any $\rvx \in \sR^d$; \label{condition:1}
    \item The point $(\rvx^*,\rvy^*)$, where $\rvx^*$ satisfies $\E_{\vmu^{\vpi}}[\delta_n(\rvx^*)\phi_{\rvx^*}(s_n)] = \vzero$ and $\rvy^* = \vzero$, is the globally asymptotically stable equilibrium of the related ODE $\dot \rvx = \bar h_1(\rvx,\rmC(\rvx)^{-1}\E_{\vmu^{\vpi}}[\delta_n(\rvx)\phi_{\rvx}(s_n)])$, where $\rmC(\rvx) \!\triangleq \! \E_{\vmu^{\vpi}}[\phi_{\rvx}(s)\phi_{\rvx}(s)^T]$; \label{condition:2}
    \item Matrix $\rmC(\rvx) \geq_L \zeta\rmI >_L \vzero, \forall \rvx \in \sR^d$ for some $\zeta > 0$, and we use the shorthand notation for $\rmC_{*} \triangleq \rmC(\rvx^*)$. We also assume that $\rmA_*$ is full rank, where $\rmA_* \!\triangleq\! \E_{\vmu^{\vpi}}[\phi_{\rvx^*}(s_n)(\phi_{\rvx^*}(s_n)\!-\!\alpha\phi_{\rvx^*}(s_{n+1}))^T \!+\!\delta_n(\rvx^*)\nabla_{\rvx}\phi_{\rvx^*}(s_n)]$;\footnote{Only when we consider the CLT result and the fastest decaying step size $\beta_n = 1/(n+1)$, we need an extra assumption, i.e., $-\rmA_{*}^T\rmC_{*}^{-1}\rmA_{*} + \frac{1}{2}\rmI$ be to Hurwitz. This condition is not needed for almost sure convergence even when $\beta_n = 1/ (n+1)$.} \label{condition:3}
    \item The Markov chain with transition kernel $\sP(s_{n+1} \!=\! s' | s_n \!=\! s) \!\triangleq\! \sum_{a\in\gA}\vpi(a|s)P(s'|s,a)$ is ergodic; \label{condition:4}
    \item $\sup_{n\geq 0} (\|\rvx_n\| + \|\rvy_n\|) < \infty$ a.s. \label{condition:5}
\end{enumerate}
The boundedness assumption imposed on $\phi_{\rvx}(s), \nabla_{\rvx}\phi_{\rvx}(s)$ and $V_{\rvx}(s)$ in Condition \ref{condition:1}, as well as the Lipschitz condition on $f_n(\rvx,\rvy)$, are in line with the assumptions made in the state-of-the-art work for nonlinear TDC algorithm \citep{xu2021sample,wang2021non}. Condition \ref{condition:2} is to ensure the globally asymptotically stability of the related ODE $\dot \rvx = \bar h_1(\rvx,\rmC(\rvx)^{-1}\E_{\vmu^{\vpi}}[\delta_n(\rvx)\phi_{\rvx}(s_n)])$. A similar assumption has been made in \citet[Section 5]{maei2009convergent} where they consider the asymptotically stable equilibrium for any trajectory of the aforementioned ODE in a compact set because their algorithms project iterates $(\rvx_n, \rvy_n)$ onto that set. Condition \ref{condition:3} aligns with current work on the nonlinear function approximation \citet{maei2009convergent,wang2021non}. In the special case of linear function approximation, Condition \ref{condition:3} is also widely used in the literature \citep{sutton2009fast,dalal2018finite,dalal2020tale,li2023sharp}. Condition \ref{condition:4} is typical for Markovian samples \citep{ma2020variance,xu2021sample,wang2021non}. Condition \ref{condition:5} ensures the stability of $(\rvx_n, \rvy_n)$, which serves the same purpose as the projection operator for the iterates $(\rvx_n,\rvy_n)$ in the original GTD2 and TDC algorithms in \citet{maei2009convergent}, where $\|\rvx_n\|$, $\|\rvy_n\|$ remain constrained by an upper bound.

\subsection{Proof of Proposition 3.3}\label{appendix:4.2}
We now explain how conditions \ref{condition:1} -- \ref{condition:5} correspond to assumptions \ref{assump:2} -- \ref{assump:5} in order to apply our main CLT result in Theorem 2.2. By Condition \ref{condition:1}, we have that
\begin{equation*}
    \|h_2(\rvx,\rvy,\xi)\| \leq (r_{\max} + (1+\alpha)C_{v})C_{\phi} + C_{\phi}^2 \|\rvy\| = O(1 + \|\rvy\|),
\end{equation*}
where $r_{\max} \triangleq \max_{(s,a,s') \in \gS \times \gA \times \gS} r(s,a,s') < \infty$. Moreover, for GTD2 algorithm,
\begin{equation*}
    \|h_1(\rvx,\rvy,\xi)\| \leq (1+\alpha)C_{\phi}^2\|\rvy\| + \|f_n(\rvx,\rvy)\| = O(1 + \|\rvy\|). 
\end{equation*}
For TDC algorithm,
\begin{equation*}
    \|h_1(\rvx,\rvy,\xi)\| \leq (r_{\max}+(1+\alpha)C_{v})C_{\phi} + \|f_n(\rvx,\rvy)\| + \alpha C_{\phi}^2 \|\rvy\| = O(1 + \|\rvy\|). 
\end{equation*}
Therefore, Assumption \ref{assump:2} is satisfied. 

Then, we turn to verifying Assumption \ref{assump:3}. For any $\rvx \in \sR^d$, to ensure $\bar h_2(\rvx,\lambda(\rvx)) = 0$, we can set $\lambda(\rvx) = \rmC(\rvx)^{-1}\E_{\vmu^{\vpi}}[\delta_n(\rvx)\phi_{\rvx}(s_n)]$, where $\rmC(\rvx)^{-1}$ is well defined by Condition \ref{condition:3}. Clearly, $\lambda(\rvx)$ is the globally asymptotically stable equilibrium of $\dot \rvy = \bar h_2(\rvx,\rvy) = \E_{\vmu}[\delta_n(\rvx),\phi_{\rvx}(s_n)] - \rmC(\rvx)\rvy$ because this ODE is linear in $\rvy$, and $\nabla_{\rvy}\bar h_2(\rvx,\lambda(\rvx)) = -\rmC(\rvx)$ is Hurwitz for any $\rvx \in \sR^d$, as stated in Condition \ref{condition:3}. By Condition \ref{condition:1}, we have $$\|\lambda(\rvx)\| \leq \|\rmC(\rvx)^{-1}\|(r_{\max}+(1+\alpha)C_{v})C_{\phi} \leq \zeta^{-1}(r_{\max}+(1+\alpha)C_{v})C_{\phi}.$$
When $\rvx = \rvx^*$, $\rvy^* = \lambda(\rvx^*) = 0$ such that $f_n(\rvx^*,\rvy^*)$ = 0 and $\hat h_1(\rvx^*) = \bar h_1(\rvx^*,\lambda(\rvx^*)) = 0$ for both GTD2 and TDC algorithms. Finally, we deal with $\nabla_{\rvx} \hat h_1(\rvx^*)$. By chain rule, we have $\nabla_{\rvx} \hat h_1(\rvx) = \nabla_{\rvx} \bar h_1(\rvx,\lambda(\rvx)) + \nabla_{\rvy}\bar h_2(\rvx,\lambda(\rvx))\nabla_{\rvx}\lambda(\rvx)$. Although $\lambda(\rvx)$ is an implicit function, after taking the derivative on both sides of $\bar h_2(\rvx,\lambda(\rvx)) = 0$ with respective to $\rvx$, we have 
$$\nabla_{\rvx} \bar h_2(\rvx,\lambda(\rvx)) + \nabla_{\rvy} \bar h_2(\rvx,\lambda(\rvx)) \nabla_{\rvx}\lambda(\rvx) = 0.$$
This further leads to $$\nabla_{\rvx}\lambda(\rvx) = -(\nabla_{\rvy}\bar h_2(\rvx,\lambda(\rvx)))^{-1}\nabla_{\rvx}\bar h_2(\rvx,\lambda(\rvx)),$$
where the matrix inverse exists because $\nabla_{\rvy}\bar h_2(\rvx,\lambda(\rvx))$ is Hurwitz for any $\rvx \in \sR^d$. Thus, we obtain the following form for $\nabla_{\rvx} \hat h_1(\rvx^*)$:
\begin{equation}
    \nabla_{\rvx} \hat h_1(\rvx^*) = \nabla_{\rvx} \bar h_1(\rvx^*,\lambda(\rvx^*)) - \nabla_{\rvy} \bar h_1(\rvx^*,\lambda(\rvx^*))(\nabla_{\rvy}\bar h_2(\rvx^*,\lambda(\rvx^*)))^{-1} \nabla_{\rvx} \bar h_2(\rvx^*,\lambda(\rvx^*)). 
\end{equation}
After algebraic calculation, for both GTD2 and TDC algorithms, $\nabla_{\rvx} \hat h_1(\rvx^*) = -\rmA_{*}^T\rmC_{*}^{-1}\rmA_{*}$. Now that $\rmA_{*}$ is full rank, for any non-zero vector $\rvv\in\sR^d$, we have $\rvv^T \nabla_{\rvx} \hat h_1(\rvx^*) \rvv = - \rvv^T \rmA_{*}\rmC_{*}^{-1}\rmA_{*} \rvv = - \rvu^T\rmC_{*} \rvu < 0$ by letting $\rvu = \rmA_{*}\rvv \neq \vzero$. Thus, $\nabla_{\rvx} \hat h_1(\rvx^*)$ is negative definite and thus Hurwitz. Additionally, Condition \ref{condition:2} indicates that $(\rvx^*,\lambda(\rvx^*))$ is the globally asymptotically stable equilibrium of the related ODE $\dot \rvx = \bar h_1(\rvx,\lambda(\rvx))$. Therefore, Assumption \ref{assump:3} is verified.

Regarding $\{\xi_n\}$ with $\xi_{n+1} = (s_n, s_{n+1})$, it can seen as the Markov chain $\{s_n\}$ on the augmented state space. Then, we have the following result.
\begin{theorem}[\cite{neal2004improving} Theorem $2$]\label{theorem:augmented_mc}
Suppose that $\{s_n\}$ is an irreducible, reversible Markov chain on the finite state space $\gV$ with transition matrix $\rmP = \{P(i,j)\}$ and stationary distribution $\vpi$. Construct a Markov chain $\{\xi_n\}$ on the augmented state space $\gE \triangleq \{(i,j): i,j \in \gV ~~s.t.~~ P(i,j) > 0\} \subseteq \gV \times \gV$ with transition matrix $\mP' = \{P'(e_{ij},e_{lk})\}$ in which $e_{ij} \triangleq (s = i, s' = j)$ and the transition probabilities $P'(e_{ij},e_{lk})$ satisfy the following two conditions: for all $e_{ij}, e_{jk} \in \gE$, 
\begin{equation}
    P'(e_{ij},e_{jk}) = P(j,k).
\end{equation}
Then, the Markov chain $\{\xi_n\}_{n\geq 0}$ is irreducible with a unique stationary distribution $\vpi'$ in which
\begin{equation}\label{eqn:stationary_augmented}
    \pi'(e_{ij}) = \pi_i P(i,j) = \pi_j P(j,i), ~~e_{ij} \in \gE.
\end{equation}
\end{theorem}
By Theorem \ref{theorem:augmented_mc}, $\{\xi_n\}$ is an ergodic Markov chain and thus satisfies Assumption \ref{assump:4}. Condition \ref{condition:4} matches Assumption \ref{assump:5}. Therefore, we can apply Lemma 2.1 (almost sure convergence) and Theorem 2.2 (CLT result) to GTD2 and TDC algorithms. The results are given as follows.
\begin{align*} 
    &\lim_{n\to\infty} \rvx_n = \rvx^* \quad \text{a.s.} ~~\text{and} \quad \lim_{n\to\infty} \rvy_n = \vzero \quad \text{a.s.} \\ 
    &\frac{1}{\sqrt{\beta_n}}(\rvx_n - \rvx^*)\! \xrightarrow{~d~} N(\vzero,\rmV_{\rvx}), ~ \frac{1}{\sqrt{\gamma_n}}\rvy_n\xrightarrow{~d~} N(\vzero,\rmV_{\rvy}), 
\end{align*}
where $\rmV_{\rvx}, \rmV_{\rvy}$ are identical for both algorithms, and
\begin{align*}
\rmV_{\rvx} = \int_0^{\infty} e^{t\rmK'} \rmU_{\rvx}e^{t(\rmK')^T}, \quad \rmV_{\rvy} = \int_0^{\infty} e^{-t\rmC_{*}} \rmU_{\rvy}e^{-t\rmC_{*}^T},    
\end{align*}
where
\begin{align}\label{eqn:exact_form_CLT_GTD2_TDC}
&\rmK_{\rvx}' = - \rmA_{*}^T \rmC_{*}^{-1} \rmA_{*} + \frac{\mathds{1}_{\{b=1\}}}{2}\rmI, \\
&\rmU_{\rvy} = \lim_{s\to\infty}\frac{1}{s}\E\left[\left(\sum_{n=1}^s \delta_n(\rvx^*)\phi_{\rvx^*}(s_n)\right)\left(\sum_{n=1}^s \delta_n(\rvx^*)\phi_{\rvx^*}(s_n)\right)^T\right], \\
&\rmU_{\rvx} = \rmA_{*}^T \rmC_{*}^{-1} \rmU_{\rvy} \rmC_{*}^{-1} \rmA_{*}.
\end{align}

\section{Simulation Setups and Additional Numerical Results}\label{appendix:simulation}
In this appendix, we provide a more detailed illustration of the numerical results. The simulations are conducted on a PC with AMD Ryzen R9 5950X, 128GB RAM and RTX 3080. 

\subsection{Distributed Learning in Section 3.1}
\subsubsection{Simulation Setup for L2-regularized Binary Classification}\label{appendix:5.1.1}
In Section 3.1, we perform the L2-regularized binary classification problem using the momentum SGD algorithm on the wikiVote graph \citep{leskovec2014snap}. Specifically, the problem has the following objective function:
\begin{equation}
\min_{\rvx \in \sR^d} \left\{f(\rvx) = \frac{1}{N}\sum_{i=1}^N F(\rvx,i) \triangleq \frac{1}{N} \sum_{i=1}^N \log\left(1+e^{\rvx^T \rvs_i}\right) -z_i \left(\rvx^T \rvs_i\right)  + \frac{\kappa}{2} \| \rvx\|^2\right\},
\vspace{-1mm}
\end{equation}
where $\{(\rvs_i,z_i)\}_{i=1}^N$ is the \textit{a9a} dataset (with $123$ features, i.e., $\rvs_i \in \sR^{123}$) from LIBSVM \citep{chang2011libsvm}, and penalty parameter $\kappa = 1$.
The momentum SGD algorithm \citep{gadat2018stochastic,li2022revisiting} employed in this simulation is given below.
\begin{algorithm}
\DontPrintSemicolon
Initial parameters $\rvx_0, \rvy_0 = \vzero$, data point $\xi_0$, step sizes $\beta_n = (n+1)^{-1}$ and $\gamma_n = (n+1)^{-0.501}$, number of iterations $T$;

\For{$n = 0$ \KwTo $T$}{
    Sample new data point: $\xi_{n+1} \leftarrow$ Sampling Strategy\;
    Compute gradient: $\rvg_n \leftarrow \nabla F(\rvx_n,\xi_{n+1})$\;
    Update momentum: $\rvy_{n+1} \leftarrow  \rvy_{n} - \gamma_{n+1} (\rvg_n + \rvy_n)$\;
    Update parameter: $\rvx_{n+1} \leftarrow \rvx_{n} + \beta_{n+1} \rvy_n$\;
}
\caption{Momentum SGD}
\label{algorithm:mSGD}
\end{algorithm}

For \textit{i.i.d.} sampling and single shuffling in the simulation, where the whole dataset is available in each iteration, we have the following schemes: at $n$-th iteration,
\begin{itemize}
    \item \textbf{\textit{i.i.d.} sampling}: $\xi_{n+1}$ is sampled from $[N]$ uniformly at random;
    \item \textbf{single shuffling}: At the beginning of the simulation, we shuffle the sequence $\{1,2,\cdots,N\}$ by the permutation operator $\sigma: [N] \to [N]$ and then sample the data point according to $\displaystyle \xi_{n+1} = \sigma(n+1 ~\text{mod} ~ N)$.
\end{itemize}

When dataset is distributed over the wikiVote graph, i.e., each node on the graph is assigned a data point, we use simple random walk (SRW) and its sampling-efficient counterpart non-backtracking random walk (NBRW) \citep{alon2007non,lee2012beyond,ben2018comparing} in the simulation. Specifically, denote by $\xi_n$ the index of the node in the $n$-th iteration and $\gN(\xi_n)$ the list of neighboring nodes of node $\xi_n$, we have
\begin{itemize}
    \item \textbf{SRW}: $\xi_{n+1}$ is sampled from $\gN(\xi_n)$ uniformly at random;
    \item \textbf{NBRW}: $\xi_{n+1}$ is sampled from $\gN(\xi_n) \backslash \{\xi_{n-1}\}$ uniformly at random. If $\gN(\xi_n) \backslash \{\xi_{n-1}\} = \emptyset$, then $\xi_{n+1} = \xi_{n-1}$.
\end{itemize}
Note that SRW and NBRW both have a stationary distribution proportional to degree distribution, while the objective function indicates that each node is treated equally, which results in the bias from SRW and NBRW. To overcome this problem, we employ importance reweighting, e.g., by modifying the momentum update step in Algorithm \ref{algorithm:mSGD} in the following form:
\begin{equation*}
\rvy_{n+1} \leftarrow  \rvy_{n} - \gamma_{n+1} \left(\rvg_n \cdot \frac{1}{d_{\xi_{n+1}}} - \rvy_n\right),
\end{equation*}
where $d_i$ is defined as the degree of node $i$. 

\subsubsection{Additional Simulation on Distributed Minimax Problem}
In this part, we consider the following minimax problem
\begin{equation}\label{eqn:minimax_problem}
\min_{\rvx \in \sR^d}\max_{\rvy\in \sR^d} \left\{f(\rvx,\rvy) = \frac{1}{N}\sum_{i=1}^N F(\rvx,\rvy,i) \triangleq \frac{1}{N} \sum_{i=1}^N -\left[\frac{1}{2}\|\rvy\|^2 - \rvb(i)^T\rvy + \rvy^T\rmA(i)\rvx\right]  + \frac{\kappa}{2} \| \rvx\|^2\right\},    
\end{equation}
and follow the same setup as in \citet{tarzanagh2022fednest} to generate the dataset. In particular, we let $\kappa = 10$, $d = 10$, $\rvb(i) = \rvb'(i) - \frac{1}{N}\sum_{i=1}^N \rvb'(i)$ and $\rmA(i) = t_i \rmI$, where $\rvb'(i) \sim N(0,\rmI)$ and $t_i$ are drawn from $(0,0.1)$ uniformly at random. We test the distributed minimax problem over the WikiVote graph \citep{leskovec2014snap}, where each node is assigned a data point, thus $889$ data points in total. To solve the minimax problem \eqref{eqn:minimax_problem}, we leverage the stochastic gradient descent ascent (SGDA) algorithm as follows, which is regarded as a special case of TTSA \citep{lin2020gradient}.
\begin{algorithm}
\DontPrintSemicolon
Initial parameters $\rvx_0, \rvy_0 = \vzero$, data point $\xi_0$, step sizes $\beta_n = (n+1)^{-1}$ and $\gamma_n = (n+1)^{-0.8}$, number of iterations $T$;

\For{$n = 0$ \KwTo $T$}{
    Sample new data point: $\xi_{n+1} \leftarrow$ Sampling Strategy\;
    Compute gradient w.r.t $\rvx$: $\rvg_n \leftarrow \nabla_{\rvx} F(\rvx_n, \rvy_n,\xi_{n+1})$\;
    Compute gradient w.r.t $\rvy$: $\rvh_n \leftarrow \nabla_{\rvy} F(\rvx_n, \rvy_n,\xi_{n+1})$\;
    Update inner parameter: $\rvy_{n+1} \leftarrow  \rvy_{n} + \gamma_{n+1} \rvh_n$\;
    Update outer parameter: $\rvx_{n+1} \leftarrow \rvx_{n} - \beta_{n+1} \rvg_n$\;
}
\caption{SGDA}
\label{algorithm:SGDA}
\end{algorithm}

In this simulation, we compare two pairs of sampling strategies for the performance ordering, i.e., SRW versus NBRW, \textit{i.i.d.} sampling versus single shuffling, which have been introduced in Appendix \ref{appendix:5.1.1}. Especially, for SRW and NBRW, we reweight the gradient in the update of both inner and outer parameters, i.e.,
\begin{align*}
&\rvy_{n+1} \leftarrow  \rvy_{n} + \gamma_{n+1} \rvh_n\cdot\frac{1}{d_{\xi_{n+1}}}, \\
&\rvx_{n+1} \leftarrow \rvx_{n} - \beta_{n+1} \rvg_n\cdot\frac{1}{d_{\xi_{n+1}}}.
\end{align*}

\begin{figure}[ht]
    \centering
    \begin{subfigure}[b]{0.48\columnwidth}
        \centering
        \includegraphics[width=\textwidth]{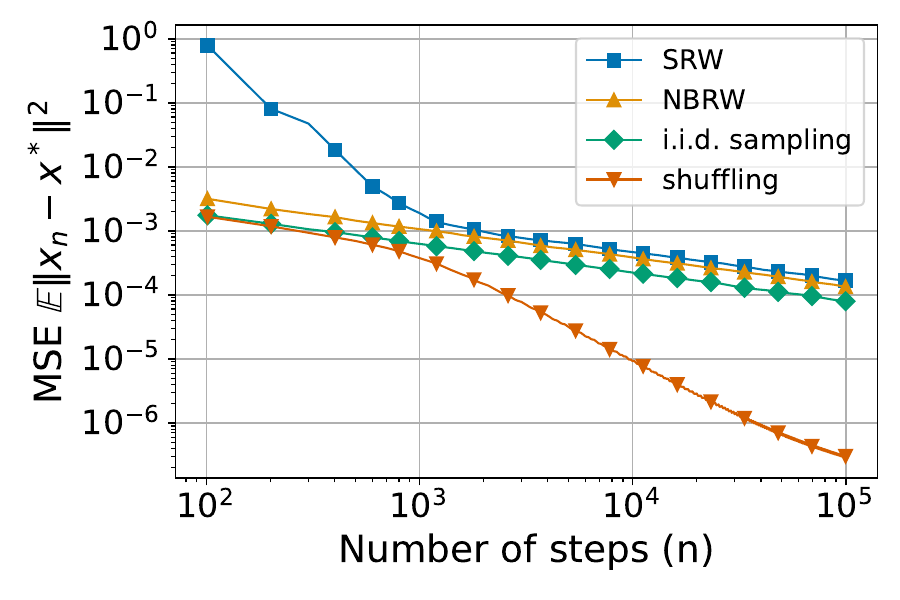}
        \caption{MSE}
        \label{fig:6a}
    \end{subfigure}
    \hfill
    \begin{subfigure}[b]{0.48\columnwidth}
        \centering
        \includegraphics[width=\textwidth]{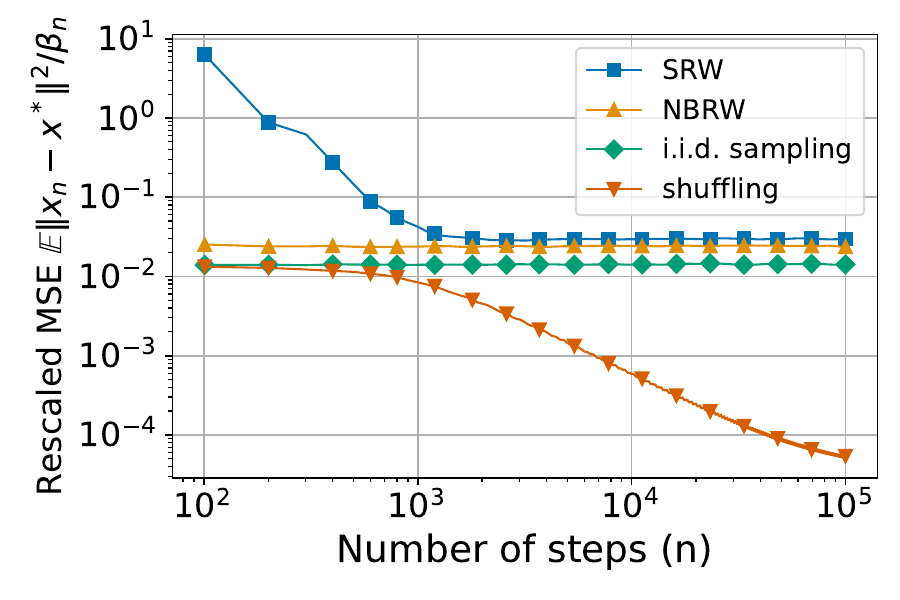}
        \caption{Rescaled MSE}
        \label{fig:6b}
    \end{subfigure}
    \hfill
    \caption{Comparison of the performance ordering in SGDA in terms of iterates $\rvx_n$.}
    \label{fig:6}
\end{figure}

\begin{figure}[ht]
    \centering
    \begin{subfigure}[b]{0.48\columnwidth}
        \centering
        \includegraphics[width=\textwidth]{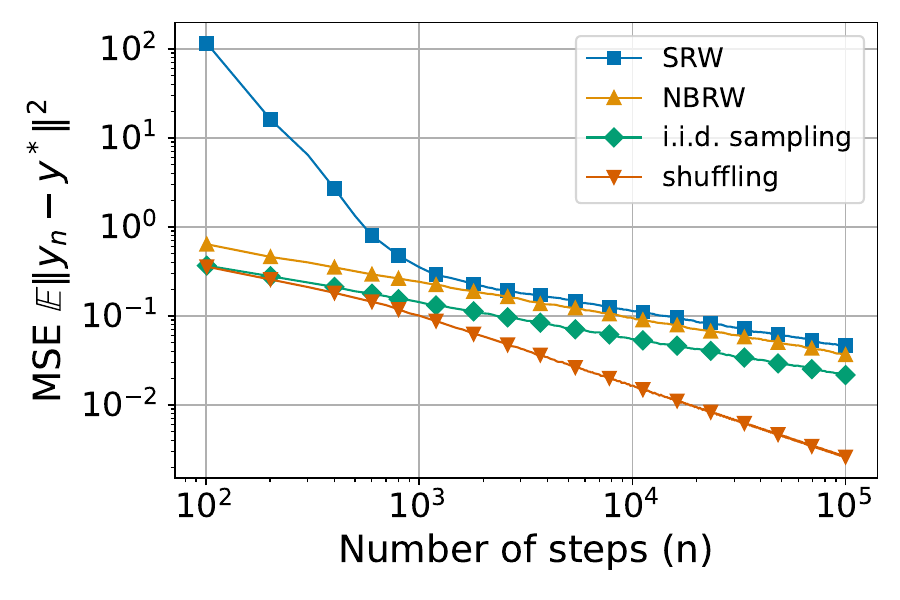}
        \caption{MSE}
        \label{fig:7a}
    \end{subfigure}
    \hfill
    \begin{subfigure}[b]{0.48\columnwidth}
        \centering
        \includegraphics[width=\textwidth]{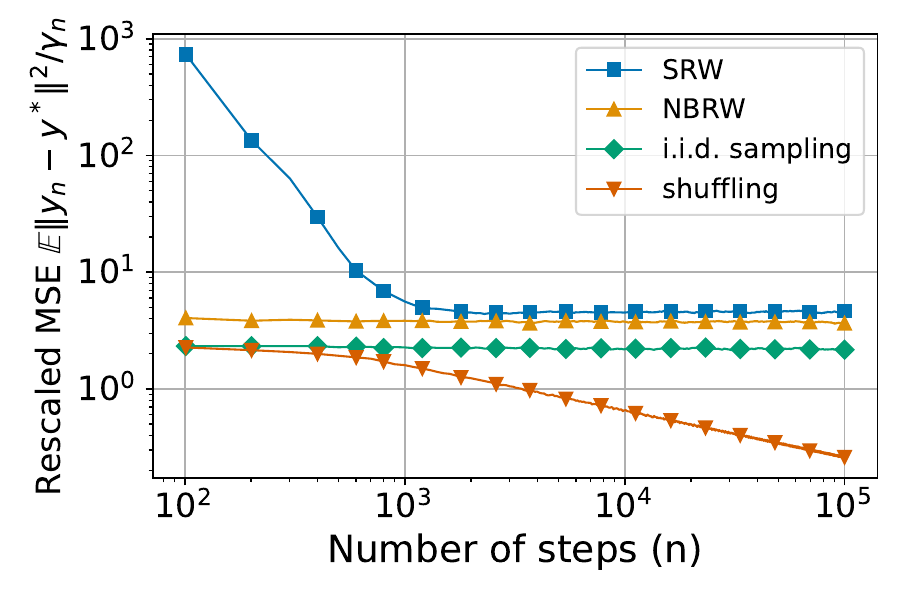}
        \caption{Rescaled MSE}
        \label{fig:7b}
    \end{subfigure}
    \hfill
    \caption{Comparison of the performance ordering in SGDA in terms of iterates $\rvy_n$.}
    \label{fig:7}
\end{figure}

In both Figure \ref{fig:6a} and Figure \ref{fig:7a}, we observe that NBRW has a smaller MSE than SRW across all time $n$ in iterates $\rvx_n$ and iterates $\rvy_n$, with a similar trend for single shuffling over \textit{i.i.d.} sampling. Figure \ref{fig:6b} and Figure \ref{fig:7b} demonstrate that for both iterates $\rvx_n$ and iterates $\rvy_n$, the rescaled MSEs of NBRW, SRW and \textit{i.i.d.} sampling approach some constants, while the curve for single shuffling still decreases in linear rate because eventually the limiting covariance matrix therein will be zero. This simulation result, along with the one in Section 3.1, demonstrates the effectiveness of Proposition 3.2 in Section 3.1, even under the finite-time regime.

\begin{figure}[ht]
    \centering
    \includegraphics[width = 0.8\textwidth]{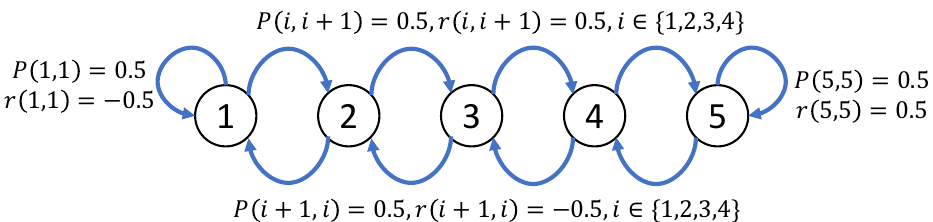}
    \caption{$5$-state random walk problem.}
    \label{fig:diagram}
\end{figure}
\subsection{Random Walk Task for GTD2 and TDC algorithms in Section 3.2}

\subsubsection{Simulation Setups and Computation of $\rmV_{\rvx}$}
For the $5$-state random walk task, the problem setting is given in Figure \ref{fig:diagram}. Using the value iteration algorithm, we obtain the true value function $W(s) = 0$ for $s = \{1,2,3,4,5\}$. In the simulation in Section 3.2, we consider the discount factor $\alpha = 0.9$, and the nonlinear function approximation $W_{x}(s) = a(s)(e^{0.1 x} - 1)$, where $a = [-2, -6, -3, -4, -5]$, and the optimal parameter $x^* = 0$. Then, $\phi_{x}(s) = 0.1 \cdot a(s) e^{0.1x}$ and $\nabla_{x}\phi_{x}(s) = 0.01 \cdot a(s) e^{0.1x}$. In both GTD2 and TDC algorithms, we set the step sizes $\beta_n = (n+1)^{-0.6}, \gamma_n = (n+1)^{-0.501}$.

Now, we leverage the expression in Appendix \ref{appendix:4.2} to calculate the theoretical value of $\rmV_{\rvx}$,\footnote{In this task, all matrices, e.g., $\rmV_{\rvx}, \rmK_{\rvx}, \rmC_*, \rmA_*, \rmU_{\rvx}, \rmU_{\rvy}$, degenerate to scalars.} which is used for Figure 3 in Section 3.2.
In particular, $\mu_i = 0.2$ for $i\in\{1,2,\cdots,5\}$, and
\begin{align*}
    \rmC_{*} = \E_{\vmu}\left[\phi_{x^*}(s_n)^2\right] = \sum_{i=1}^5\frac{1}{5}\cdot 0.01\cdot a(i)^2 = 0.18.
\end{align*}
Note that $\rmV_{\rvx}$ now becomes $\rmV_{\rvx} = 2\rmK_{\rvx}^{-1}\cdot \rmU_{\rvx} = \frac{1}{2}\rmC_*^{-1}\rmU_{\rvy} = \frac{25}{9}\rmU_{\rvy}$. Then, we run the simulation to estimate $\rmU_{\rvy}$ of the form in \eqref{eqn:exact_form_CLT_GTD2_TDC}, which gives $\rmU_{\rvy} \approx 0.0174$ with $100$ independent trials. Thus, we have $\rmV_{\rvx} \approx 0.0484$.

\subsubsection{Additional Choice of Nonlinear Function Approximation}
In this part, we conduct the $5$-state random walk task for GTD2 and TDC algorithms with step sizes $\beta_n = (n+1)^{-0.6}, \gamma_n= (n+1)^{-0.501}$, and another choice of the nonlinear function approximation, i.e.,  $W_{x}(s) = 0.1\cdot (x + \sin(x))$, which becomes the ground truth $W(s) = 0$ by setting $x^* = 0$. Then, we have 
$$\rmC_* = \E_{\vmu}\left[\phi_{x^*}(s_n)^2\right] = \sum_{i=1}^5\frac{1}{5}\cdot 0.04\cdot a(i)^2 = 0.72,$$
such that $\rmV_{\rvx} = \frac{1}{2}\rmC_* \rmU_{\rvy} = \frac{25}{36}\rmU_{\rvy}$. Similarly, we run the simulation to compute $\rmU_{\rvy} \approx 0.1384$ so that $\rmV_{\rvx} \approx 0.0961$.

\begin{figure}[ht]
    \centering
    \begin{subfigure}[b]{0.48\columnwidth}
        \centering
        \includegraphics[width=\textwidth]{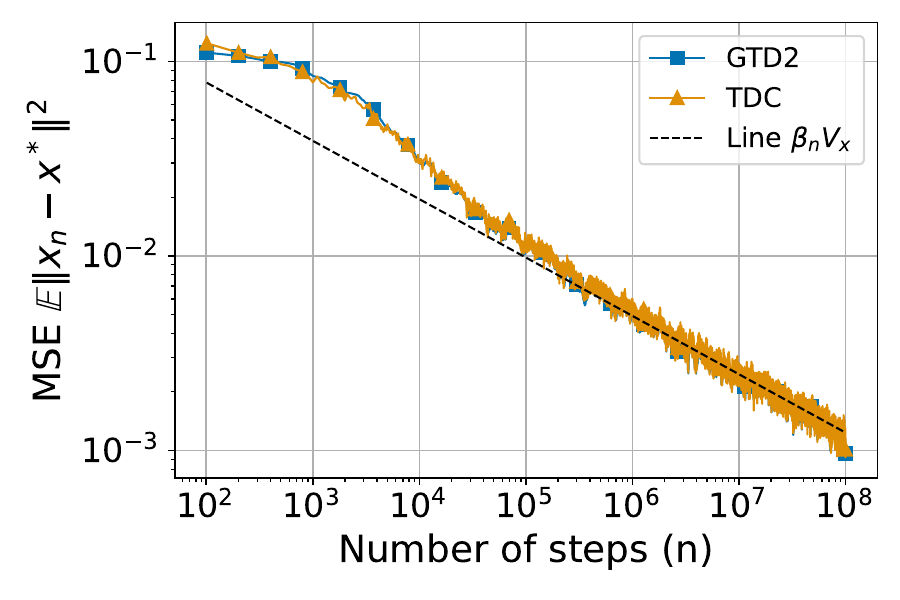}
        \caption{MSE}
        \label{fig:4a}
    \end{subfigure}
    \hfill
    \begin{subfigure}[b]{0.48\columnwidth}
        \centering
        \includegraphics[width=\textwidth]{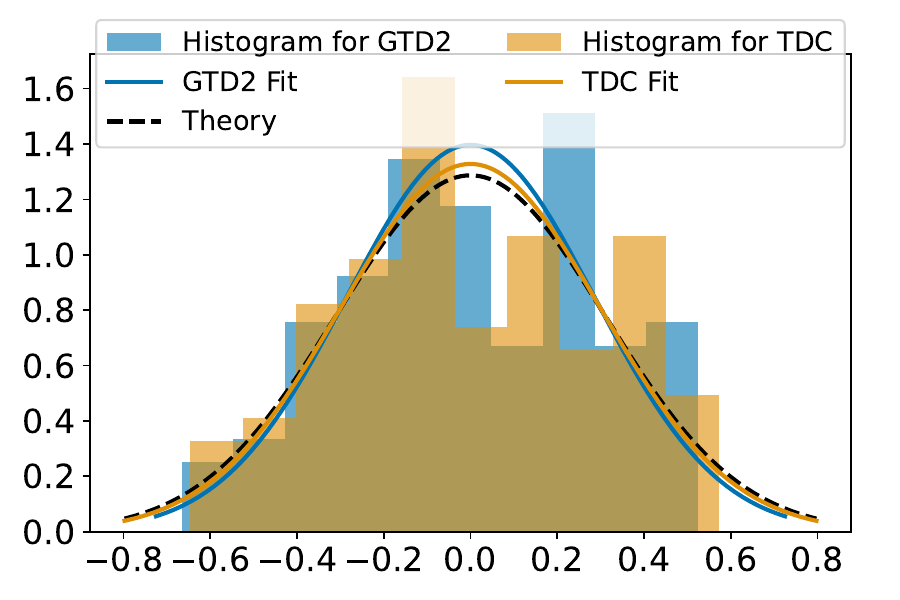}
        \caption{Histograms of $\beta_n^{-1/2}x_n$ at $n = 10^8$}
        \label{fig:4b}
    \end{subfigure}
    \hfill
    \caption{Comparison of nonlinear GTD2 and TDC algorithms in the $5$-state random walk task with the nonlinear function approximation $W_x(s) = 0.1 \cdot(x + \sin(x))$.}
    \label{fig:4}
\end{figure}

Figure \ref{fig:4} shows the long-term performance of both the GTD2 and TDC algorithms, as well as the deviation from the optimal value $x^*$ at time $n = 10^8$. This is in agreement with our Proposition 3.3 in Section 3.2. Specifically, we show in \ref{fig:4a} that for this choice of nonlinear function approximation, the GTD2 and TDC algorithms achieve almost the same performance starting from $n= 10^4$. This is because $W_x(s) = 0.1 \cdot (x + \sin(x))$ acts more like a linear function in $x$ than the selection $W_x(s) = a(s)(e^{0.1x}-1)$ in Section 3.2, which means the effect of $f_n(\rvx,\rvy)$ introduced by the nonlinear approximation is reduced in the iterates $\rvx_n$ for GTD2 and TDC algorithms, thus diminishing the performance gap between these two algorithms. Figure \ref{fig:4b} represents the histogram of $\beta_n^{-1/2} x_n$ for both algorithms from $100$ independent experiments. Their experimental density curves approach the theoretical Gaussian curve with zero mean and variance $\rmV_{\rvx} = 0.0961$. 

\section{Useful Theoretical Results}\label{appendix:technical_lemmas}
\begin{lemma}[Abel Transformation]\label{lemma:abel_transformation}
    Suppose $\{f_k\}$ and $\{g_k\}$ are two sequences. Then,
    \begin{equation*}
        \sum_{k=m}^n f_k(g_{k+1 - g_k}) = f_n g_{n+1} - f_m g_m - \sum_{k=m+1}^n g_k(f_k - f_{k-1}).
    \end{equation*}
\end{lemma}

\begin{lemma}[Burkholder Inequality, \cite{davis1970intergrability}, \citet{hall2014martingale} Theorem 2.10]\label{lemma:Burkholder Inequality}
Given a Martingale difference sequence $\{M_{i,n}\}_{i=1}^n$, for $p\geq 1$ and some positive constant $C_p$, we have
\begin{equation}
    \E\left[\left\|\sum_{i=1}^n M_{i,n}\right\|^p\right] \leq C_p \E\left[\left(\sum_{i=1}^n \left\|M_{i,n}\right\|^2\right)^{p/2}\right]
\end{equation}
\end{lemma}

\begin{theorem}[Martingale CLT, \citet{delyon2000stochastic} Theorem 30]\label{theorem:clt_martingale} 
If a Martingale difference array $\{X_{n,i}\}$ satisfies the following condition: for some $\tau > 0$,
\begin{equation}
    \sum_{k = 1}^n \E\left[\|X_{n,k}\|^{2+\tau} | \gF_{k-1}\right] \xrightarrow[]{\mathbb{P}} 0, \quad \sup_{n} \sum_{k=1}^n \E\left[\|X_{n,k}\|^{2} | \gF_{k-1}\right] < \infty, \quad \sum_{k=1}^n \E\left[X_{n,k}X_{n,k}^T | \gF_{k-1}\right] \xrightarrow[]{\mathbb{P}} \mV,
\end{equation}
then 
\begin{equation}
    \sum_{i=1}^n X_{n,i} \xrightarrow[]{~d~} N(0,\mV). 
\end{equation}
\end{theorem}

\begin{lemma}[\cite{duflo1996algorithmes} Proposition 3.I.2]\label{lemma:matrix_norm_inequality}
    For a Hurwitz matrix $\rmH$, there exist some positive constants $C, b$ such that for any $n$,
    \begin{equation}
        \left\|e^{\rmH n}\right\| \leq Ce^{-bn}.
    \end{equation} 
\end{lemma}

\begin{lemma}[\cite{fort2015central} Lemma 5.8]\label{lemma:matrix_product_norm_inequality}
    For a Hurwitz matrix $\rmA$, denote by $-r$, $r>0$, the largest real part of its eigenvalues. Let a positive sequence $\{\gamma_n\}$ such that $\lim_n \gamma_n = 0$. Then for any $0<r'<r$, there exists a positive constant $C$ such that for any $k < n$,
    \begin{equation}
        \left\|\prod_{j=k}^n (\rmI + \gamma_j\rmA)\right\| \leq C e^{-r' \sum_{j=k}^n \gamma_j}.
    \end{equation}
\end{lemma}

\begin{lemma}[\cite{fort2015central} Lemma 5.9, \cite{mokkadem2006convergence} Lemma 10]\label{lemma:upperbound_of_sequence_x}
    Let $\{\gamma_n\}$ be a positive sequence such that $\lim_n \gamma_n = 0$ and $\sum_n \gamma_n = \infty$. Let $\{\epsilon_n, n\geq 0\}$ be a nonnegative sequence. Then, for $b > 0$, $p \geq 0$,
    \begin{equation}
        \limsup_n \gamma_{n}^{-p} \sum_{k=1}^n \gamma_k^{p+1} e^{-b \sum_{j=k+1}^n \gamma_j} \epsilon_k \leq \frac{1}{C(b,p)} \limsup_n \epsilon_n
    \end{equation}
    for some constant $C(b,p) > 0$. 
    
    When $p = 0$ and define a positive sequence $\{w_n\}$ satisfying $w_{n-1}/w_n = 1 + o(\gamma_n)$, we have
    \begin{equation}\label{eqn:66}
        \sum_{k=1}^n \gamma_k e^{-b \sum_{j=k+1}^n \gamma_j} \epsilon_k = \begin{cases}
            O(w_n), & \quad \text{if~} \epsilon_n = O(w_n), \\ o(w_n), & \quad \text{if~} \epsilon_n = o(w_n).
        \end{cases}
    \end{equation}
\end{lemma}

\begin{lemma}[\cite{fort2015central} Lemma 5.10]\label{lemma:matrix_inequality}
    For any matrices $A,B,C$,
    \begin{equation}
        \|ABA^T - CBC^T\| \leq \|A-C\| \|B\| (\|A\|+\|C\|).
    \end{equation}
\end{lemma}

\begin{lemma}\label{lemma:closed_form_solution_lyapunov_equation}
    Suppose $\rmK$ is a Hurwitz matrix. Then, for any positive semi-definite matrix $\rmU$, there exists a unique positive semi-definite matrix $\rmV$ such that $\rmK\rmV + \rmV\rmK^T +\rmU = 0$ (Lyapunov equation), where the closed form of $\rmV$ is given by
    \begin{equation}
        \rmV = \int_{0}^{\infty} e^{t\rmK} \rmU e^{t\rmK^T} dt.
    \end{equation}
\end{lemma}
Lemma \ref{lemma:closed_form_solution_lyapunov_equation} come from Theorem 3.16 \cite{chellaboina2008nonlinear}. Nevertheless, they necessitate a positive definite matrix $\rmU$ so that the solution $\rmV$ is also positive definite. Throughout this paper, we do not require the solution $\rmV$ to be positive definite so that the matrix $\rmU$ can be relaxed to be positive semi-definite. This relaxation does not change any steps as in the proof of Theorem 3.16 \cite{chellaboina2008nonlinear}, and is thus omitted here.
 
\end{document}